\documentclass[preprint,12pt]{elsarticle}

\usepackage{float}
\usepackage{url}
\usepackage{amssymb}
\usepackage{amsmath}
\usepackage{amsthm}
\usepackage{bm}
\usepackage{multirow}
\usepackage{booktabs,multirow,array}
\usepackage{siunitx}
\newcolumntype{L}[1]{>{\raggedright\arraybackslash}p{#1}} 
\newcolumntype{C}[1]{>{\centering\arraybackslash}p{#1}}   
\usepackage{placeins}

\usepackage{booktabs}
\usepackage{tabularx}
\usepackage{makecell}
\usepackage{multirow}
\usepackage{algorithm}
\usepackage{algpseudocode}
\usepackage{rotating}
\usepackage{pdflscape} 
\usepackage{booktabs}
\usepackage{tabularx}
\usepackage{makecell}
\usepackage{graphicx}  
\usepackage{xcolor}

\usepackage{ragged2e}  
\usepackage{amsmath}
\usepackage{bm}
\usepackage{booktabs}
\usepackage{makecell}
\usepackage{array}
\usepackage{tabularx}

\newcolumntype{Y}{>{\centering\arraybackslash}X}
\newcolumntype{Y}[1]{>{\raggedright\arraybackslash}p{#1}}
\newcolumntype{Z}[1]{>{\centering\arraybackslash}p{#1}}

\usepackage{booktabs}
\usepackage{siunitx}
\journal{}

\begin{document}

\begin{frontmatter}



\title{BiLoGo: Reducing the smoothing of localised responses in Transformer-based neural operators for sheet forming}
\author[1]{Yingxue Zhao}
\author[1]{Haoran Li}
\author[1]{Haosu Zhou}
\author[2]{Tobias Pfaff}
\author[1]{Nan Li\corref{cor1}}

\affiliation[1]{organization={Dyson School of Design Engineering}, 
               addressline={Imperial College London}, 
               city={London},
               country={UK}}

\affiliation[2]{organization={Prometheus}, 
               country={UK}}
\cortext[cor1]{Corresponding author. 
E-mail address: {n.li09@imperial.ac.uk} (N. Li)}


\begin{abstract}
In engineering design, trial-and-error Finite Element (FE) simulation of sheet forming can become computationally expensive. Neural-network-based learned simulators have been developed to accelerate design. In sheet forming, the contact between the tools and the blank can result in localised responses, such as thinning around sharp radii, on the side walls of deep-drawn corners or near reinforcing ribs. These localised responses often determine whether a component can be formed without failure. Learned simulators therefore need to capture them accurately and efficiently. Among these learned simulators, Transformer-based neural operators propagate information efficiently across the mesh, but tend to smooth such localised responses.
This study develops BiLoGo, a Bipartite Local--Global learned simulator with an FE-guided design that connects local node--element message passing with a Transformer-based neural operator for long-range information exchange. BiLoGo consists of three modules. First, a bidirectional node--element GraphSAGE model learns the coupling between nodal and elemental features on a bipartite graph, and predicts each quantity on its native FE entity without interpolation. Second, a dual-stream Transformer-based neural operator processes the blank deformation and the tool contact with two separate sets of tokens, so that the relatively localised contact information is not averaged with the state of the whole blank. Third, a fusion and element update module combines the global update with the local features by gated residual fusion, and updates the element features conditioned on the kinematics of the element nodes. The model is evaluated on a dome-shaped cold forming case and two corner-shaped hot stamping cases, one with ribs. Tested against single-level and hierarchical graph-based models and Transformer-based neural operators, BiLoGo gives the lowest positional and threshold-based thinning errors on all three cases. Near the ribs, the smoothing index of the predicted thinning increases from 0.41--0.56 for the Transformer-based neural operators to 0.94 for BiLoGo-U. Compared with the most accurate existing model, the hierarchical graph network CAtt-BiUGNN, BiLoGo-T reduces the mean positional error by 15.5\% on the Corner case and BiLoGo-U by 8.3\% on the Corner+Rib case, with 66--69\% less training memory.
\end{abstract}

\begin{highlights}
\item BiLoGo couples node--element message passing with a Transformer neural operator
\item A forming dataset with reinforcing ribs of varying position and size is introduced
\item Smoothing index near ribs: 0.94 for BiLoGo-U vs 0.41--0.56 for token-only models
\item Lowest positional error on all cases with 66--69\% less memory than a hierarchical GNN
\item Spatially local tokens reduce the errors near local geometric features
\end{highlights}



\end{frontmatter}

\section{Introduction}
\label{sec:introduction}

Finite Element (FE) simulation is widely used in the design of sheet-formed components.
However, the design often follows a trial-and-error process, in which each iteration requires a full explicit dynamic analysis.
This can limit the number of design candidates that can be evaluated within the available design time.
Machine Learning (ML) models, including learned simulators that predict the evolution of the FE solution step by step, have been developed to accelerate this process \cite{Attar2021Feasibility,Wang2026Springback,Zhou2025Multilevel,Luo2026StampFormer,Zhao2026RUGNN}.
For sheet forming, a learned simulator needs to predict both the structural-scale deformation and the local mechanical response, including the coupled nodal displacement and elemental thinning, over an autoregressive rollout \cite{Zhao2026RUGNN,Zhao2026CAttBiGNN,Wang2026Springback}.
In many forming problems, the most important part of these responses is localised.
The largest thinning, which is an important indicator of whether a component can be formed without splitting, is often concentrated in small regions, such as the side walls of deep-drawn corners.
In addition, local geometric features, such as reinforcing ribs, are often added to a component to increase its stiffness \cite{Cha2018BeadFormability,Ott2021BeadVariants}.
Although these features are small relative to the overall component, they can affect the local material flow and may result in localised thinning in their vicinity \cite{Cha2018BeadFormability}.
Therefore, accurate prediction of these localised responses is the primary requirement for a learned simulator.
In addition to accuracy, the computational requirements determine whether a learned simulator can be used in practical engineering design.
This is particularly important for localised responses, since resolving them requires a denser mesh, which increases the size of the graph and the GPU memory consumption.
A learned simulator therefore needs to maintain its accuracy with lower computational requirements.

Graph-based models, such as Graph Neural Networks (GNNs), are well suited to learned simulation \cite{Pfaff2021MeshGraphNet,Maurizi2022MechanicalFields,Zhao2024GNNReview,Chen2024PIERS}.
They operate on graphs, in which the relations between nodes are described by edges, and an FE mesh can be converted naturally into such a graph.
In sheet forming, the nodal motion is governed by internal forces, external loading and contact, while the deformation of an element, and therefore its thinning, is determined by the relative motion of its nodes \cite{Zhao2026CAttBiGNN,Gao2024FEIH,Wang2026NEHGNN}.
These node--element relations can be represented by a graph in which nodes and elements are separate entities, so that each quantity is predicted on its own FE entity without interpolation and the potential loss of information \cite{Zhao2026CAttBiGNN}.
In addition, since message passing exchanges information between neighbouring entities, GNNs are well suited to capturing the local interactions in the FE model.
However, one learned step usually covers many solver time steps \cite{Pfaff2021MeshGraphNet,Iparraguirre2026MGNT}.
Larger learned steps reduce the number of autoregressive steps in a rollout, which reduces the inference time and the accumulation of prediction errors \cite{Brandstetter2022MPPDE}.
Within one learned step, the contact between the tools and the metal sheet, referred to as the blank, changes with the tool position.
These contact changes and other mechanical disturbances can affect regions far from where they occur.
On a fine mesh, reaching such distances by message passing alone requires many message-passing layers, and information from distant nodes can be compressed excessively in the fixed-size node features \cite{Cao2023BSMS,Fortunato2022MSMGN,Iparraguirre2026MGNT,Alon2021Bottleneck}.
A learned simulator for sheet forming therefore needs both local FE processing and long-range information propagation.

Hierarchical GNNs achieve long-range propagation through graph coarsening.
Hierarchical models such as MultiScale MeshGraphNets, MAgNET, BSMS-GNN and ROBIN reduce the number of message-passing operations on the original mesh by processing information on coarser graphs \cite{Fortunato2022MSMGN,Deshpande2024MAgNET,Cao2023BSMS,Wurth2025ROBIN}.
For sheet forming, a multi-level graph-based model with spatial edge weights inspired by convolution kernels predicts the post-stamped thickness from the blank shape \cite{Zhou2025Multilevel}.
RUGNN processes local interactions on the fine graph, propagates information over long distances on the coarsest graph, and refines local features after upsampling \cite{Zhao2026RUGNN}.
CAtt-BiUGNN extends this design to a node--element bipartite graph hierarchy for coupled nodal and elemental prediction \cite{Zhao2026CAttBiGNN}.
In addition to extending the receptive field, these hierarchical models preserve local information through their structure.
First, information is compressed gradually over several graph levels.
Second, each fine-level node is connected only to its nearest coarse-level nodes.
Third, in CAtt-BiUGNN, nodes and elements are kept as separate entities at every level.
Fourth, skip connections pass the features stored at each level back to the fine graph during upsampling \cite{Zhao2026RUGNN,Zhao2026CAttBiGNN}.
The importance of gradual compression was shown in RUGNN, where a large node reduction ratio between levels caused a loss of local information \cite{Zhao2026RUGNN}.
Edge-specific inter-level weights can improve the information transfer between graph levels compared with message passing between levels \cite{Li2026Crashworthiness}.
However, the number of these weights grows with the number of inter-level edges and therefore with the mesh size \cite{Zhao2026CAttBiGNN,Li2026MaskMorph}.
As a result, hierarchical models with such weights can require large GPU memory on fine meshes.

Neural operators provide another approach to long-range propagation.
They learn mappings between function spaces, and many of them exchange information over the whole domain within each layer, for example through integral kernels, Fourier transforms or attention \cite{Lu2021DeepONet,Li2021FNO,Kovachki2023NeuralOperator}.
In contrast, a message-passing layer of a mesh-based GNN only exchanges information between neighbouring entities.
Among neural operators, Transformer-based neural operators replace the graph hierarchy with a single learned compression step \cite{Wu2024Transolver,Luo2025TransolverPP,Iparraguirre2026MGNT,Zhou2026Transolver3}.
Transolver assigns mesh points to a small number of learned physical slices, performs global interaction between the resulting tokens, and maps the result back to the mesh \cite{Wu2024Transolver}.
This makes global propagation computationally efficient, since the cost of the global interaction depends on the number of tokens rather than on the number of mesh points, and no latent grid is required \cite{Wu2024Transolver,Alkin2024UPT}.
However, the features of all mesh entities are recovered from the same few tokens, which may smooth localised responses with large spatial gradients.
Combining token-based global interaction with local message passing can improve the accuracy, as shown for particle-based fluid simulation \cite{Wu2024Transolver} and for impact and deformation problems in solid mechanics \cite{Iparraguirre2026MGNT}.
However, these models predict nodal or particle quantities, and the accuracy near localised responses was not assessed separately.
In addition, how the local information lost in the compression can be kept and recovered, in particular for element-level responses such as thinning, has received limited attention.

To address these limitations, a Bipartite Local--Global (BiLoGo) learned simulator is developed in this study.
BiLoGo uses a Transformer-based neural operator for long-range propagation instead of a graph hierarchy.
Local information is processed on a node--element graph before and after the global module, and is then combined with the global update.
BiLoGo consists of three modules.
First, a bidirectional node--element GraphSAGE model is applied before and after the global module.
It exchanges information between nodes and elements through the FE connectivity.
Nodes and elements are kept as separate entities, and each quantity is predicted on its own FE entity.
Second, a dual-stream Transformer-based neural operator is used for global propagation.
The deformation state of the blank and the tool contact information are compressed into two separate sets of tokens.
This is because the tool contact is relatively localised and changes with the tool position.
Its information may therefore be averaged out if it is compressed together with the state of the whole blank.
The contact tokens then query the state tokens through cross-attention, and the result is mapped back to the mesh.
Third, a fusion and element update module combines the global update with the local features.
Gated fusion is applied separately to nodes and elements, so that the global update does not overwrite the local information.
The element features are then updated conditioned on the relative positions and displacements of the element nodes.
The tokeniser in the second module can be replaced, and two variants are therefore developed.
BiLoGo-T uses the learned physical slices of Transolver \cite{Wu2024Transolver}.
BiLoGo-U uses the supernode pooling and position-based decoder of UPT \cite{Alkin2024UPT}, and the supernodes are used directly as tokens without further compression.
All other components are the same in both variants.

BiLoGo is evaluated on three forming problems.
In the Dome dataset, the blank is gradually formed into a double-curvature shape.
The thinning is mainly concentrated near the dome radius and changes relatively smoothly with the overall geometry, so that the localised response is relatively weak.
In the Corner and Corner+Rib datasets, the localised responses are stronger and are resolved on finer meshes.
In the Corner dataset, the global geometry varies, and the largest thinning is concentrated near the corner.
In the Corner+Rib dataset, the global geometry is fixed and only the position and size of the ribs vary, so that the localised response is mainly caused by the change of local geometric features.
The three datasets are used to examine how the benefit of BiLoGo changes from a response with weaker localisation to responses with stronger localisation.
In addition, the Corner and Corner+Rib datasets allow the localisation caused by changes in the global geometry and by local geometric features to be examined separately.

The main contributions of this work are as follows.
First, a Corner+Rib dataset is introduced to study forming problems in which local geometric features change the response only in local regions, which have received limited attention for learned simulators. In this dataset, the global geometry is fixed and only the position and size of the reinforcing ribs vary.
Second, BiLoGo is developed, which combines a bidirectional node--element GraphSAGE model and a fusion and element update module with a dual-stream Transformer-based neural operator for coupled nodal and elemental prediction. Compared with Transformer-based neural operators that use the same types of tokeniser, BiLoGo reduces the smoothing of the predicted thinning near the corner and the ribs.
Third, BiLoGo is compared with single-level and hierarchical graph-based models and Transformer-based neural operators on the three datasets. It gives the lowest displacement and threshold-based thinning errors on all three datasets, and requires considerably less training memory than the most accurate existing model, a hierarchical graph network. In addition, its number of parameters does not change with the mesh. Therefore, it does not require more parameters on finer meshes, and simulations with different mesh topologies may be used to train the same model.
Fourth, the global module of BiLoGo can be replaced, and two variants, BiLoGo-T with physical slices and BiLoGo-U with spatial supernodes, are compared. The results indicate that assigning the mesh entities to tokens by their features can be sufficient when the geometry varies at the scale of the whole component, whereas compressing the mesh entities by their spatial positions may be preferred when the response is concentrated near local geometric features.

The remainder of this paper is organised as follows.
Section~\ref{sec:related} reviews mesh-based GNNs with FE-structured and hierarchical representations, neural operators, and machine learning models for sheet forming.
Section~\ref{sec:problem} defines the learned update problem and analyses the properties of the explicit FE update that guide the model design.
Section~\ref{sec:dataset} describes the generation of the forming datasets.
Section~\ref{sec:graph} describes the graph construction and the input and output features.
Section~\ref{sec:method} introduces the three modules of BiLoGo and the two tokenisers.
Section~\ref{sec:setup} describes the training settings of BiLoGo.
Section~\ref{sec:results} introduces the baseline models and evaluation metrics, presents the results on prediction accuracy, architecture ablations and computational cost, and discusses the selection of the global module and the limitations of this work.
Section~\ref{sec:conclusion} concludes the paper.
\ref{app:slices} shows the slice assignment of BiLoGo-T on the Dome and Corner datasets.

\section{Related work}
\label{sec:related}

This section reviews the work related to BiLoGo, from general learned simulators to sheet forming.
Section~\ref{sec:related_gnn} reviews mesh-based GNNs, including FE-structured and hierarchical representations.
Section~\ref{sec:related_operators} reviews neural operators, including the token-based operators used as the global module of BiLoGo.
Section~\ref{sec:related_forming} reviews machine learning models for sheet forming.

\subsection{Mesh-based graph neural networks in computational mechanics}
\label{sec:related_gnn}

GNNs have been widely used as learned simulators in computational mechanics, since an FE mesh can be represented directly as a graph.
MeshGraphNets \cite{Pfaff2021MeshGraphNet} established the encoder--processor--decoder framework for mesh-based learned simulation, in which node and edge features are updated by message passing along mesh edges and the model is rolled out autoregressively.
Similar message-passing models have been used to predict stress, strain and deformation fields \cite{Maurizi2022MechanicalFields,Zhao2024GNNReview,Gladstone2024SteadyGNN}, dynamic responses of deformable bodies \cite{Chen2024PIERS,Li2025DynamicImpactGNN}, contact between deformable bodies \cite{Dubey2026ContactGNN,Tesan2025DigitalHuman}, the crashworthiness of vehicle panel components \cite{Li2026Crashworthiness,Li2026MaskMorph}, and local phenomena such as crack propagation and plastic strain localisation \cite{Perera2022CrackGNN,Zhou2024CrackMPNN,Jakobsen2026PlasticLocalization}.
Reviews of these developments are given in \cite{Zhao2024GNNReview,Herrmann2024DLMechanicsReview}.
Physical knowledge has also been included in graph models, for example by training with the residual of the governing equations, which reduces the need for simulation data \cite{Wurth2024PIMGN}, or by building the laws of thermodynamics into the model structure \cite{Hernandez2024ThermoGNN,Tierz2025LocalThermoGNN}.

Most mesh-based GNNs are node-centred.
Several studies have introduced FE-structured representations, in which elements are represented as separate entities.
A finite element-inspired hypergraph network represents elements as hyperedges connecting their nodes \cite{Gao2024FEIH,Gao2024FSI}.
Heterogeneous graph networks have been used for nonlinear structural analysis \cite{Li2025HeterogeneousNonlinear} and tube forming \cite{Wang2026NEHGNN}.
CAtt-BiGNN represents mesh nodes and element centres as two node types connected by directed node--element edges, so that nodal displacement and elemental thinning are predicted on their own entities \cite{Zhao2026CAttBiGNN}.
These representations describe the local node--element relationships.
However, on fine meshes, single-level message passing requires many layers to propagate information over long distances, which has led to the development of hierarchical GNNs.

Hierarchical GNNs extend the receptive field by exchanging information between graphs of different resolutions.
The coarse graphs can be generated from coarser meshes \cite{Fortunato2022MSMGN,Zhao2026RUGNN}, including the mesh levels of adaptive mesh refinement, which change during the simulation \cite{Perera2024AdaptiveMesh}.
They can also be constructed by merging the nodes within each cell of a background grid \cite{Zhou2025Multilevel}, from the connectivity of the original mesh \cite{Cao2023BSMS,Yang2022AMGNET,Wurth2025ROBIN,Deshpande2024MAgNET}, or learned together with the dynamics \cite{Deng2025EvoMesh}.
For example, BSMS-GNN selects the nodes on every other frontier of a breadth-first search \cite{Cao2023BSMS}, AMGNet and ROBIN coarsen the graph by algebraic multigrid \cite{Yang2022AMGNET,Wurth2025ROBIN}, and MAgNET partitions the graph into fully connected subgraphs and contracts each of them into one coarse node \cite{Deshpande2024MAgNET}.
In contrast, EvoMesh learns which nodes are kept at each coarser level from the current physical state \cite{Deng2025EvoMesh}.
Other studies use multiscale autoencoders or rotation-equivariant multiscale processing \cite{Barwey2023GraphAutoencoder,Lino2022RotationMultiscale}, and some combine local graph processing with regional or segment-level processing \cite{Mousavi2025RIGNO,Lei2025M4GN}.
Graph U-Net architectures \cite{Deshpande2024MAgNET,Zhou2025Multilevel,Zhao2026RUGNN,Li2026MaskMorph} pass information down to coarse levels and back to the fine level, and use skip connections to reuse the fine-level features during upsampling.
CAtt-BiUGNN applies this approach to a node--element bipartite hierarchy, in which nodes and elements are coarsened separately and connected to coarser levels through $k$-nearest-neighbour mappings with learnable edge-specific weights \cite{Zhao2026CAttBiGNN}.
Compared with a single-level GNN, a hierarchy reduces the number of message-passing layers needed to reach distant regions \cite{Fortunato2022MSMGN,Cao2023BSMS}.
This can mitigate the over-smoothing of node features in deep message-passing stacks, and can reduce the computational time and memory \cite{Cao2023BSMS,Lei2025M4GN}.
However, when the inter-level mappings use edge-specific weights, the number of parameters grows with the mesh size, and the GPU memory can become large on fine meshes \cite{Zhao2026CAttBiGNN,Li2026MaskMorph}.

\subsection{Neural operators in computational mechanics}
\label{sec:related_operators}

Neural operators learn mappings between function spaces \cite{Lu2021DeepONet,Li2021FNO,Kovachki2023NeuralOperator}.
They have been applied to solid mechanics problems, including heterogeneous deformation, geometry-dependent mechanical response and formability evaluation \cite{Rezaei2025FiniteOperator,Kaewnuratchadasorn2025GeoPhysicsNO,Nguyen2026FormabilityDeepONet}.

Neural operators can be grouped by how the global interaction is parameterised \cite{Kovachki2023NeuralOperator,Azizzadenesheli2024NOReview}.
First, branch--trunk operators, such as DeepONet, combine a branch network for the input function with a trunk network for the output location, and are supported by a universal approximation theorem for operators \cite{Lu2021DeepONet}.
The trunk network can evaluate the output at arbitrary locations, while the input function is evaluated at the same set of sensor locations for all samples, so that an input on a different mesh is first sampled at these locations \cite{Lu2021DeepONet}.
Second, graph-based operators, such as the graph neural operator, integrate over a fixed-radius neighbourhood and can be applied directly to irregular meshes.
However, the range of each layer is limited when the radius is small, and the cost can increase considerably when the radius is enlarged \cite{Kovachki2023NeuralOperator,Azizzadenesheli2024NOReview}.
Third, grid-based operators, such as the Fourier Neural Operator (FNO), perform the global interaction in the Fourier domain with a quasi-linear cost on a regular grid \cite{Li2021FNO}.
For irregular meshes, the mesh needs to be mapped to a latent grid, as in the Geometry-Informed Neural Operator (GINO) \cite{Li2023GINO}, and a fine three-dimensional latent grid can require large GPU memory and long training times \cite{Alkin2024UPT}, as also observed in Section~\ref{sec:results_cost}.
Fourth, attention-based operators can be applied directly to irregular meshes and complex geometries, but standard attention scales quadratically with the number of mesh points \cite{Azizzadenesheli2024NOReview}.
GNOT reduces this cost by using efficient attention \cite{Hao2023GNOT}.
Token-based operators reduce the cost by compressing the mesh into a small number of tokens before the global interaction.
Transolver assigns mesh points to a small number of learned physical slices based on their features \cite{Wu2024Transolver}, and Transolver++ further introduces locally adaptive slicing \cite{Luo2025TransolverPP}.
The Universal Physics Transformer (UPT) aggregates information from the mesh to spatially distributed supernodes within a fixed radius, and recovers point-wise outputs through a Perceiver-type decoder queried by spatial position \cite{Alkin2024UPT}.

Several models combine local message passing with global attention.
Graph\-GPS combines a message-passing layer and a global attention layer in every block \cite{Rampasek2022GraphGPS}.
Similarly, adding a Transolver layer in parallel to the message passing of a particle-based GNN reduced its position error by 62\% \cite{Wu2024Transolver}.
EAGLE pools clusters of mesh nodes into tokens, performs attention between these tokens and maps the result back to the mesh \cite{Janny2023EAGLE}.
HCMT uses hierarchical and contact mesh Transformers to model collisions between flexible bodies \cite{Yu2024HCMT}.
GAOT places a Transformer processor between multiscale graph-based encoders and decoders \cite{Wen2025GAOT}, and MGN-T places a global Transformer processor between local message-passing stages for solid mechanics \cite{Iparraguirre2026MGNT}.
Compared with these models, BiLoGo processes the nodes and elements as separate FE entities and predicts the thinning on the elements, compresses the blank state and the tool contact into separate tokens, and combines the global update with the local features through gated fusion and an element update based on the nodal kinematics.

Overall, hierarchical GNNs propagate information over long distances by message passing on one or more coarser graphs, whereas many neural operators exchange information over the whole domain within each layer \cite{Kovachki2023NeuralOperator}.
In addition, token-based operators perform the global interaction between a small number of tokens.
Compared with single-level GNNs, some hierarchical GNNs with edge-specific weights, Transformers that process all mesh points in every layer, and grid-based operators, whose regular latent grid becomes large in three dimensions, this can considerably reduce the memory and runtime on large meshes \cite{Wu2024Transolver,Alkin2024UPT}, as also observed in Section~\ref{sec:results_cost}.
However, neural operators tend to learn low-frequency components more easily than high-frequency components, which can lead to smoother predictions for responses with sharp gradients \cite{Khodakarami2026SpectralBias}.
In this work, Transolver and UPT also give smoother thinning near the ribs than the FE result, as shown in Section~\ref{sec:results_smoothing}.
Operators that compress the mesh before the global interaction, such as GINO, Transolver and UPT, may be further affected by how this compression is performed.
In Transolver, the slices group the mesh points by their features, so points that are far apart but have similar features are assigned to the same token.
This allows distant regions with similar physical states to interact directly \cite{Wu2024Transolver}.
However, it may also smooth responses that are concentrated in small regions, since their features are combined with those of distant regions in the same token, as analysed in Section~\ref{sec:method_compression}.
In contrast, the latent grid of GINO and the supernodes of UPT keep the spatial locality of the compression, and enforcing locality has been shown to benefit neural operators in physical simulations \cite{Chen2025LocalityNO}.
Hierarchical GNNs keep the local information through gradual compression and skip connections, but can require large GPU memory on fine meshes, as reviewed in Section~\ref{sec:related_gnn}.
This motivates the use of local node--element processing together with the global module in BiLoGo, and the comparison between slice-based and spatially local tokenisers in this work.

\subsection{Machine learning models for sheet forming}
\label{sec:related_forming}

Machine learning methods have been applied to many problems in sheet forming.
Scalar-output models and physics-guided models have been used to predict quantities such as forming limits and formability \cite{Nguyen2025FormingLimits,Nguyen2026FormabilityDeepONet}.
Graph and point cloud models have also been used to classify the drawability of U-shaped deep-drawn parts with diverse real-world geometries \cite{Lehrer2027GeometryAugmented}.
These models are efficient, but they do not provide the spatial distribution of the mechanical fields.
To predict full fields, image-based Convolutional Neural Networks (CNNs) have been developed for thinning and deformation prediction, feasibility assessment, geometry optimisation under manufacturing constraints and blank shape optimisation \cite{Attar2021HFQ,Attar2021Feasibility,Attar2023Optimisation,Zhou2025BlankShape}.
Related work has also used deep learning to increase the mesh density of physical fields in forming simulations \cite{Xu2022SuperMeshing}, and multimodal models have combined material and geometry information to predict physical fields in stamping \cite{Luo2026StampFormer}.
However, image-based models require the FE results to be mapped onto regular grids, which can reduce geometric accuracy and make it difficult to represent the tool--blank interaction.

In sheet forming, graph-based models have been developed for real-time thickness prediction from the blank shape \cite{Zhou2025Multilevel}, deformation prediction in hot stamping \cite{Zhao2025BatteryBox,Zhao2026RUGNN} and springback prediction under material variations \cite{Wang2026Springback}.
Most of these models predict nodal quantities.
Element-level quantities, such as thinning, are either not predicted or are stored as nodal attributes.
A small number of studies represent nodes and elements as different graph entities, for example for wrinkling analysis in tube forming \cite{Wang2026NEHGNN} and for coupled nodal displacement and elemental thinning prediction in sheet forming \cite{Zhao2026CAttBiGNN}, as reviewed in Section~\ref{sec:related_gnn}.
However, localised responses have received limited attention in these studies.
These include the localised thinning in regions such as deep-drawn corners, the change of this thinning under large geometric variations, and the localised thinning caused by local geometric features, such as reinforcing ribs.

\section{Problem formulation}
\label{sec:problem}

This section defines the learned update problem and analyses the explicit dynamic FE update that the learned simulator approximates.
The analysis is presented in three parts.
Section~\ref{sec:problem_fe} describes how nodes, elements, contact forces and internal forces interact within one explicit update cycle.
Section~\ref{sec:problem_learned} shows that one learned step corresponds to many explicit cycles, so the learned update can depend on information from distant regions of the blank.
Section~\ref{sec:problem_local} shows that the element deformation and thinning are nevertheless mainly determined by the nodes of each element.
These properties guide the design of the model described in Section~\ref{sec:method}.

\subsection{Explicit dynamic FE update}
\label{sec:problem_fe}

The blank is discretised by a mesh with nodes $v \in \mathcal{V}$ and elements $i \in \mathcal{T}$.
In all datasets of this study, three-node linear triangular shell elements are used.
The set of nodes connected to element $i$ is denoted by $\mathcal{N}(i)$, and the set of elements connected to node $v$ is denoted by $\mathcal{M}(v)$.
In the explicit FE solver, time is discretised into solver time steps of size $\Delta t$, and quantities at solver time step $s$ are denoted by the superscript $s$.
The update over one solver time step is referred to as one explicit cycle.
The position of node $v$ is $\mathbf{x}_v^s$, its reference position is $\mathbf{x}_v^0$, and its displacement is $\mathbf{u}_v^s = \mathbf{x}_v^s - \mathbf{x}_v^0$.
A detailed derivation of the FE update is given in \cite{Zhao2026CAttBiGNN}, and only the relations used in this work are summarised here.

The semi-discrete equation of motion is
\begin{equation}
\mathbf{M}\ddot{\mathbf{u}}^s = \mathbf{P}_{\mathrm{ext}}^s - \mathbf{I}_{\mathrm{int}}^s,
\label{eq:motion}
\end{equation}
where $\mathbf{M}$ is the lumped mass matrix, $\mathbf{P}_{\mathrm{ext}}^s$ is the external nodal force vector and $\mathbf{I}_{\mathrm{int}}^s$ is the internal nodal force vector.
The nodal kinematics are updated by the central-difference scheme,
\begin{equation}
\dot{\mathbf{u}}^{s+\frac{1}{2}} = \dot{\mathbf{u}}^{s-\frac{1}{2}} + \Delta t\,\ddot{\mathbf{u}}^s,
\qquad
\mathbf{x}^{s+1} = \mathbf{x}^s + \Delta t\,\dot{\mathbf{u}}^{s+\frac{1}{2}}.
\label{eq:central_difference}
\end{equation}
Here, $\Delta\mathbf{u}^{s+1} = \mathbf{x}^{s+1} - \mathbf{x}^{s}$ is the nodal displacement increment.
The nodal displacement increment is then transferred to the integration points $q$ of each element.
With the Boolean extraction matrix $\mathbf{S}_i$, which selects the degrees of freedom of the nodes in $\mathcal{N}(i)$, the strain increment is
\begin{equation}
\Delta\boldsymbol{\varepsilon}_{iq}^{s+1} \approx \mathbf{B}_{iq}^s \mathbf{S}_i \Delta\mathbf{u}^{s+1},
\label{eq:strain}
\end{equation}
where $\mathbf{B}_{iq}^s$ is the strain--displacement matrix of element $i$ at integration point $q$.
The stress $\boldsymbol{\sigma}_{iq}$ and the history variables $\boldsymbol{\eta}_{iq}$ are updated by the history-dependent constitutive operator $\mathcal{H}$,
\begin{equation}
\left(\boldsymbol{\sigma}_{iq}^{s+1}, \boldsymbol{\eta}_{iq}^{s+1}\right) = \mathcal{H}\left(\boldsymbol{\sigma}_{iq}^{s}, \boldsymbol{\eta}_{iq}^{s}, \Delta\boldsymbol{\varepsilon}_{iq}^{s+1}, T_{iq}^{s}, \Delta t\right),
\label{eq:constitutive}
\end{equation}
where $T_{iq}^{s}$ is the temperature.
The thinning ratio is defined as
\begin{equation}
\theta_{iq}^{s} = 1 - \frac{h_{iq}^{s}}{h_{iq}^{0}} = 1 - \exp\left(\varepsilon_{33,iq}^{s}\right),
\label{eq:thinning}
\end{equation}
where $h_{iq}^{s}$ and $h_{iq}^{0}$ are the current and initial thicknesses, and $\varepsilon_{33,iq}^{s}$ is the logarithmic through-thickness strain.
Finally, the updated stresses are assembled into the internal force vector,
\begin{equation}
\mathbf{I}_{\mathrm{int}}^{s+1} = \sum_{i \in \mathcal{T}} \mathbf{S}_i^{\top} \sum_{q} w_q \left(\mathbf{B}_{iq}^{s+1}\right)^{\top} \boldsymbol{\sigma}_{iq}^{s+1} J_{iq}^{s+1},
\label{eq:assembly}
\end{equation}
where $w_q$ is the quadrature weight and $J_{iq}^{s+1}$ is the Jacobian at integration point $q$.
Eqs.~\eqref{eq:motion}--\eqref{eq:assembly} form one explicit update cycle.
The increment size is limited by the stability condition $\Delta t \leq \Delta t_{\mathrm{crit}}$, where $\Delta t_{\mathrm{crit}}$ is approximately the time taken by a stress wave to cross the smallest element.

Two properties of this cycle are noted.
First, nodes and elements carry different quantities and exchange information in a fixed order.
The strain increment of an element is computed from the displacement increments of its own nodes (Eq.~\eqref{eq:strain}), and the element stresses are then assembled into the forces of the same nodes (Eq.~\eqref{eq:assembly}).
Second, the two force terms in Eq.~\eqref{eq:motion} have different spatial characteristics and different roles.
In sheet forming, $\mathbf{P}_{\mathrm{ext}}$ is dominated by tool--blank contact forces.
These forces act only on the nodes in contact, change with the tool position, and depend mainly on the gap between each node and the tool surface and on the local tool normal.
In contrast, $\mathbf{I}_{\mathrm{int}}$ is defined over the whole blank and depends on its current deformation and stress state.
The contact force determines where and how the blank is loaded, while the internal force determines how the blank resists this loading, and the nodal motion is driven by the difference between the two.

These properties guide the model design.
First, nodes and elements are represented as separate entities that exchange information along their incidence relations.
Second, the contact input is processed separately from the blank state, since the two inputs have different spatial characteristics and roles.

\subsection{Learned update as a composition of explicit cycles}
\label{sec:problem_learned}

\begin{figure}[t]
\centering
\includegraphics[width=\textwidth,trim=50 188 25 132,clip]{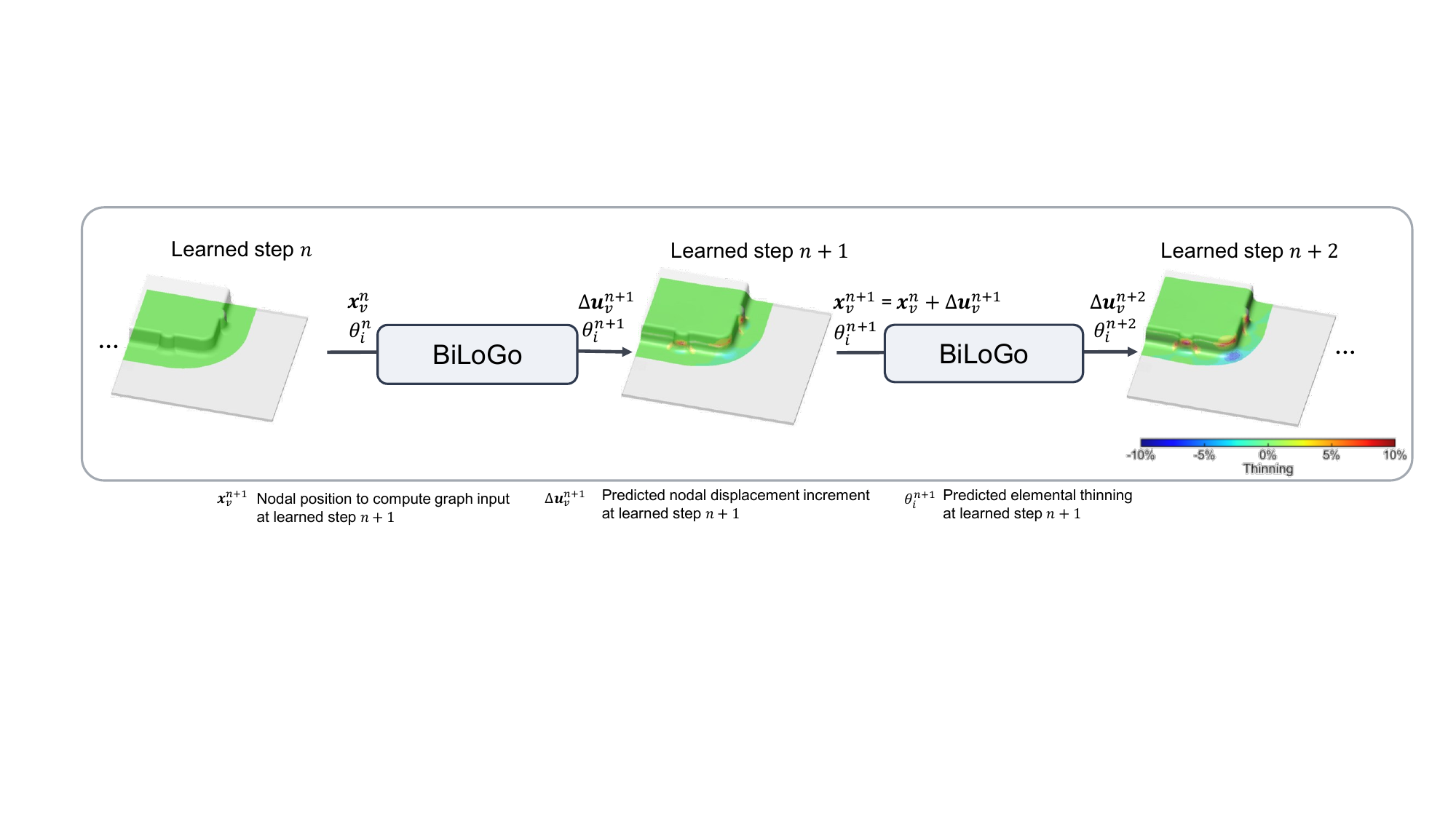}
\caption{Autoregressive rollout of BiLoGo, shown for a Corner+Rib simulation. At each learned step, the predicted displacement increment updates the nodal positions, the predicted thinning is used as the input of the next step, and the contact state is recomputed from the predicted geometry and the prescribed tool positions.}
\label{fig:autoregressive}
\end{figure}

The learned simulator advances the state over a learned step of size $\Delta T = t^{n+1} - t^{n}$, where the superscript $n$ denotes the learned step.
The blank state $\mathcal{S}^n$ describes the current configuration and deformation of the blank, including the nodal positions, the most recent nodal displacement increments and the element thinning.
The contact state $\mathcal{C}^n$ describes the distance and orientation of the blank relative to the tools, as described in detail in Section~\ref{sec:method_inputs}.
Given $\mathcal{S}^n$ and $\mathcal{C}^n$ at learned step $n$, the learned simulator $\mathcal{F}_{\phi}$ with parameters $\phi$ predicts
\begin{equation}
\left(\Delta\hat{\mathbf{u}}^{n+1}, \hat{\boldsymbol{\theta}}^{n+1}\right) = \mathcal{F}_{\phi}\left(\mathcal{S}^n, \mathcal{C}^n\right),
\label{eq:learned_update}
\end{equation}
where $\Delta\hat{\mathbf{u}}^{n+1}$ is the nodal displacement increment and $\hat{\boldsymbol{\theta}}^{n+1}$ is the element thinning at the next learned step.
Following \cite{Zhao2026CAttBiGNN}, the thinning of element $i$, $\theta_i^{n}$, is the element-level value of the integration-point thinning defined in Eq.~\eqref{eq:thinning}.
During rollout, the nodal positions are updated as $\mathbf{x}^{n+1} = \mathbf{x}^{n} + \Delta\hat{\mathbf{u}}^{n+1}$, the predicted thinning is used as the input of the next step, and the contact state is recomputed from the predicted geometry and the prescribed tool positions.
The input and output features are defined in Section~\ref{sec:graph}.
The autoregressive rollout is illustrated in Fig.~\ref{fig:autoregressive}.

One learned step covers $m = \Delta T / \Delta t$ explicit cycles, so $\mathcal{F}_{\phi}$ approximates the composition of $m$ cycles of Eqs.~\eqref{eq:motion}--\eqref{eq:assembly}.
The strain increment of element $i$ depends only on the nodes in $\mathcal{N}(i)$ (Eq.~\eqref{eq:strain}), and the internal force at node $v$ depends only on the elements in $\mathcal{M}(v)$ (Eq.~\eqref{eq:assembly}).
One explicit cycle therefore couples each node only with the nodes of its neighbouring elements, which is referred to here as one element ring.
After $m$ cycles, the state of a node can depend on nodes up to $m$ element rings away.
This also applies to the contact force.
Although it acts only on the nodes in contact, as described in Section~\ref{sec:problem_fe}, its effect is transferred to other regions of the blank through the internal force over these cycles.
Since one learned step usually covers many explicit cycles, information from distant regions of the blank can affect the update of a node within one learned step.
Therefore, in addition to local node--element exchange, the learned simulator needs a mechanism that exchanges both the contact state and the blank state across the whole blank within one learned step.
On the Corner dataset, adding graph downsampling and upsampling to a recurrent single-level GNN reduced the final mean positional Euclidean error from 1.21~mm to 0.52~mm \cite{Zhao2025BatteryBox}, which indicates that long-range propagation is needed for this problem.

\subsection{Local element-level response}
\label{sec:problem_local}

Section~\ref{sec:problem_fe} shows that, within one explicit cycle, the strain increment of an element depends only on its own nodes.
A similar relation holds for the total deformation of an element over a learned step.
Let $a$, $b$ and $c$ denote the three nodes of element $i$.
The edge matrices of the element in the reference and current configurations are
\begin{equation}
\mathbf{D}_i^{0} = \left[\mathbf{X}_b - \mathbf{X}_a,\ \mathbf{X}_c - \mathbf{X}_a\right],
\qquad
\mathbf{D}_i^{n} = \left[\mathbf{x}_b^{n} - \mathbf{x}_a^{n},\ \mathbf{x}_c^{n} - \mathbf{x}_a^{n}\right],
\label{eq:edge_matrices}
\end{equation}
where $\mathbf{X}_a$, $\mathbf{X}_b$ and $\mathbf{X}_c$ are the reference nodal positions expressed in the in-plane coordinate system of the reference element, so that $\mathbf{D}_i^{0} \in \mathbb{R}^{2\times2}$, and $\mathbf{D}_i^{n} \in \mathbb{R}^{3\times2}$.
For a linear triangular element, the deformation gradient of the mid-surface is constant within the element and is given by
\begin{equation}
\mathbf{F}_i^{n} = \mathbf{D}_i^{n} \left(\mathbf{D}_i^{0}\right)^{-1},
\qquad
\mathbf{C}_i^{n} = \left(\mathbf{F}_i^{n}\right)^{\top} \mathbf{F}_i^{n},
\label{eq:deformation_gradient}
\end{equation}
where $\mathbf{C}_i^{n} \in \mathbb{R}^{2\times2}$ is the in-plane right Cauchy--Green tensor, which describes the in-plane stretching of the element.
Eqs.~\eqref{eq:edge_matrices} and \eqref{eq:deformation_gradient} show that the in-plane deformation of element $i$ depends only on the positions of its three nodes, even though the motion of these nodes depends on distant regions, as shown in Section~\ref{sec:problem_learned}.
The thinning is then obtained from the plane-stress constitutive update (Eq.~\eqref{eq:constitutive}), and, for plastic deformation, it is additionally constrained by the plastic incompressibility condition \cite{Zhao2026CAttBiGNN}.

The deformation gradient $\mathbf{F}_i^{n}$ describes the membrane deformation of the mid-surface, but not the change in curvature.
For the Corner+Rib simulations, the thinning estimated from the change in element area, $1 - A_i^{0}/A_i^{n}$, where $A_i^{0}$ and $A_i^{n}$ are the reference and current element areas, has a correlation coefficient of 0.9993 with the element thinning exported by the FE solver at the final state.
This indicates that the thinning is mainly determined by membrane deformation in the present simulations.
For processes in which bending has a larger effect on the thinning, this may not hold, as discussed in Section~\ref{sec:results_limitations}.

In summary, the nodal motion depends on information from distant regions, while the element deformation and thinning are mainly determined by the kinematics of the element nodes.
Since the element deformation is given by the spatial gradient of the nodal displacement within each element, small errors in the predicted nodal positions can lead to large errors in the element deformation.
Eqs.~\eqref{eq:edge_matrices} and \eqref{eq:deformation_gradient} are therefore not used to compute the thinning directly from the predicted nodal positions.
Instead, they indicate that the element prediction mainly requires the current kinematics of the element nodes and the motion of these nodes over the learned step.
Therefore, the long-range information is mainly needed for the nodal motion, while the element prediction can be obtained from the current kinematics of the element nodes together with the nodal information passed to the element through the node--element message passing.

\FloatBarrier
\section{Dataset generation}
\label{sec:dataset}

Three forming scenarios are used in this study.
The Dome and Corner datasets were introduced in \cite{Zhao2026CAttBiGNN} and represent cold forming and hot stamping with variations in the global tool geometry.
The Corner+Rib dataset is introduced in this work.
It uses a fixed Corner geometry with a pair of reinforcing ribs, whose position and size vary between simulations, so that the deformation and thinning change mainly in local regions.
The three datasets therefore represent increasingly localised responses: the thinning is distributed over the dome in the Dome dataset, concentrated near the corner in the Corner dataset, and changes mainly near the ribs in the Corner+Rib dataset.
Section~\ref{sec:dataset_workflow} describes the common simulation workflow, and Sections~\ref{sec:dataset_dome}--\ref{sec:dataset_rib} describe the three datasets.

\FloatBarrier
\subsection{Simulation workflow}
\label{sec:dataset_workflow}

In sheet forming, the blank is clamped between the die and the blank holder, and the relative motion between the punch and the die draws the blank into the die cavity.
The blank-holder force controls the flow of material from the flange into the die.
The distance travelled by the moving tool is referred to as the stroke.
In cold forming, the blank is formed at room temperature.
In hot stamping, the blank is heated before forming and is formed and quenched by cold tools, so that its temperature decreases during the process \cite{Shao2017HFQFormability,Attar2021Feasibility}.

All datasets were generated with the automated workflow developed in \cite{Attar2021Feasibility} and used in \cite{Zhao2026RUGNN,Zhao2026CAttBiGNN}.
For each dataset, combinations of the design variables were sampled by Latin Hypercube Sampling (LHS) \cite{McKay1979LHS} within the prescribed ranges.
For each sampled design, the tool geometries were created in SolidWorks, meshed in HyperMesh and simulated in PAM-STAMP.
The tools were modelled as rigid bodies and the blank as a deformable body.
The blank was discretised with three-node linear triangular shell elements with five integration points through the thickness.
Only one quarter of each component was simulated, with symmetry boundary conditions on the two symmetry planes, and the same blank geometry was used for all simulations within a dataset.

For each simulation, 11 states were saved over the complete forming process, from the undeformed blank to the final formed state.
These states are referred to as stages 1 to 11 in Figs.~\ref{fig:dome_dataset}--\ref{fig:rib_dataset}.
The learned simulator therefore performs a 10-step autoregressive rollout, in which each learned step covers many solver time steps.
The solver time step is computed automatically by PAM-STAMP, with a mean value of approximately 3.45~$\mu$s, and no mass is added to the blank by mass scaling.

\FloatBarrier
\subsection{Dome dataset}
\label{sec:dataset_dome}

\begin{figure}[t]
\centering
\includegraphics[width=\textwidth,trim=71 65 52 71,clip]{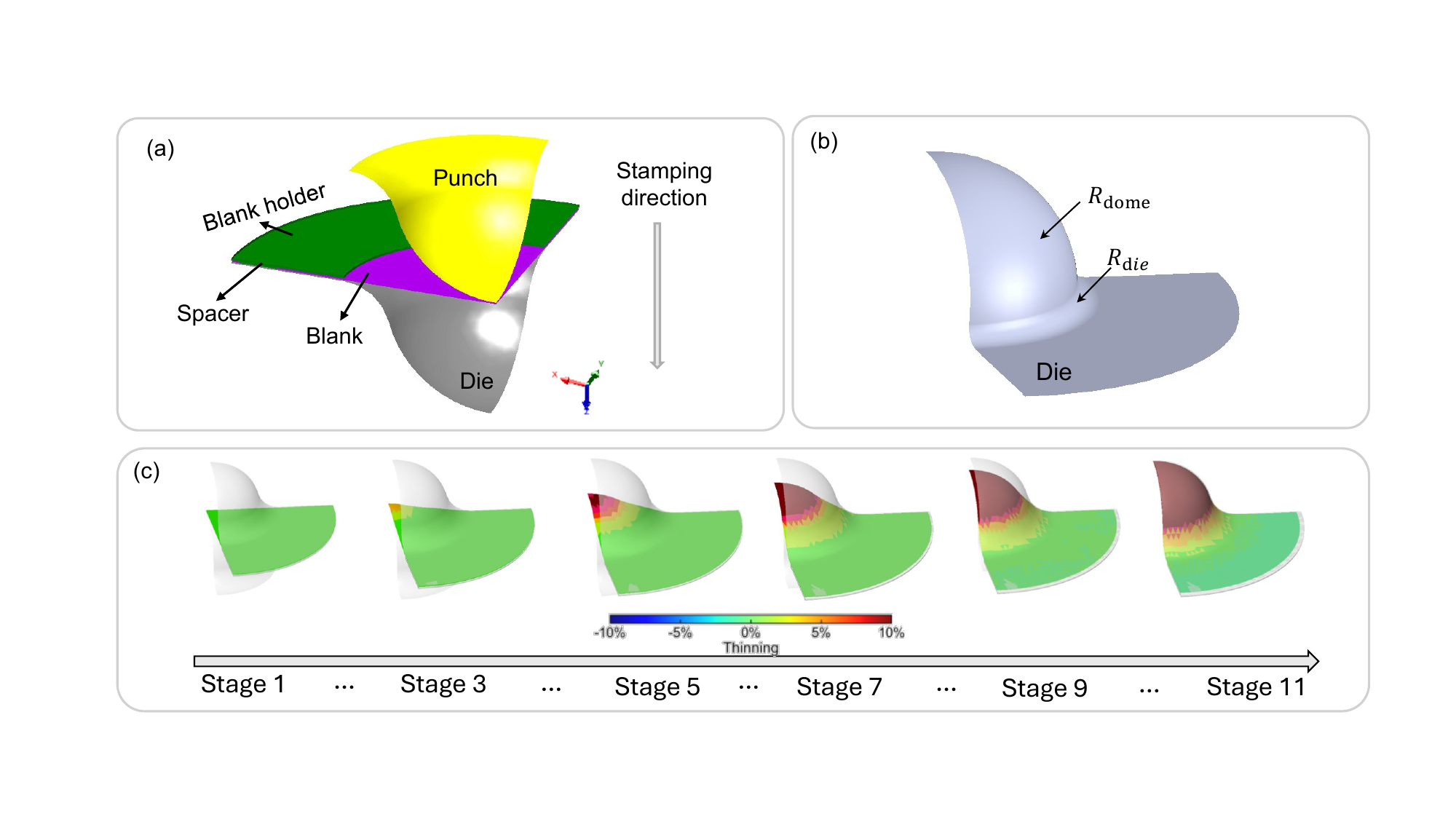}
\caption{Dome dataset. (a) Tool set-up of the quarter model. (b) Die geometry and design variables $R_{\mathrm{dome}}$ and $R_{\mathrm{die}}$. (c) Thinning distribution of one simulation over the 11 saved stages.}
\label{fig:dome_dataset}
\end{figure}

The Dome dataset represents the cold forming of a double-curvature component, which leads to a more complex stress and deformation state than a single-curvature geometry \cite{Zhao2026CAttBiGNN}.
The tool set-up, the design variables and the forming process of one simulation are shown in Fig.~\ref{fig:dome_dataset}.
The die, punch, blank holder and spacers are rigid, and the die is fully constrained.
The punch moves vertically at a constant speed of 500~mm/s.
A blank-holder force of 200~kN is applied to control the material flow in the flange, and spacers with a thickness of 2.2~mm are placed between the die and the blank holder.
The blank is made of cold-worked Al2008, described by an isotropic material model from the PAM-STAMP material database, with an initial thickness of 2~mm and a friction coefficient of 0.12.
The die geometry is defined by the dome radius $R_{\mathrm{dome}}$ and the die radius $R_{\mathrm{die}}$, with $60~\mathrm{mm} \leq R_{\mathrm{dome}} \leq 120~\mathrm{mm}$ and $10~\mathrm{mm} \leq R_{\mathrm{die}} \leq 20~\mathrm{mm}$.
The punch geometry is obtained by a 2~mm offset of the die.
The blank is discretised with 611 nodes and 1,120 triangular elements.
The dataset contains 50 training, 25 validation and 25 test simulations, which were sampled independently to avoid overlap between the subsets.

\FloatBarrier
\subsection{Corner dataset}
\label{sec:dataset_corner}

\begin{figure}[t]
\centering
\includegraphics[width=\textwidth,trim=71 68 52 71,clip]{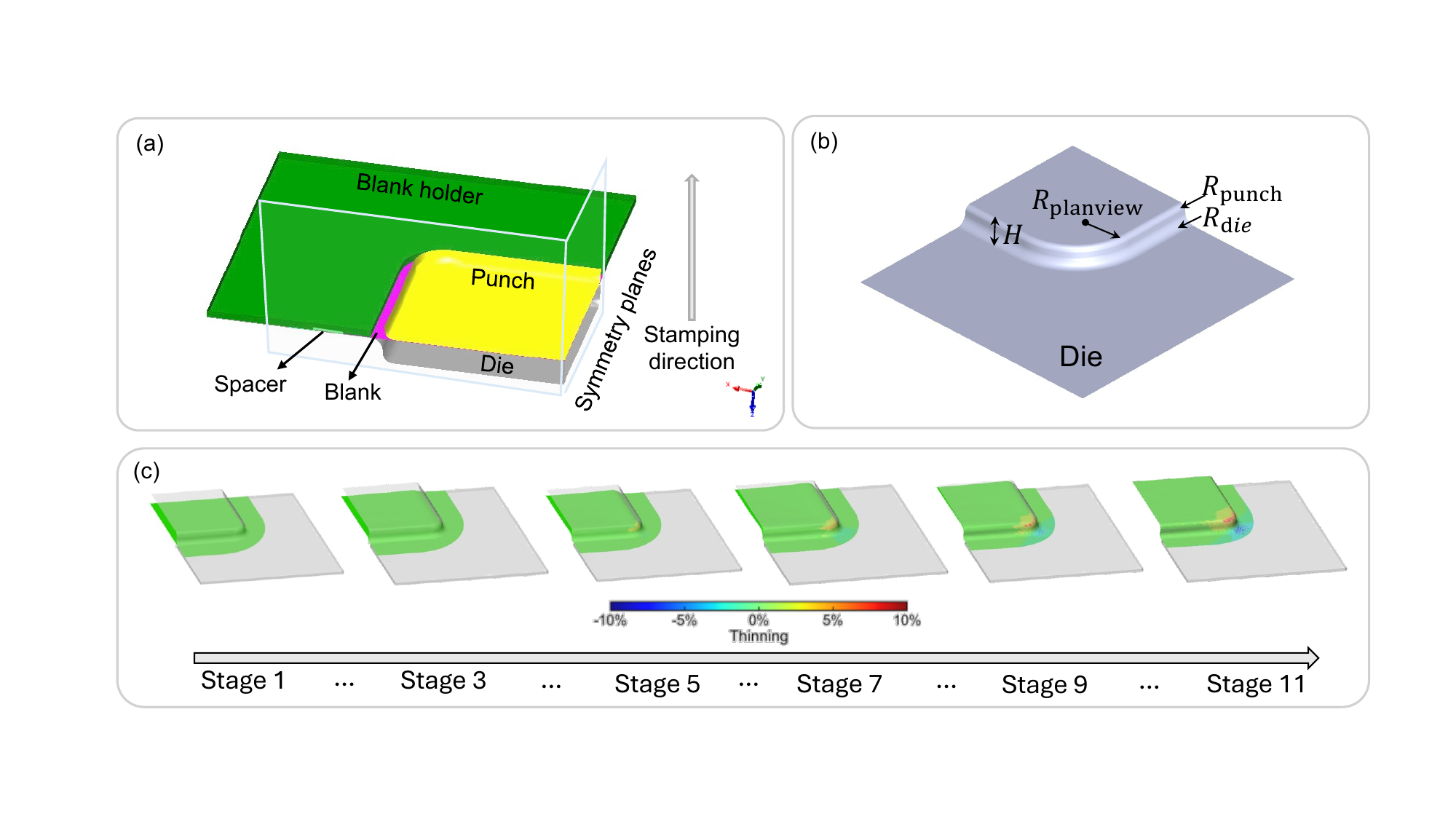}
\caption{Corner dataset. (a) Tool set-up of the quarter model. (b) Die geometry and design variables $R_{\mathrm{die}}$, $R_{\mathrm{punch}}$, $R_{\mathrm{planview}}$ and $H$. (c) Thinning distribution of one simulation over the 11 saved stages.}
\label{fig:corner_dataset}
\end{figure}

The Corner dataset represents the hot stamping of a deep-drawn corner, which is a typical feature of battery box components \cite{Zhao2025BatteryBox,Zhao2026CAttBiGNN}.
The tool set-up, the design variables and the forming process of one simulation are shown in Fig.~\ref{fig:corner_dataset}.
The die geometry is defined by four design variables: the die radius $R_{\mathrm{die}}$, the punch radius $R_{\mathrm{punch}}$, the plan-view radius $R_{\mathrm{planview}}$ and the height $H$, with $25~\mathrm{mm} \leq R_{\mathrm{die}} \leq 40~\mathrm{mm}$, $20~\mathrm{mm} \leq R_{\mathrm{punch}} \leq 27.3~\mathrm{mm}$, $70~\mathrm{mm} \leq R_{\mathrm{planview}} \leq 250~\mathrm{mm}$ and $60~\mathrm{mm} \leq H \leq 120~\mathrm{mm}$.
The punch geometry is obtained by a 2.3~mm offset of the die.
The blank is discretised with 4,130 nodes and 8,008 triangular elements.
The blank is made of AA6082, described by an elastic--viscoplastic material model for hot-stamping conditions \cite{Mohamed2012AA6082,Attar2021Feasibility,Zhao2026RUGNN,Zhao2026CAttBiGNN}.
The blank is heated to 500~$^{\circ}$C, and the tools are kept at 25~$^{\circ}$C.
The stamping speed is 500~mm/s, the spacer thickness is 2.3~mm, the blank thickness is 2~mm, the friction coefficient is 0.1, and a blank-holder force of 200~kN is applied in the direction opposite to the stamping motion.
The punch is fully constrained, and the die moves in the negative $z$-direction with a stroke determined by the height of the die.
The dataset contains 100 training, 25 validation and 25 test simulations.

\FloatBarrier
\subsection{Corner+Rib dataset}
\label{sec:dataset_rib}

\begin{figure}[t]
\centering
\includegraphics[width=\textwidth,trim=58 61 63 82,clip]{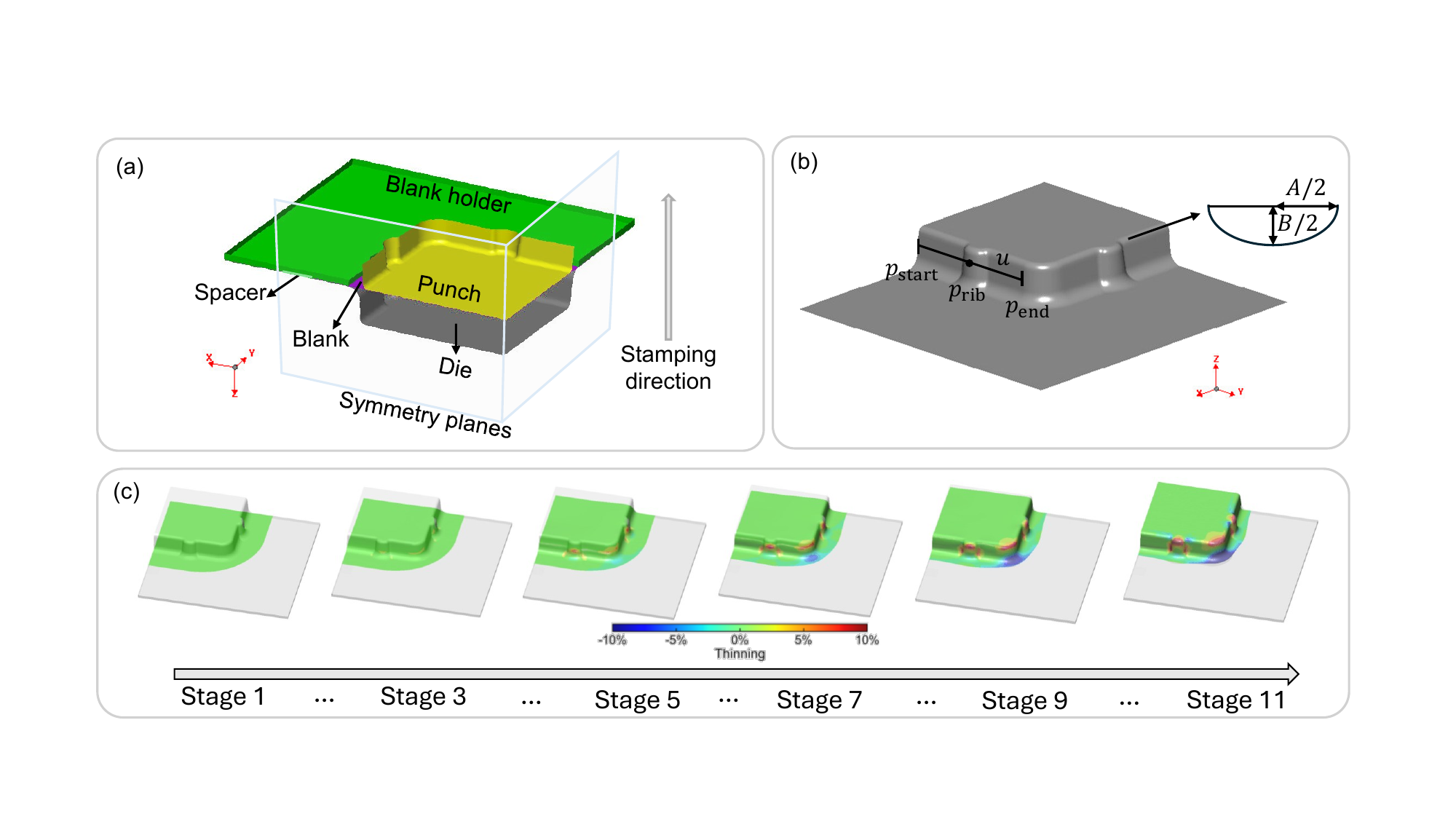}
\caption{Corner+Rib dataset. (a) Tool set-up of the quarter model. (b) Die geometry of the quarter model with the rib parameters. The rib centre $\mathbf{p}_{\mathrm{rib}}$ is located on the straight top edge between $\mathbf{p}_{\mathrm{start}}$ at the symmetry plane and $\mathbf{p}_{\mathrm{end}}$ at the corner fillet by the position parameter $u$ (Eq.~\eqref{eq:rib_position}), and $A/2$ and $B/2$ are the semi-axes of the half-elliptical rib along and normal to the side wall. (c) Thinning distribution of one simulation over the 11 saved stages.}
\label{fig:rib_dataset}
\end{figure}

The Corner+Rib dataset is used to study the effect of local geometric features.
It uses the same process conditions, material, blank thickness and friction coefficient as the Corner dataset.
The global geometry of the Corner component is fixed for all simulations, with each design variable set to the centre of its range in the Corner dataset: $R_{\mathrm{die}} = 32.5$~mm, $R_{\mathrm{punch}} = 23.65$~mm, $R_{\mathrm{planview}} = 160$~mm and $H = 90$~mm.
A pair of reinforcing ribs is added to the two side walls next to the corner, and only the position and size of the ribs vary between simulations.
The tool set-up, the rib parameters and the forming process of one simulation are shown in Fig.~\ref{fig:rib_dataset}.

In the top view, each rib has the shape of half an ellipse.
The straight edge of the half ellipse lies on the side wall, and the elliptical arc extends outwards from the wall.
The rib is extruded from the top face of the Corner geometry to the local bottom face of the side wall, so its height follows the height of the component.
The rib geometry is defined by three parameters.
The position parameter $u$ defines the centre of the rib along the straight top edge of the side wall,
\begin{equation}
\mathbf{p}_{\mathrm{rib}} = \mathbf{p}_{\mathrm{start}} + \frac{u}{100}\left(\mathbf{p}_{\mathrm{end}} - \mathbf{p}_{\mathrm{start}}\right),
\label{eq:rib_position}
\end{equation}
where $\mathbf{p}_{\mathrm{start}}$ is the end of the straight edge at the symmetry plane, $\mathbf{p}_{\mathrm{end}}$ is the end of the straight edge at the corner fillet, and $u$ is given in per cent.
The semi-axis $A/2$ is the half-length of the rib along the side wall, so the full length of the rib along the wall is $A$.
The semi-axis $B/2$ is the distance by which the rib extends outwards from the side wall.
The ranges of the three parameters are listed in Table~\ref{tab:rib_parameters}.
The position is limited to the central half of the edge to avoid interference between the rib and the corner or the symmetry plane.
The two ribs use the same values of $u$, $A/2$ and $B/2$.
A draft angle of 6$^{\circ}$ is applied to the side faces of the ribs, and a fillet with a radius of 15~mm is applied at the root of the ribs.
These two values are fixed for all simulations.
The three parameters were sampled as continuous variables by LHS within the ranges in Table~\ref{tab:rib_parameters}.

\begin{table}[t]
\centering
\caption{Design variables of the Corner+Rib dataset.}
\label{tab:rib_parameters}
\small
\begin{tabular}{llp{6.2cm}}
\hline
Parameter & Range & Description \\
\hline
$u$ & 25--75\% & Position of the rib centre along the top edge \\
$A/2$ & 35--50~mm & Half-length of the rib along the side wall \\
$B/2$ & 22.5--30~mm & Outward extent of the rib from the side wall \\
\hline
\end{tabular}
\end{table}

The blank is a quarter blank of $670 \times 670$~mm, discretised with 16,049 nodes and 31,595 triangular elements, with a nominal element size of approximately 5~mm.
The rigid tool surfaces were meshed with minimum and maximum element sizes of 1~mm and 5~mm, respectively.
This mesh contains approximately four times as many elements as the Corner mesh and 28 times as many as the Dome mesh, so that the deformation and thinning near the ribs can be resolved.
The dataset contains 100 training, 50 validation and 50 test simulations.

\FloatBarrier
\section{Graph construction and input--output features}
\label{sec:graph}

This section describes how the FE results in Section~\ref{sec:dataset} are converted into graphs for the learned simulator, and defines its input and output features.
The design follows the properties of the FE update identified in Section~\ref{sec:problem_fe}: nodes and elements are represented as separate entities, and the blank state and the tool contact are treated as separate inputs.

\FloatBarrier
\subsection{Node--element bipartite graph}
\label{sec:graph_construction}

\begin{figure}[t]
\centering
\includegraphics[width=\textwidth,trim=65 77 85 59,clip]{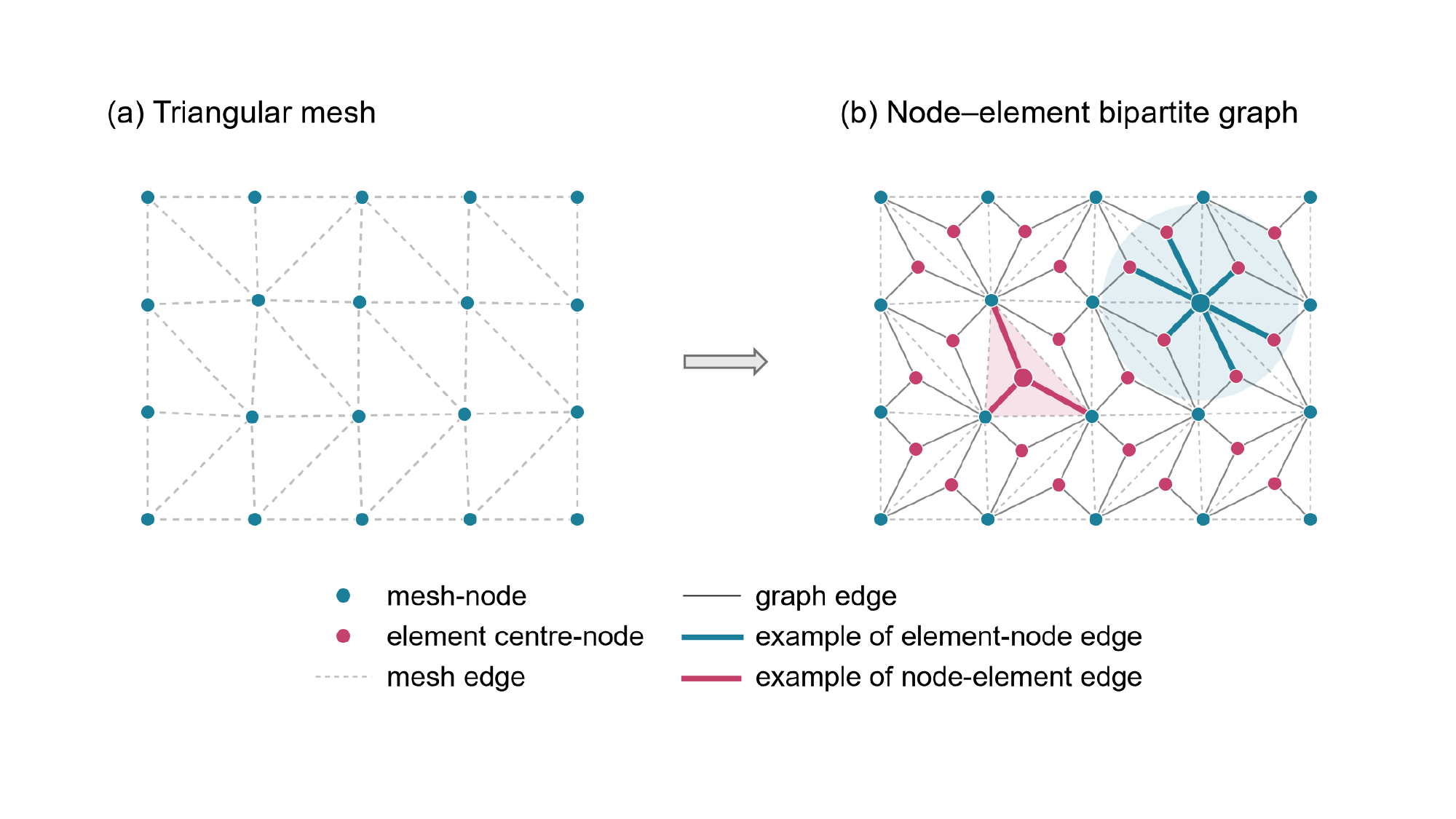}
\caption{Construction of the node--element bipartite graph. (a) Triangular mesh. (b) Bipartite graph with mesh-nodes and element centre-nodes. Each element centre-node is connected to the three nodes of its element, and information is passed in both directions along each edge. The highlighted edges show the neighbourhood used to update an element from its three nodes (magenta, Eq.~\eqref{eq:mp_element}) and the neighbourhood used to update a node from its neighbouring elements (teal, Eq.~\eqref{eq:mp_node}).}
\label{fig:bipartite_graph}
\end{figure}

Following \cite{Zhao2026CAttBiGNN}, the FE mesh at each saved state is represented as a node--element bipartite graph.
Each mesh node and each element is a graph entity.
Each element is represented by a graph node located at the element centre, which is referred to as an element centre-node, while the mesh nodes are referred to as mesh-nodes.
Each element is connected to its three nodes by directed node-to-element and element-to-node edges, which correspond to the two directions of information transfer in the FE update: the element strain is computed from its nodes (Eq.~\eqref{eq:strain}), and the element stress is assembled into the forces of the same nodes (Eq.~\eqref{eq:assembly}).
The original mesh edges between nodes are not used, since in the FE update two nodes interact only through the elements they share.
The construction of the graph is shown in Fig.~\ref{fig:bipartite_graph}.
The edges define which entities exchange information and carry no features.

For the Corner+Rib mesh, for example, the graph contains 16,049 node entities, 31,595 element entities and 94,785 edges in each direction.

\FloatBarrier
\subsection{Input features}
\label{sec:graph_inputs}

The input features at learned step $n$ are listed in Table~\ref{tab:inputs}.
They are divided into a state stream and a contact stream.
The state stream describes the current configuration and deformation of the blank.
For each node, it contains the current position $\mathbf{x}_v^{n}$, the most recent displacement increment $\Delta\mathbf{u}_v^{n}$, the boundary indicator $\mathbf{b}_v$ and the global input $\mathbf{l}$.
For each element, it contains the centre position $\mathbf{x}_i^{n}$, the mean displacement increment of its nodes $\overline{\Delta\mathbf{u}}_i^{n}$, the current thinning $\theta_i^{n}$ and the global input $\mathbf{l}$.
The displacement increment $\Delta\mathbf{u}_v^{n}$ describes the most recent nodal motion, in the same way as the nodal velocity in the central-difference update (Eq.~\eqref{eq:central_difference}), and the current thinning provides the deformation history of each element.
Following \cite{Zhao2026CAttBiGNN}, the boundary indicator $\mathbf{b}_v = (b_{v,x}, b_{v,y}, b_{v,z}) \in \{0,1\}^{3}$ marks the constrained degrees of freedom of node $v$, where each component equals 1 if the displacement in the corresponding direction is constrained.
The global input $\mathbf{l} = [\Delta T, L]$ contains the tool stroke $L$ and the time interval $\Delta T = L/(N v_{\mathrm{s}})$ between two learned steps, where $N = 10$ is the number of learned steps and $v_{\mathrm{s}}$ is the stamping speed.
The same global input is used at every learned step of a simulation.
On the Corner+Rib dataset, the tool stroke is the same for all simulations, so $\mathbf{l}$ provides no variation across simulations.
In the standardisation of the BiLoGo inputs, its zero standard deviation is therefore replaced by 1.

\begin{table}[t]
\centering
\caption{Input and output features of BiLoGo at learned step $n$.}
\label{tab:inputs}
\small
\begin{tabular}{lll}
\hline
Stream & Entity & Features \\
\hline
State & Node $v$ & $\mathbf{s}_v^{n} = [\mathbf{x}_v^{n},\ \Delta\mathbf{u}_v^{n},\ \mathbf{b}_v,\ \mathbf{l}]$ \\
State & Element $i$ & $\mathbf{s}_i^{n} = [\mathbf{x}_i^{n},\ \overline{\Delta\mathbf{u}}_i^{n},\ \theta_i^{n},\ \mathbf{l}]$ \\
Contact & Node $v$ & $\mathbf{c}_v^{n}$: proximity $c(d)$ and surface normal $\mathbf{n}$ for each tool, $\mathbf{l}$ \\
Contact & Element $i$ & $\mathbf{c}_i^{n}$: mean of the node contact inputs over $v \in \mathcal{N}(i)$, $\mathbf{l}$ \\
\hline
Output & Node $v$ & displacement increment $\Delta\mathbf{u}_v^{n+1}$ \\
Output & Element $i$ & thinning $\theta_i^{n+1}$ \\
\hline
\end{tabular}
\end{table}

The contact stream describes the relation between the blank and the tools.
For each node and each of the three tools (die, punch and blank holder), the shortest distance $d$ from the node to the tool surface mesh and the normal $\mathbf{n}$ of the closest tool triangle are computed.
The distance is converted into a proximity value $c(d)$, which increases as the node approaches the tool.
The contact input of node $v$ contains the proximity value $c(d)$ and the surface normal $\mathbf{n}$ for each of the three tools, together with the global input $\mathbf{l}$.
The contact input of each element is the mean of the contact inputs of its nodes, together with the global input $\mathbf{l}$.
These quantities describe the gap and the tool normal, on which the contact forces in $\mathbf{P}_{\mathrm{ext}}$ mainly depend, as described in Section~\ref{sec:problem_fe}.
During rollout, the distances and normals are recomputed at every learned step from the predicted geometry and the prescribed tool positions, so that training and rollout use the same contact definition.

For the Dome and Corner datasets, the proximity is computed as the inverse of the non-negative shortest distance, $c(d) = 1/\max(d, \epsilon)$, where $\epsilon$ is a small constant that avoids division by zero.
This follows the contact computation in \cite{Zhao2026CAttBiGNN}, so that the results can be compared with the graph-based models and baselines reported in that work.

For the Corner+Rib dataset, the inverse distance may lead to unstable training.
The ribs create narrow regions in which some nodes lie very close to the tool surfaces.
For these nodes, the inverse distance can become very large, and the values remain large compared with those of the other nodes after standardisation.
This can make the gradient-based optimisation unstable.
In addition, since the inverse distance changes rapidly as $d$ approaches zero, small errors in the predicted positions near the tools can cause large changes in the contact input during rollout.
Therefore, a Cauchy proximity is used for all models on the Corner+Rib dataset,
\begin{equation}
c(d) = \frac{1}{1 + \left(d/s\right)^{2}},
\label{eq:cauchy_proximity}
\end{equation}
where $s = 2$~mm is a length scale, which is set equal to the blank thickness, so that the proximity equals 0.5 when a node is one blank thickness away from the tool surface.
It is bounded between 0 and 1, so the nodes closest to the tools no longer produce extreme values.
It also decreases smoothly with distance and still gives larger values to nodes closer to the tool, so the information on which nodes are near the tools is kept.

The two streams are kept separate because they have different spatial characteristics and different roles.
The information in the contact stream that drives the deformation is concentrated near the contact regions and changes with the tool position, while the state stream varies over the whole blank.
Keeping them separate allows the model to process and compress each stream according to its own characteristics, and to compute the response to the contact condition with respect to the current state of the blank, as described in Section~\ref{sec:method}.

\FloatBarrier
\subsection{Output features and normalisation}
\label{sec:graph_outputs}

The output of each node is the displacement increment $\Delta\mathbf{u}_v^{n+1}$ over the next learned step, and the output of each element is the thinning $\theta_i^{n+1}$ at the next learned step.
The displacement is predicted as an increment because the nodal positions are updated as $\mathbf{x}^{n+1} = \mathbf{x}^{n} + \Delta\hat{\mathbf{u}}^{n+1}$ during rollout, as described in Section~\ref{sec:problem_learned}.
Following \cite{Zhao2026CAttBiGNN}, the thinning is predicted as the absolute value at the next step, and the predicted value is used as the thinning input of the following step.
Both outputs are predicted on their own entities, so the element thinning does not need to be interpolated to the nodes.

All input and output features are standardised with the mean and standard deviation computed on the training set.
The only exception is the Cauchy proximity, which is not standardised, since it already lies between 0 and 1.
The predicted outputs are converted back to physical values with the same statistics before they are used to update the state.

\FloatBarrier
\section{BiLoGo model development}
\label{sec:method}

\begin{figure}[t]
\centering
\includegraphics[width=\textwidth,trim=20 17 16 54,clip]{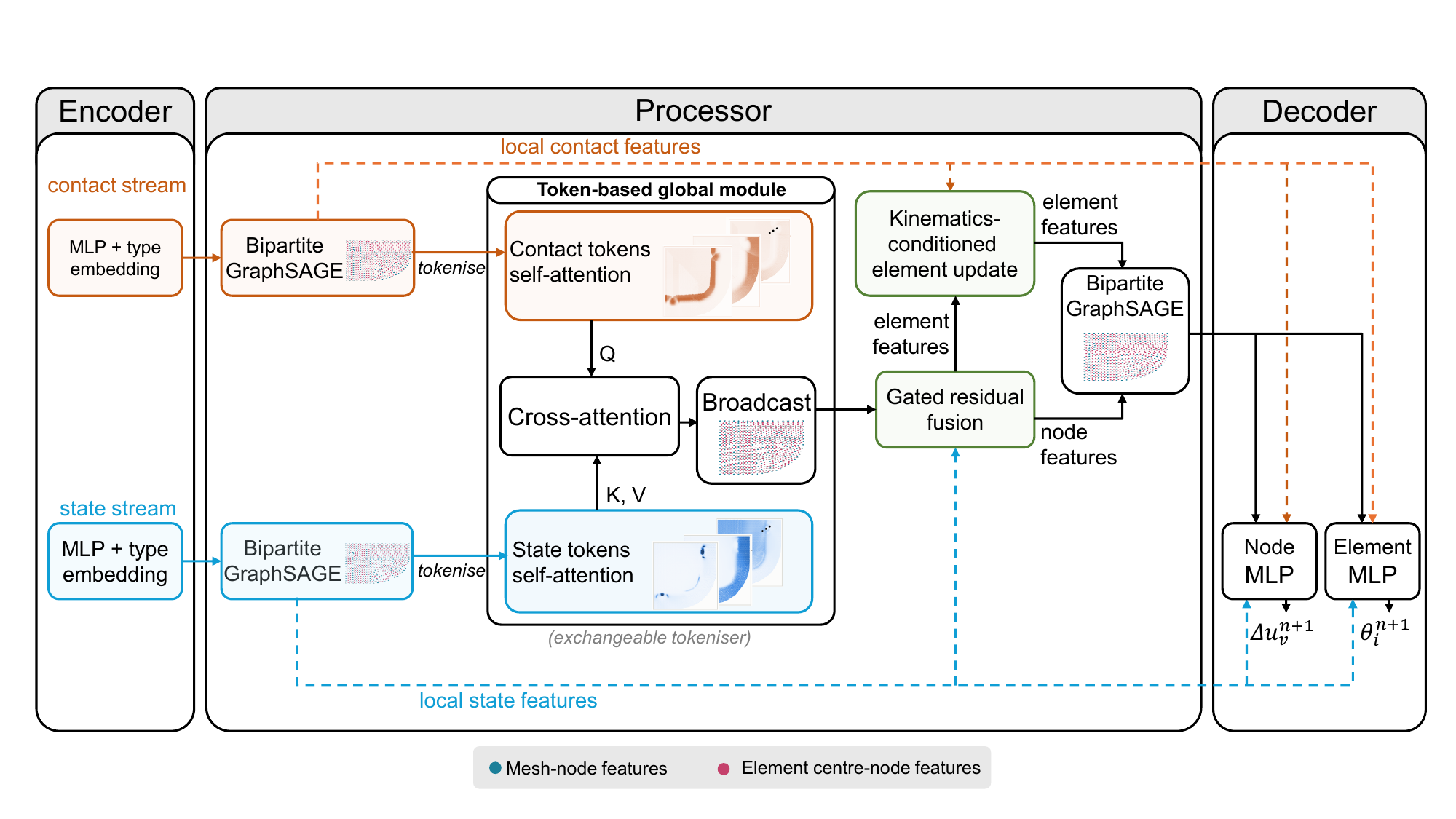}
\caption{Architecture of BiLoGo. The contact and state streams are encoded separately and processed by the bipartite GraphSAGE layers of the bidirectional node--element GraphSAGE model. In the global module, each stream is compressed into tokens and processed by self-attention, and the contact tokens query the state tokens through cross-attention. The result is broadcast back to the nodes and elements, combined with the local state features by the gated residual fusion, and used in the kinematics-conditioned element update. After the bipartite GraphSAGE layers that follow the global module, the node and element decoders predict the displacement increment and the thinning. Dashed lines show the local features passed directly to later components. The gated residual fusion and the kinematics-conditioned element update, which form the fusion and element update module, are highlighted in green. The tokeniser is either the physical-slice tokeniser (BiLoGo-T) or the spatial supernode tokeniser (BiLoGo-U).}
\label{fig:architecture}
\end{figure}

The architecture of the Bipartite Local--Global (BiLoGo) learned simulator is shown in Fig.~\ref{fig:architecture}.
At each learned step, BiLoGo takes the blank state and the tool contact as two separate input streams on the node--element bipartite graph described in Section~\ref{sec:graph}, and predicts the nodal displacement increment $\Delta\hat{\mathbf{u}}^{n+1}$ and the element thinning $\hat{\boldsymbol{\theta}}^{n+1}$ (Eq.~\eqref{eq:learned_update}).
BiLoGo consists of three modules.
First, a bidirectional node--element GraphSAGE model performs local message passing before and after the global module.
Second, a dual-stream Transformer-based neural operator uses a tokeniser to compress each stream into tokens, performs the global interaction and maps the result back to the nodes and elements.
Third, a fusion and element update module combines the global update with the local features and updates the element features conditioned on the kinematics of the element nodes.
The tokeniser in the second module can be replaced.
BiLoGo-T uses learned physical slices, and BiLoGo-U uses spatially distributed supernodes.
Apart from the tokeniser, all components are the same in all variants of BiLoGo.

The remainder of this section is organised as follows.
Section~\ref{sec:method_compression} analyses the information loss in token-based global propagation and derives the requirements for the model design.
Sections~\ref{sec:method_inputs}--\ref{sec:method_decoder} then describe the components in the order in which information passes through the model: the input encoding in Section~\ref{sec:method_inputs}, the local message passing before the global module in Section~\ref{sec:method_mp}, the global module in Section~\ref{sec:method_global}, the gated fusion in Section~\ref{sec:method_fusion}, the kinematics-conditioned element update in Section~\ref{sec:method_kinematic}, and the local message passing after the global module and the output decoders in Section~\ref{sec:method_decoder}.
The first module is therefore described in Sections~\ref{sec:method_mp} and \ref{sec:method_decoder}, the second in Section~\ref{sec:method_global}, and the third in Sections~\ref{sec:method_fusion} and \ref{sec:method_kinematic}.

\FloatBarrier
\subsection{Information loss in token-based global propagation}
\label{sec:method_compression}

The global module required in Section~\ref{sec:problem_learned} compresses the node and element features into a small number of tokens, performs global interaction between the tokens, and maps the result back to the mesh.
This subsection describes how this compression affects the local information, and derives the requirements for the model design.
Let $\mathbf{h}_e \in \mathbb{R}^{d}$ denote the latent feature of entity $e$, where $e$ is either a node or an element, and let $N_e = |\mathcal{V}| + |\mathcal{T}|$ be the number of entities.

In slice-based compression \cite{Wu2024Transolver}, each entity is assigned to $K$ tokens with weights $a_{ek} \geq 0$, $\sum_{k=1}^{K} a_{ek} = 1$, which are computed from the entity features.
The tokens and the recovered global features are
\begin{equation}
\mathbf{z}_k = \frac{\sum_{e} a_{ek} \mathbf{h}_e}{\sum_{e} a_{ek}},
\qquad
\mathbf{g}_e = \sum_{k=1}^{K} a_{ek} \tilde{\mathbf{z}}_k,
\label{eq:slice}
\end{equation}
where $\tilde{\mathbf{z}}_k$ are the tokens after global interaction.
In matrix form, the recovered features are $\mathbf{G} = \mathbf{A}\tilde{\mathbf{Z}}$, where $\mathbf{A} \in \mathbb{R}^{N_e \times K}$ collects the assignment weights.
The rank of $\mathbf{G}$ is therefore at most $K$, and the recovered feature of every entity is a convex combination of the same $K$ vectors.
This bound applies to the global update, while the subsequent layers also use the local features, so the final prediction is not limited to rank $K$.
Three consequences follow.
First, entities with similar assignment weights receive similar global features, even if they are far apart.
Neighbouring entities, which tend to have similar features, also tend to receive similar global features.
Second, since the assignment is based on features rather than position, a feature that changes sharply over short distances, such as the contact input near the edges of the contact regions, can be averaged with smoother features within the same token.
Third, a field that varies smoothly over the blank may be represented with a small $K$, whereas a response concentrated in several small regions may require the tokens to separate these regions, which is not guaranteed by feature-based assignment.

In supernode-based compression \cite{Alkin2024UPT}, each token $k$ is associated with a position $\mathbf{p}_k$ and aggregates only the entities within a radius $r$ of $\mathbf{p}_k$.
The recovered feature of each entity is computed by attention, with a query based on the position of the entity.
The compression is therefore spatially local, and its resolution is determined by the number and spacing of the supernodes rather than by the feature-based assignment.

The three consequences above lead to the following requirements for the model design.
First, if the local feature of an entity is replaced by its recovered global feature, the part of the local feature that is outside the span of the $K$ tokens is lost (first consequence).
The local features may therefore need to be kept and combined with the global update, with a balance that can differ between entities.
Together with the analysis in Section~\ref{sec:problem_local}, this also suggests that the element-level response may need to be recovered locally after the global update.
Second, the contact features may be compressed separately from the state features (second consequence), which is consistent with the separate inputs suggested in Section~\ref{sec:problem_fe} and is examined by the ablation study in Section~\ref{sec:results_components}.
Third, the tokeniser may affect how well localised responses are preserved (third consequence), so the tokeniser is made exchangeable and compared in Section~\ref{sec:results_global}.
In addition, since nodes and elements are different entities, it may be beneficial to distinguish them when they enter the latent space and to recover them separately.

\FloatBarrier
\subsection{Input encoding}
\label{sec:method_inputs}

Section~\ref{sec:method_compression} suggests that nodes and elements may need to be distinguished when they enter the latent space.
The model receives the four inputs defined in Section~\ref{sec:graph}: the state and contact inputs of the nodes and of the elements.
Each of the four inputs is encoded by its own Multilayer Perceptron (MLP), followed by layer normalisation.
Each MLP consists of two linear layers with a GELU activation.
A learnable entity-type embedding, $\mathbf{t}_{\mathrm{node}}$ for nodes and $\mathbf{t}_{\mathrm{elem}}$ for elements, is added to the encoded features of both streams:
\begin{equation}
\mathbf{h}_e^{S,0} = \mathrm{MLP}^{S}_{\tau(e)}\left(\mathbf{s}_e^{n}\right) + \mathbf{t}_{\tau(e)},
\qquad
\mathbf{h}_e^{C,0} = \mathrm{MLP}^{C}_{\tau(e)}\left(\mathbf{c}_e^{n}\right) + \mathbf{t}_{\tau(e)}.
\label{eq:encoder}
\end{equation}
Here, the superscripts $S$ and $C$ denote the state and contact streams, $\tau(e) \in \{\mathrm{node}, \mathrm{elem}\}$ is the type of entity $e$, and $\mathbf{h}_e^{S,0}, \mathbf{h}_e^{C,0} \in \mathbb{R}^{d}$ are the encoded features.
Since nodes and elements are later compressed into the same tokens, the type embedding allows the tokeniser and the global interaction to distinguish the two entity types.

\FloatBarrier
\subsection{Local node--element message passing}
\label{sec:method_mp}

Section~\ref{sec:problem_fe} shows that, within one explicit cycle, an element exchanges information only with its own nodes, first from the nodes to the element and then from the element to the nodes.
These local relations may therefore require only a small receptive field, while the long-range dependence identified in Section~\ref{sec:problem_learned} is handled by the global module.
Local processing is performed on the node--element bipartite graph in Fig.~\ref{fig:bipartite_graph}.
Each layer, referred to as a bipartite GraphSAGE layer in Fig.~\ref{fig:architecture}, consists of two half-steps.
It first updates the element features from their nodes, and then updates the node features from the updated elements:
\begin{align}
\mathbf{m}_i &= \frac{1}{|\mathcal{N}(i)|} \sum_{v \in \mathcal{N}(i)} \phi_{v \to i}\left(\mathbf{h}_v\right),
&
\mathbf{h}_i &\leftarrow \mathrm{LN}\left(\mathbf{h}_i + \beta_{i}\, \psi_{i}\left(\left[\mathbf{h}_i, \mathbf{m}_i\right]\right)\right),
\label{eq:mp_element}
\\
\mathbf{m}_v &= \frac{1}{|\mathcal{M}(v)|} \sum_{i \in \mathcal{M}(v)} \phi_{i \to v}\left(\mathbf{h}_i\right),
&
\mathbf{h}_v &\leftarrow \mathrm{LN}\left(\mathbf{h}_v + \beta_{v}\, \psi_{v}\left(\left[\mathbf{h}_v, \mathbf{m}_v\right]\right)\right).
\label{eq:mp_node}
\end{align}
Here, $\mathbf{m}_i$ and $\mathbf{m}_v$ are the aggregated messages to element $i$ and node $v$, $\phi_{v \to i}$, $\phi_{i \to v}$, $\psi_i$ and $\psi_v$ are two-layer MLPs, $[\cdot,\cdot]$ denotes concatenation, $\mathrm{LN}$ denotes layer normalisation, and $\beta_i$ and $\beta_v$ are learnable scalars that scale the residual updates, initialised to 0.1 so that each layer initially makes a small change.
In Eq.~\eqref{eq:mp_node}, $\mathbf{h}_i$ is the element feature already updated by Eq.~\eqref{eq:mp_element}.

Each half-step follows the form of GraphSAGE \cite{Hamilton2017GraphSAGE}.
The features of the neighbours are averaged, combined with the feature of the receiving entity by concatenation, and passed through an MLP.
Three changes are made for the present problem.
First, the graph contains two entity types, so each direction of information transfer has its own message function, $\phi_{v \to i}$ or $\phi_{i \to v}$, and its own update function, $\psi_i$ or $\psi_v$.
Second, the update is added to the existing feature as a residual and followed by layer normalisation, so that several layers can be stacked without changing the scale of the features.
Third, the messages are computed from the entity features only, without edge features, and mean aggregation is used in both directions.
The order of the two updates follows the FE update, in which the element strain is computed from the nodal displacement (Eq.~\eqref{eq:strain}) and the element stress is then assembled into the nodal force (Eq.~\eqref{eq:assembly}).

This form is simpler than the edge-aware cross-attention used in CAtt-BiGNN \cite{Zhao2026CAttBiGNN}, in which the geometry of the mesh is described by edge features and the messages are weighted by attention.
In BiLoGo, the positions of the nodes and element centres are part of the entity inputs described in Section~\ref{sec:graph_inputs}, so the geometry is already contained in the entity features, and the edges are only used to define the connectivity.
Mean aggregation is therefore used in the local message passing, which avoids computing edge features and attention weights on every edge and can reduce the computational cost.
However, the deformation of an element depends on the positions of its nodes relative to each other, as explained in Section~\ref{sec:problem_local}, and such relative quantities are difficult to obtain from the mean of the absolute node features.
The attention-based aggregation from the nodes to the element is therefore kept in the kinematics-conditioned element update described in Section~\ref{sec:method_kinematic}, where it is combined with the relative geometry of the element nodes and applied after the global module.

Two message-passing layers are applied to each stream before the global module.
Each layer has its own parameters, and the two streams use separate layers.
The two streams are processed independently at this stage and do not exchange information until the cross-attention in the global module described in Section~\ref{sec:method_global}.
Two layers extend the receptive field of each entity by two element rings, which covers the local node--element relations, while the long-range information is provided by the global module.
The resulting features, $\mathbf{h}_e^{S}$ and $\mathbf{h}_e^{C}$, are referred to as the local state features and the local contact features.
They are the input to the global module, and they are also kept for the gated fusion, the kinematics-conditioned element update and the decoders.

\FloatBarrier
\subsection{Global module: tokenisation and token interaction}
\label{sec:method_global}

Section~\ref{sec:problem_learned} shows that the learned update can depend on information from distant regions of the blank within one learned step.
Section~\ref{sec:method_compression} suggests that the contact and state streams may be compressed separately, and that the tokeniser may affect how well localised responses are preserved.
For each stream, the local features of all nodes and elements are collected into one set of $N_e$ entities and compressed by the tokeniser of that stream.
Nodes and elements are compressed into the same tokens, so that the global interaction couples both entity types.
The two streams use separate tokenisers, so that the contact tokens are formed from the local contact features only and are not averaged with the local state features.

The tokens of each stream are first processed by a Transformer block with pre-layer normalisation,
\begin{equation}
\mathbf{Z} \leftarrow \mathbf{Z} + \mathrm{MHA}\left(\mathrm{LN}(\mathbf{Z}), \mathrm{LN}(\mathbf{Z}), \mathrm{LN}(\mathbf{Z})\right),
\qquad
\mathbf{Z} \leftarrow \mathbf{Z} + \mathrm{MLP}\left(\mathrm{LN}(\mathbf{Z})\right),
\label{eq:token_block}
\end{equation}
where $\mathbf{Z}$ is the matrix of tokens, $\mathrm{MHA}(\mathbf{Q}, \mathbf{K}, \mathbf{V})$ denotes multi-head attention with queries $\mathbf{Q}$, keys $\mathbf{K}$ and values $\mathbf{V}$, and $\mathrm{MLP}$ is a two-layer MLP with a GELU activation, which forms the feed-forward part of the Transformer block.
This step exchanges information between all regions of the blank within each stream.
The state tokens $\mathbf{Z}^{S}$ and the contact tokens $\mathbf{Z}^{C}$ are then combined by cross-attention, in which the contact tokens are the queries and the state tokens are the keys and values:
\begin{equation}
\mathbf{R} = \mathbf{Z}^{C} + \mathrm{MHA}\left(\mathrm{LN}(\mathbf{Z}^{C}), \mathrm{LN}(\mathbf{Z}^{S}), \mathrm{LN}(\mathbf{Z}^{S})\right),
\qquad
\mathbf{R} \leftarrow \mathbf{R} + \mathrm{MLP}\left(\mathrm{LN}(\mathbf{R})\right).
\label{eq:cross_attention}
\end{equation}
In Eq.~\eqref{eq:cross_attention}, each contact token collects the state information that is relevant to its contact condition.
This follows Section~\ref{sec:problem_fe}: the contact force determines where and how the blank is loaded, and the current state of the blank determines how it responds.
The resulting tokens $\mathbf{R}$ are mapped back to the nodes and elements by the recovery operation of the contact-stream tokeniser, which gives the recovered global feature $\mathbf{g}_e$ of each entity.
The global interaction in Eqs.~\eqref{eq:token_block} and \eqref{eq:cross_attention} is the same for both tokenisers described below.

\FloatBarrier
\subsubsection{Physical-slice tokeniser (BiLoGo-T)}
\label{sec:method_slicing}

In BiLoGo-T, each stream is compressed into $K$ tokens by learned physical slicing \cite{Wu2024Transolver}.
The assignment weights of entity $e$ are computed from its local feature,
\begin{equation}
\mathbf{a}_e = \mathrm{softmax}\left(\mathbf{W}_a \mathbf{h}_e\right),
\label{eq:slice_weights}
\end{equation}
where $\mathbf{a}_e = [a_{e1}, \dots, a_{eK}]$ and $\mathbf{W}_a \in \mathbb{R}^{K \times d}$ is a learnable matrix, separate for the two streams.
The tokens are computed by Eq.~\eqref{eq:slice}, and the recovered features are computed with the assignment weights of the contact stream.
The number of tokens $K$ is small compared with $N_e$, so the global interaction is inexpensive, and the memory required by the compression and recovery grows linearly with $N_e$.

\FloatBarrier
\subsubsection{Spatial supernode tokeniser (BiLoGo-U)}
\label{sec:method_supernode}

In BiLoGo-U, each stream is compressed into $N_s$ spatially distributed supernodes, adapted from the supernode pooling and the Perceiver-type decoder of UPT \cite{Alkin2024UPT}.
Each entity is assigned a position $\mathbf{X}_e$ in the reference configuration: the in-plane coordinates of the node in the initial blank, or the in-plane coordinates of the element centre.
The reference configuration is used so that the positions of the tokens do not change during the rollout.
A subset of $N_s$ entities is selected as supernodes.
The selection is random during training and fixed for each simulation case during evaluation, and the same supernodes are used for both streams.
Each supernode $k$ aggregates the features of the entities within a radius $r$ of its position, keeping at most $N_{\max}$ nearest entities:
\begin{equation}
\mathbf{z}_k = \frac{1}{|\mathcal{R}(k)|} \sum_{e \in \mathcal{R}(k)} \mathrm{MLP}\left(\left[\tilde{\mathbf{h}}_e, \tilde{\mathbf{h}}_k\right]\right),
\qquad
\tilde{\mathbf{h}}_e = \mathbf{W}_{\mathrm{in}} \mathbf{h}_e + \mathrm{PE}\left(\mathbf{X}_e\right).
\label{eq:supernode_pooling}
\end{equation}
Here, $\mathcal{R}(k)$ is the set of entities assigned to supernode $k$, $\tilde{\mathbf{h}}_k$ is the projected feature of the supernode itself, $\mathbf{W}_{\mathrm{in}}$ is a learnable matrix, and $\mathrm{PE}$ is a sinusoidal positional encoding.
The supernodes are used directly as tokens, without the further compression into a smaller set of latent tokens used in the original UPT, since this compression led to a higher thinning error on the Corner+Rib dataset in our experiments.
After the global interaction, the recovered feature of each entity is computed by a Perceiver-type cross-attention block, in which the query is computed from the positional encoding of the entity and the keys and values are the tokens $\mathbf{R}$.
In this work, $N_s = 2048$, $r = 20$~mm and $N_{\max} = 32$.
The state and contact streams use the same supernode locations, and each stream forms its own $N_s$ tokens at these locations.

Compared with Eq.~\eqref{eq:slice}, each token in Eq.~\eqref{eq:supernode_pooling} contains information from a local region of the blank only, and the recovery of each entity depends on its position.
The pooling therefore does not mix distant regions, although the subsequent self-attention exchanges information between all tokens.
The number and spacing of the supernodes determine the spatial resolution of the global update.
The cost of the self-attention grows with $N_s^2$, and the cost of the position-queried recovery grows with $N_e N_s$, so BiLoGo-U is more expensive than BiLoGo-T.

\FloatBarrier
\subsection{Entity-wise gated fusion of global and local features}
\label{sec:method_fusion}

Section~\ref{sec:method_compression} shows that the recovered global features have limited rank, so the local features may need to be kept and combined with the global update, with a balance that can differ between entities.
The recovery of local detail from reduced representations has also been studied in other contexts, such as mesh-based super-resolution, local field reconstruction and the representation of discontinuities \cite{Barwey2025MeshSuperResolution,Xu2022SuperMeshing,GuevaraGarban2025LocalFields,Lai2026LocallyEnhanced}.
The recovered global feature $\mathbf{g}_e$ is combined with the local state feature $\mathbf{h}_e^{S}$ by gated fusion:
\begin{gather}
\mathbf{y}_e = \mathrm{LN}\left(\mathbf{W}_y\, \mathrm{LN}\left(\mathbf{h}_e^{S}\right)\right),
\qquad
\boldsymbol{\gamma}_e = \sigma\left(\mathrm{MLP}_{\gamma}\left(\left[\mathrm{LN}(\mathbf{g}_e), \mathbf{y}_e\right]\right)\right),
\nonumber \\
\mathbf{f}_e = \mathbf{g}_e + \alpha\, \boldsymbol{\gamma}_e \odot \mathbf{W}_c \mathbf{y}_e.
\label{eq:fusion}
\end{gather}
Here, $\mathbf{y}_e$ is the projected local state feature, $\boldsymbol{\gamma}_e \in (0,1)^{d}$ is the gate, $\sigma$ is the sigmoid function, $\odot$ denotes element-wise multiplication, $\mathbf{W}_y$ and $\mathbf{W}_c$ are learnable matrices, $\alpha$ is a learnable scalar, and $\mathbf{f}_e$ is the fused feature.

Eq.~\eqref{eq:fusion} has the form of a gated residual connection \cite{Lim2021TFT}, in which the residual branch is scaled by a learnable scalar initialised to zero \cite{Bachlechner2021ReZero}.
In BiLoGo, the recovered global feature forms the main path and the local state feature forms the gated residual branch.
The fusion therefore keeps the local state information alongside the global update, instead of replacing it.
The gate is computed for each entity and each feature channel from both the global update and the local feature.
Where the global update already describes the response, the gate can stay small; where the response varies over short distances, for example near reinforcing ribs, more local information can be added.
Separate fusion modules with their own parameters are used for nodes and elements, since their features describe different quantities.
The fusion is applied before the post-global message passing, so that the subsequent local layers operate on features that already contain the restored local information.

The gated fusion uses only the local state features.
The local contact features are passed to the outputs through the kinematics-conditioned element update described in Section~\ref{sec:method_kinematic} and as direct inputs to the decoders described in Section~\ref{sec:method_decoder}, so they are not added again in the fusion.

\FloatBarrier
\subsection{Kinematics-conditioned element update}
\label{sec:method_kinematic}

Section~\ref{sec:problem_local} shows that the element deformation and thinning are mainly determined by the kinematics of the three element nodes, and Section~\ref{sec:method_compression} shows that this local information may not be fully preserved in the recovered global features.
In addition, the local message passing does not provide the relative geometry of the element nodes, as described in Section~\ref{sec:method_mp}.
After the fusion, the element features are updated using the kinematics of the element nodes.
This is performed in three steps.
First, a kinematic descriptor is computed for each node of the element from the current rollout state.
Second, the descriptor is combined with the local contact features of the node to form a message from the node to the element.
Third, the three messages are aggregated by attention and added to the element feature as a residual update.

For each element $i$ and each of its nodes $v \in \mathcal{N}(i)$, a 16-dimensional kinematic descriptor is computed:
\begin{equation}
\mathbf{k}_{iv} = \left[\mathbf{r}_{iv}^{0},\ \|\mathbf{r}_{iv}^{0}\|,\ \mathbf{r}_{iv}^{n},\ \|\mathbf{r}_{iv}^{n}\|,\ \boldsymbol{\delta}_{iv}^{n},\ \|\boldsymbol{\delta}_{iv}^{n}\|,\ \mathbf{r}_{iv}^{n} - \mathbf{r}_{iv}^{0},\ \|\mathbf{r}_{iv}^{n} - \mathbf{r}_{iv}^{0}\|\right].
\label{eq:kinematic_descriptor}
\end{equation}
Here, $\mathbf{r}_{iv}^{0} = \mathbf{x}_v^{0} - \bar{\mathbf{x}}_i^{0}$ and $\mathbf{r}_{iv}^{n} = \mathbf{x}_v^{n} - \bar{\mathbf{x}}_i^{n}$ are the positions of node $v$ relative to the element centre in the reference and current configurations, where $\mathbf{x}_v^{0}$ is the position of node $v$ in the initial blank and $\bar{\mathbf{x}}_i$ is the mean position of the three element nodes.
$\boldsymbol{\delta}_{iv}^{n} = \Delta\mathbf{u}_v^{n} - \overline{\Delta\mathbf{u}}_i^{n}$ is the displacement increment of node $v$ relative to the mean increment of the element nodes, and $\|\cdot\|$ denotes the Euclidean norm.

The descriptor follows the element kinematics in Section~\ref{sec:problem_local}.
Since the edge vectors in Eq.~\eqref{eq:edge_matrices} are differences between the relative node positions, for example $\mathbf{x}_b^{n} - \mathbf{x}_a^{n} = \mathbf{r}_{ib}^{n} - \mathbf{r}_{ia}^{n}$, the descriptors of the three nodes contain the information that determines the in-plane deformation gradient in Eq.~\eqref{eq:deformation_gradient}.
Relative quantities are used so that the descriptor does not depend on the position of the element in the blank.
The same node--element geometry is described by the edge features of CAtt-BiGNN \cite{Zhao2026CAttBiGNN}, in which it is encoded on every edge and updated in every processor layer, whereas BiLoGo uses it once, in the node-to-element direction after the global module.
The descriptors are computed from the current rollout state at every learned step and are standardised with the statistics of the training set.

The descriptors of the three nodes are combined with the local contact features of the nodes and aggregated to the element by attention:
\begin{equation}
\mathbf{q}_{iv} = \mathrm{MLP}_{q}\left(\left[\mathrm{MLP}_{k}(\mathbf{k}_{iv}),\ \mathbf{h}_v^{C}\right]\right),
\qquad
w_{iv} = \underset{v \in \mathcal{N}(i)}{\mathrm{softmax}}\left(\mathrm{MLP}_{w}\left(\mathrm{LN}\left(\left[\mathbf{f}_i, \mathbf{q}_{iv}\right]\right)\right)\right),
\label{eq:kinematic_attention}
\end{equation}
\begin{equation}
\mathbf{f}_i \leftarrow \mathbf{f}_i + \beta_u\, \mathrm{MLP}_{u}\left(\left[\mathbf{f}_i,\ \textstyle\sum_{v \in \mathcal{N}(i)} w_{iv}\, \mathbf{q}_{iv}\right]\right).
\label{eq:kinematic_update}
\end{equation}
Here, $\mathbf{f}_i$ is the fused feature of element $i$ (Eq.~\eqref{eq:fusion}), $\mathbf{h}_v^{C}$ is the local contact feature of node $v$ after the pre-global message passing, $\mathbf{q}_{iv}$ is the message from node $v$ to element $i$, $w_{iv}$ is its attention weight, $\mathrm{MLP}_{k}$, $\mathrm{MLP}_{q}$, $\mathrm{MLP}_{w}$ and $\mathrm{MLP}_{u}$ are two-layer MLPs, and $\beta_u$ is a learnable scalar initialised to zero.
Eq.~\eqref{eq:kinematic_attention} is an attention pooling over the three nodes of the element, and Eq.~\eqref{eq:kinematic_update} is a residual update with the same zero initialisation as in Eq.~\eqref{eq:fusion}.
The local contact features $\mathbf{h}_v^{C}$ are included because the contact condition of the nodes affects the deformation of the element, for example through the contact pressure and friction.
The attention weights allow the three nodes to contribute differently, for example when only some of the nodes are in contact.
This step changes only the element features.
It is placed after the fusion and before the post-global message passing, so that the updated element information is passed to the nodes by the following element-to-node updates.

\FloatBarrier
\subsection{Post-global local refinement and output decoders}
\label{sec:method_decoder}

After the kinematics-conditioned element update, two layers of local bipartite message passing (Eqs.~\eqref{eq:mp_element} and \eqref{eq:mp_node}) are applied to the fused node and element features.
These two layers have the same form as the pre-global layers in Section~\ref{sec:method_mp}, but their own parameters, and they operate on a single set of fused features instead of two separate streams.
Neighbouring entities receive similar recovered global features, and this refinement restores the local differences between them.
It also passes the updated element information to the nodes, in the same way as the element stresses are assembled into the nodal forces in the FE update.
The resulting node and element features are denoted by $\mathbf{o}_v$ and $\mathbf{o}_i$.

The node decoder predicts the standardised displacement increment from the refined feature and the local features of both streams,
\begin{equation}
\Delta\tilde{\mathbf{u}}_v^{n+1} = \mathrm{MLP}_{\mathrm{node}}\left(\left[\mathbf{h}_v^{S},\ \mathbf{h}_v^{C},\ \mathbf{o}_v\right]\right).
\label{eq:node_decoder}
\end{equation}
The element decoder additionally uses the refined features of the element nodes, aggregated by attention:
\begin{gather}
\lambda_{iv} = \underset{v \in \mathcal{N}(i)}{\mathrm{softmax}}\left(\mathrm{MLP}_{\lambda}\left(\left[\mathbf{o}_i, \mathbf{o}_v\right]\right)\right),
\nonumber \\
\tilde{\theta}_i^{n+1} = \mathrm{MLP}_{\mathrm{elem}}\left(\left[\mathbf{h}_i^{S},\ \mathbf{h}_i^{C},\ \mathbf{o}_i,\ \textstyle\sum_{v \in \mathcal{N}(i)} \lambda_{iv}\, \mathbf{o}_v\right]\right).
\label{eq:element_decoder}
\end{gather}
Here, $\Delta\tilde{\mathbf{u}}_v^{n+1}$ and $\tilde{\theta}_i^{n+1}$ are the standardised nodal displacement increment and the standardised element thinning at the next learned step, which are converted to physical values with the statistics of the training set.
The direct inputs $\mathbf{h}_e^{S}$ and $\mathbf{h}_e^{C}$ ensure that the local features of both streams reach the outputs, even if part of this information is not kept by the global module.
The aggregation of the node features in the element decoder follows the node-to-element transfer in Eq.~\eqref{eq:strain}.

\FloatBarrier
\section{Training settings}
\label{sec:setup}

This section describes the training settings of BiLoGo.

BiLoGo was trained with the Adam optimiser \cite{Kingma2015Adam}, with an initial learning rate of $1\times10^{-4}$ that decays exponentially to $1\times10^{-6}$.
The batch size is one simulation.
The loss follows \cite{Zhao2026CAttBiGNN} and is the unweighted sum of the mean squared errors of the standardised displacement increment and the standardised thinning:
\begin{equation}
\begin{split}
\mathcal{L} = {} & \frac{1}{3 N_{p} |\mathcal{V}|} \sum_{n=0}^{N_p-1} \sum_{v \in \mathcal{V}} \sum_{d \in \{x,y,z\}} \left( \Delta\tilde{u}_{v,d}^{\,n+1} - \Delta\tilde{u}_{v,d}^{\ast\,n+1} \right)^{2} \\
& + \frac{1}{N_{p} |\mathcal{T}|} \sum_{n=0}^{N_p-1} \sum_{i \in \mathcal{T}} \left( \tilde{\theta}_i^{\,n+1} - \tilde{\theta}_i^{\ast\,n+1} \right)^{2}.
\end{split}
\label{eq:loss}
\end{equation}
Here, $\mathcal{V}$ and $\mathcal{T}$ are the sets of nodes and elements, $N_p$ is the number of learned steps of a simulation, $d$ denotes the Cartesian direction, the tilde denotes standardised quantities, and the superscript $\ast$ denotes the FE solution.

For the Dome and Corner datasets, BiLoGo was trained with teacher forcing for 1000 epochs.
With teacher forcing, the input of each learned step is computed from the FE solution, and the loss is computed over all $N_p = 10$ learned steps of each simulation.
For the Corner+Rib dataset, BiLoGo was first trained with teacher forcing for 500 epochs, followed by an autoregressive stage of 50 epochs, which reduces the difference between the inputs seen in training and in rollout.
In the autoregressive stage, one window of five consecutive learned steps is taken from each simulation in each epoch.
The starting step of the window changes between epochs, so that all six possible starting steps are used.
The first step of the window uses the FE state as input.
In the following four steps, the state is updated with the predicted displacement increment and the predicted thinning of the previous step, and the contact input is recomputed from the updated geometry.
The predicted state is detached from the computational graph, so the gradient is not propagated between steps.
This is similar to the pushforward trick \cite{Brandstetter2022MPPDE}.
In this stage, the loss of each step is computed with Eq.~\eqref{eq:loss} for a single step, the losses of the five steps are summed, and the model parameters are updated once for each simulation.
The checkpoint with the lowest validation error is used for testing.
The hyperparameters of BiLoGo are listed in Table~\ref{tab:hyperparameters}.
The two variants differ only in the tokeniser.

\begin{table}[t]
\centering
\caption{Hyperparameters of BiLoGo.}
\label{tab:hyperparameters}
\small
\begin{tabular}{lll}
\hline
Component & Hyperparameter & Value \\
\hline
All & Latent dimension $d$ & 128 \\
Local message passing & Pre-global layers per stream & 2 \\
 & Post-global layers & 2 \\
Global interaction & Attention heads & 4 \\
 & Self-attention blocks per stream & 1 \\
 & Cross-attention blocks & 1 \\
BiLoGo-T & Tokens per stream $K$ & 16 \\
BiLoGo-U & Supernode tokens per stream $N_s$ & 2048 \\
 & Supernode radius $r$ & 20~mm \\
 & Maximum entities per supernode $N_{\max}$ & 32 \\
\hline
\end{tabular}
\end{table}

\FloatBarrier
\section{Results and discussion}
\label{sec:results}

This section evaluates BiLoGo on the three forming datasets and discusses the results.
Section~\ref{sec:results_setup} introduces the baselines, the ablation studies and the evaluation metrics.
Section~\ref{sec:results_overall} compares the prediction accuracy of all models, and Section~\ref{sec:results_smoothing} compares the local accuracy near the corner and the ribs.
Section~\ref{sec:results_components} presents the ablation study on the Corner and Corner+Rib datasets, and Section~\ref{sec:results_global} examines the effect of the tokeniser in the global module.
Section~\ref{sec:results_cost} compares the computational cost, and Section~\ref{sec:results_limitations} discusses the selection of the global module and the limitations of this work.

\FloatBarrier
\subsection{Baselines, ablations and evaluation metrics}
\label{sec:results_setup}

Three groups of baseline models are compared with BiLoGo.
The first group contains single-level graph-based models.
MeshGraphNets (MGN) \cite{Pfaff2021MeshGraphNet} performs message passing directly on the original mesh, and the Recurrent GNN (RGNN) \cite{Chen2024PIERS} adds recurrent updates to MGN.
FEIH-GNN \cite{Gao2024FEIH} is an FE-inspired hypergraph model.
CAtt-BiGNN \cite{Zhao2026CAttBiGNN} predicts the nodal displacement and the element thinning on a node--element bipartite graph with edge-aware cross-attention.
The second group contains hierarchical graph-based models.
UMGN \cite{Zhao2026RUGNN} extends MGN with downsampling and upsampling between graph levels, RUGNN \cite{Zhao2026RUGNN} adds recurrent updates to UMGN, and CAtt-BiUGNN \cite{Zhao2026CAttBiGNN} is the hierarchical version of CAtt-BiGNN.
The third group contains Transformer-based neural operators.
Transolver \cite{Wu2024Transolver} uses learned physical slicing, and UPT \cite{Alkin2024UPT} compresses the mesh into spatially distributed supernodes.
GNN-Transformer is adapted from MGN-T \cite{Iparraguirre2026MGNT}, which places a global Transformer processor between local message-passing stages.
Since MGN-T is node-centred and does not predict element-level quantities, GNN-Transformer applies this structure to the node--element bipartite graph used in this work.

For the Dome dataset, BiLoGo-T is compared with MGN, RGNN, FEIH-GNN, CAtt-BiGNN, Transolver and GNN-Transformer.
For the Corner dataset, BiLoGo-T is compared with UMGN, RUGNN, CAtt-BiUGNN, Transolver and GNN-Transformer.
For the Corner+Rib dataset, BiLoGo-T and BiLoGo-U are compared with CAtt-BiUGNN, Transolver, GNN-Transformer and UPT.
Following \cite{Zhao2026CAttBiGNN}, the single-level graph-based models are used for the Dome dataset, which has a small mesh, and the hierarchical graph-based models are used for the larger Corner and Corner+Rib meshes.
The results of the graph-based baselines on the Dome dataset and of all baselines on the Corner dataset are taken from \cite{Zhao2026CAttBiGNN}, using the seed-0 runs for the Corner dataset.
All models use the same data and contact encoding on each dataset, as described in Section~\ref{sec:graph_inputs}.
On the Corner+Rib dataset, all baselines are trained with the same procedure as BiLoGo, including the autoregressive stage, as described in Section~\ref{sec:setup}.

Two groups of ablation studies are performed.
First, the contribution of each component of BiLoGo is examined on the Corner dataset, and the results are presented in Section~\ref{sec:results_components}.
One component is removed at a time: the kinematics-conditioned element update, the gated fusion, the cross-attention, the separate token streams for the state and contact inputs, and the bipartite GraphSAGE layers before and after the global module.
In addition, the same five components are removed from BiLoGo-U on the Corner+Rib dataset.
Second, the effect of the global module is examined on the Corner+Rib dataset by replacing the tokeniser of BiLoGo while keeping all other components, and the results are presented in Section~\ref{sec:results_global}.
In addition to the physical slices of BiLoGo-T and the supernodes of BiLoGo-U, the locally adaptive slicing of Transolver++ \cite{Luo2025TransolverPP} with 16 tokens per stream and the latent grid of GINO \cite{Li2023GINO} with a resolution of 48 are used in this comparison.

The evaluation metrics follow \cite{Zhao2026CAttBiGNN}.
Each test simulation is rolled out autoregressively over the 10 learned steps, starting from the undeformed blank.
The metrics are evaluated at the final learned step, since the final geometry and thinning distribution are the main quantities used to assess the formability of a design.
Each metric is computed for each test simulation and then averaged over the test set.
Two metrics are used for the nodal displacement.
The mean positional Euclidean error, $e_{\mathrm{pos}}$, is the mean Euclidean distance between the predicted and FE nodal positions.
The position Mean Absolute Error (MAE) is the mean absolute difference between the predicted and FE coordinates over all nodes and the three coordinate directions.
Two metrics are used for the element thinning.
The threshold-based thinning error is the mean absolute thinning error over the elements whose FE thinning magnitude exceeds 0.5\%, so that the regions with almost no deformation are excluded.
The top-$p$\% critical relative thinning error measures the accuracy in the most critical regions.
For each test simulation, the elements are ranked by their FE thinning at the final learned step, and the $p$\% of elements with the largest thinning are selected as the critical thinning region.
The relative error is then calculated between the mean predicted thinning and the mean FE thinning over this region.
For the Corner and Corner+Rib datasets, in which both thinning and thickening are pronounced, the $p$\% of elements with the largest thickening are also selected as the critical thickening region, and the relative errors of the two regions are averaged \cite{Zhao2026CAttBiGNN}.
Following \cite{Zhao2026CAttBiGNN}, $p = 5$ is used for the Dome dataset, in which the high thinning is distributed over a larger region, and $p = 1$ is used for the Corner dataset, in which it is concentrated near the corner.
For the Corner+Rib dataset, $p = 0.25$ is used, since its mesh contains approximately four times as many elements as the Corner mesh, which gives a similar number of critical elements.

The computational cost is compared on the Corner and Corner+Rib datasets in terms of the number of trainable parameters, the training time per epoch without validation, the peak GPU memory during training and inference, and the inference time of a full 10-step rollout.
The cost measurements on the Corner dataset were performed on an NVIDIA L4 GPU, and those on the Corner+Rib dataset were performed on an NVIDIA A40 GPU.
All models of each dataset were measured on the same GPU.

\FloatBarrier
\subsection{Overall comparison}
\label{sec:results_overall}

\FloatBarrier
\subsubsection{Dome dataset}
\label{sec:results_dome}
The results on the Dome dataset are listed in Table~\ref{tab:results_dome}.
BiLoGo-T gives the lowest displacement errors.
Compared with GNN-Transformer, which gives the second-lowest mean positional Euclidean error, BiLoGo-T reduces this error by 3.8\%.
Compared with CAtt-BiGNN, BiLoGo-T reduces the mean positional Euclidean error by 15.7\% and the position MAE by 15.1\%.
CAtt-BiGNN gives the lowest threshold-based thinning error of 0.377\%, followed by BiLoGo-T with 0.396\%.
For the top-5\% critical relative thinning error, GNN-Transformer gives the lowest value of 1.52\%, followed by BiLoGo-T with 1.61\%.
In contrast to the Corner and Corner+Rib datasets, Transolver and GNN-Transformer give displacement errors close to that of BiLoGo-T on the Dome dataset.
This may be attributed to the deformation pattern of the Dome component.
The displacement and the thinning have similar spatial patterns, and both are concentrated near the top of the dome.
In addition, the geometry of the dome is smooth, and the deformation changes gradually over the blank.
Such a response may be represented by a small number of feature-based tokens.

\begin{table}[!htbp]
\centering
\caption{Prediction accuracy on the Dome dataset at the final learned step of the 10-step rollout. The values are averaged over the test simulations. The best value in each column is shown in bold, and the second-best value is underlined.}
\label{tab:results_dome}
\resizebox{\textwidth}{!}{%
\begin{tabular}{lcccc}
\hline
Model & $e_{\mathrm{pos}}$ (mm) & Position MAE (mm) & Threshold-based thinning error (\%) & Top-5\% critical thinning error \\
\hline
MGN \cite{Pfaff2021MeshGraphNet} & 0.5137 & 0.2385 & 0.581 & 2.49\% \\
FEIH-GNN \cite{Gao2024FEIH} & 0.5291 & 0.2488 & 0.502 & 2.67\% \\
CAtt-BiGNN \cite{Zhao2026CAttBiGNN} & 0.4001 & 0.1893 & \underline{0.377} & 1.80\% \\
Transolver \cite{Wu2024Transolver} & 0.3510 & \underline{0.1681} & 0.612 & 2.64\% \\
GNN-Transformer & \underline{0.3505} & 0.1691 & 0.602 & \textbf{1.52\%} \\
BiLoGo-T & \textbf{0.3371} & \textbf{0.1608} & \textbf{0.396} & \underline{1.61\%} \\
\hline
\end{tabular}}
\end{table}

\begin{figure}[!htbp]
\centering
\includegraphics[width=\textwidth]{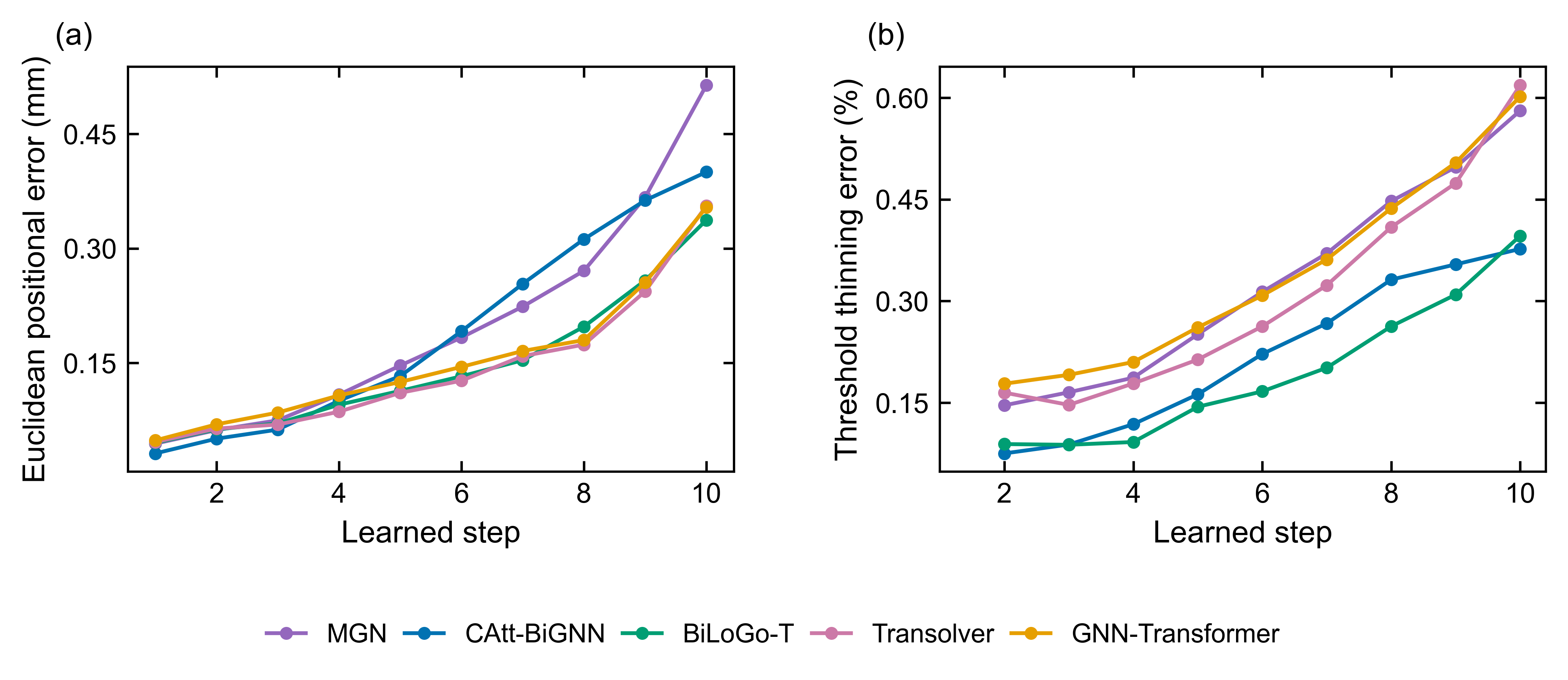}
\caption{Errors over the 10 learned steps of the rollout on the Dome test set. (a) Mean positional Euclidean error. (b) Threshold-based thinning error. The lines show the mean over the test simulations. The thinning error is shown from the second step, since few elements exceed the thinning threshold at the first step.}
\label{fig:dome_rollout}
\end{figure}

The evolution of the errors over the rollout on the Dome dataset is shown in Fig.~\ref{fig:dome_rollout}.
The positional errors of BiLoGo-T, Transolver and GNN-Transformer are similar at all learned steps.
CAtt-BiGNN gives the lowest positional error in the first three steps, but its error is larger than those of these three models from the fifth step onwards. The positional error of MGN increases most rapidly in the last two steps, and MGN gives the largest error at the final step. BiLoGo-T and CAtt-BiGNN give lower threshold-based thinning errors than MGN, Transolver and GNN-Transformer at all learned steps. The threshold-based thinning error of BiLoGo-T is lower than that of CAtt-BiGNN from the fourth to the ninth step, whereas CAtt-BiGNN gives a lower error at the final step. 

\begin{figure}[!htbp]
\centering
\includegraphics[width=\textwidth]{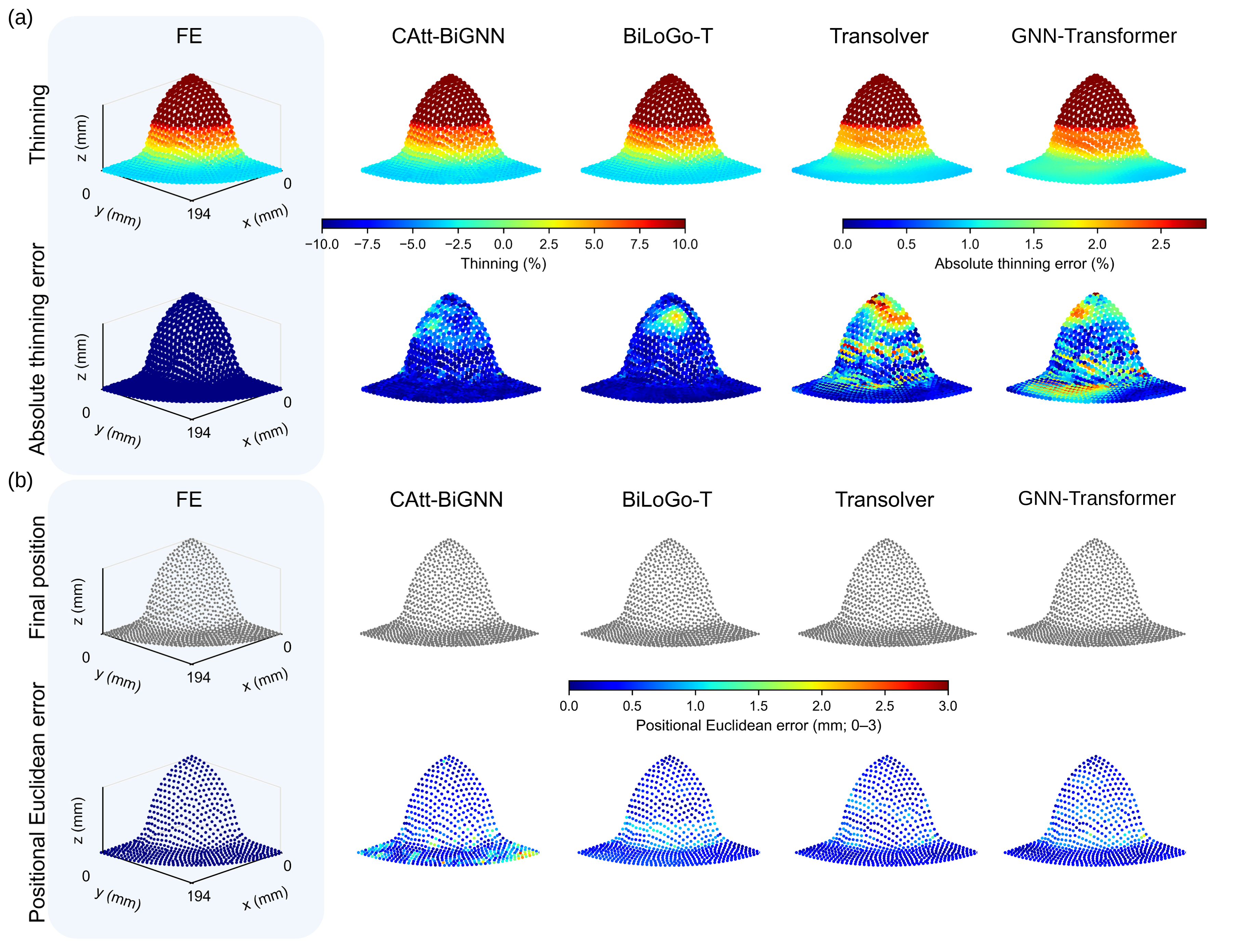}
\caption{Final-step comparison for an example simulation from the Dome test set. (a) Thinning (top) and absolute thinning error (bottom). (b) Final positions (top) and positional Euclidean errors (bottom).}
\label{fig:dome_fields}
\end{figure}

The thinning fields of an example simulation from the test set are shown in Fig.~\ref{fig:dome_fields}(a).
In the FE result, the thinning increases gradually from the flange to the top of the dome, and all models reproduce this overall distribution.
The thinning errors of Transolver and GNN-Transformer are distributed over the side of the dome and the flange.
The errors of CAtt-BiGNN and BiLoGo-T are smaller, and the largest error of BiLoGo-T is located near the top of the dome.
The final positions and the positional errors of the same simulation are shown in Fig.~\ref{fig:dome_fields}(b).
The positional errors of all models are small.
The largest positional errors of CAtt-BiGNN occur in the flange, whereas those of the other models occur on the lower part of the dome.

\FloatBarrier
\subsubsection{Corner dataset}
\label{sec:results_corner}
The results on the Corner dataset are listed in Table~\ref{tab:results_corner}.
BiLoGo-T gives the lowest displacement errors and the lowest threshold-based thinning error.
Compared with CAtt-BiUGNN, which gives the second-lowest errors, BiLoGo-T reduces the mean positional Euclidean error by 15.5\%, the position MAE by 15.0\% and the threshold-based thinning error by 8.1\%.
The top-1\% critical relative thinning error of BiLoGo-T is 3.57\%, which is higher than the value of 3.02\% obtained by CAtt-BiUGNN in the run used here, but within the range of the three CAtt-BiUGNN runs reported in \cite{Zhao2026CAttBiGNN}, $3.52 \pm 0.52$\%.
Among the Transformer-based neural operators, Transolver gives the largest errors.
GNN-Transformer, which adds local message passing before and after the global module, reduces the mean positional Euclidean error from 1.023~mm to 0.420~mm.
BiLoGo-T further reduces this error to 0.271~mm, and the effect of its components is examined in Section~\ref{sec:results_components}.

\begin{table}[!htbp]
\centering
\caption{Prediction accuracy on the Corner dataset at the final learned step of the 10-step rollout. The values are averaged over the test simulations. The best value in each column is shown in bold, and the second-best value is underlined.}
\label{tab:results_corner}
\resizebox{\textwidth}{!}{%
\begin{tabular}{lcccc}
\hline
Model & $e_{\mathrm{pos}}$ (mm) & Position MAE (mm) & Threshold-based thinning error (\%) & \makecell[c]{Top-1\% critical relative\\thinning error} \\
\hline
UMGN \cite{Zhao2026RUGNN} & 0.4971 & 0.2299 & 0.428 & 14.20\% \\
RUGNN \cite{Zhao2026RUGNN} & 0.4005 & 0.1865 & 0.396 & 13.27\% \\
CAtt-BiUGNN \cite{Zhao2026CAttBiGNN} & \underline{0.3209} & \underline{0.1490} & \underline{0.185} & \textbf{3.02\%} \\
Transolver \cite{Wu2024Transolver} & 1.0231 & 0.4787 & 0.480 & 10.75\% \\
GNN-Transformer & 0.4199 & 0.1989 & 0.353 & 8.89\% \\
BiLoGo-T & \textbf{0.2713} & \textbf{0.1266} & \textbf{0.170} & \underline{3.57\%} \\
\hline
\end{tabular}}
\end{table}

In addition, BiLoGo-T requires less GPU memory than CAtt-BiUGNN, as compared in Section~\ref{sec:results_cost}.

\begin{figure}[!htbp]
\centering
\includegraphics[width=\textwidth]{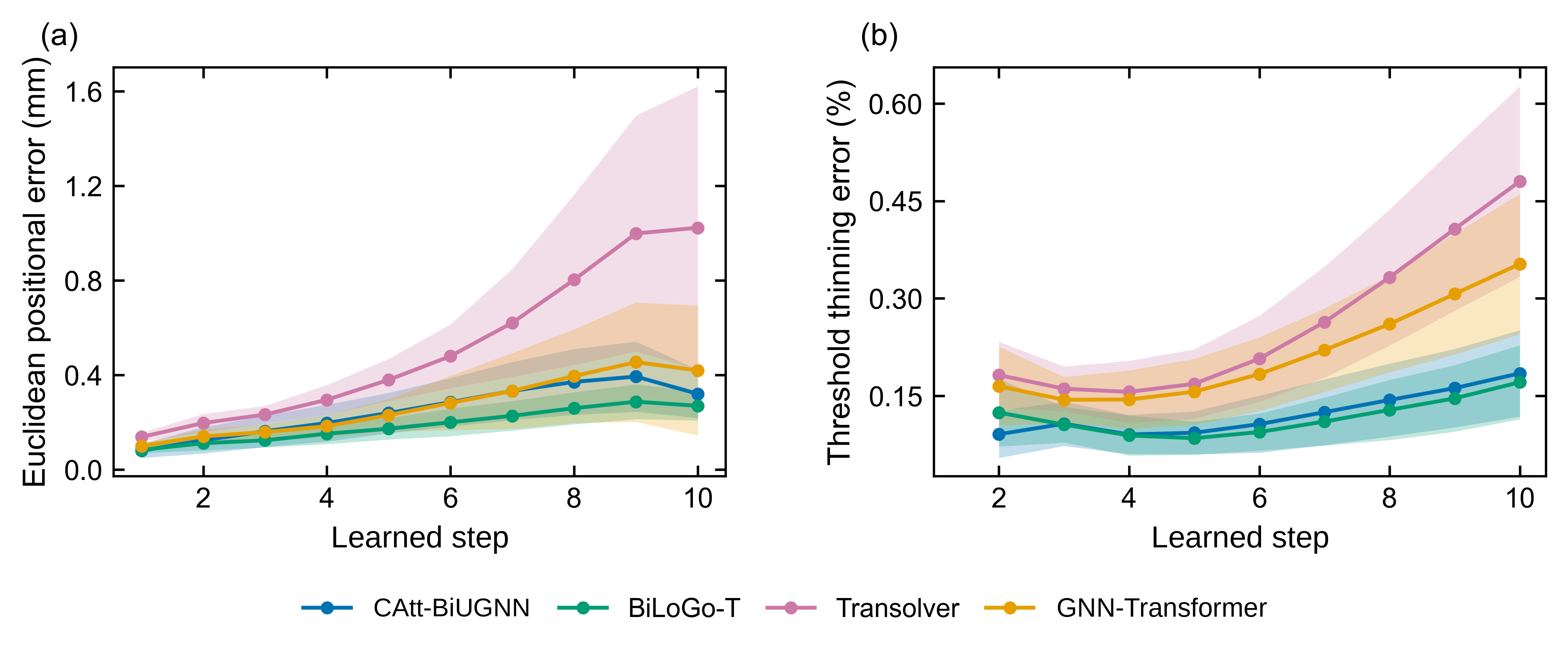}
\caption{Errors over the 10 learned steps of the rollout on the Corner test set. (a) Mean positional Euclidean error. (b) Threshold-based thinning error. The lines show the mean over the test simulations, and the shaded regions show the standard deviation. The thinning error is shown from the second step, since few elements exceed the thinning threshold at the first step.}
\label{fig:corner_rollout}
\end{figure}

The evolution of the errors over the rollout on the Corner dataset is shown in Fig.~\ref{fig:corner_rollout}.
BiLoGo-T gives the lowest positional error from the second learned step onwards, and the difference from the other models increases towards the end of the rollout.
The threshold-based thinning errors of BiLoGo-T and CAtt-BiUGNN are similar over the whole rollout and considerably lower than those of GNN-Transformer and Transolver.

\begin{figure}[!htbp]
\centering
\includegraphics[width=\textwidth]{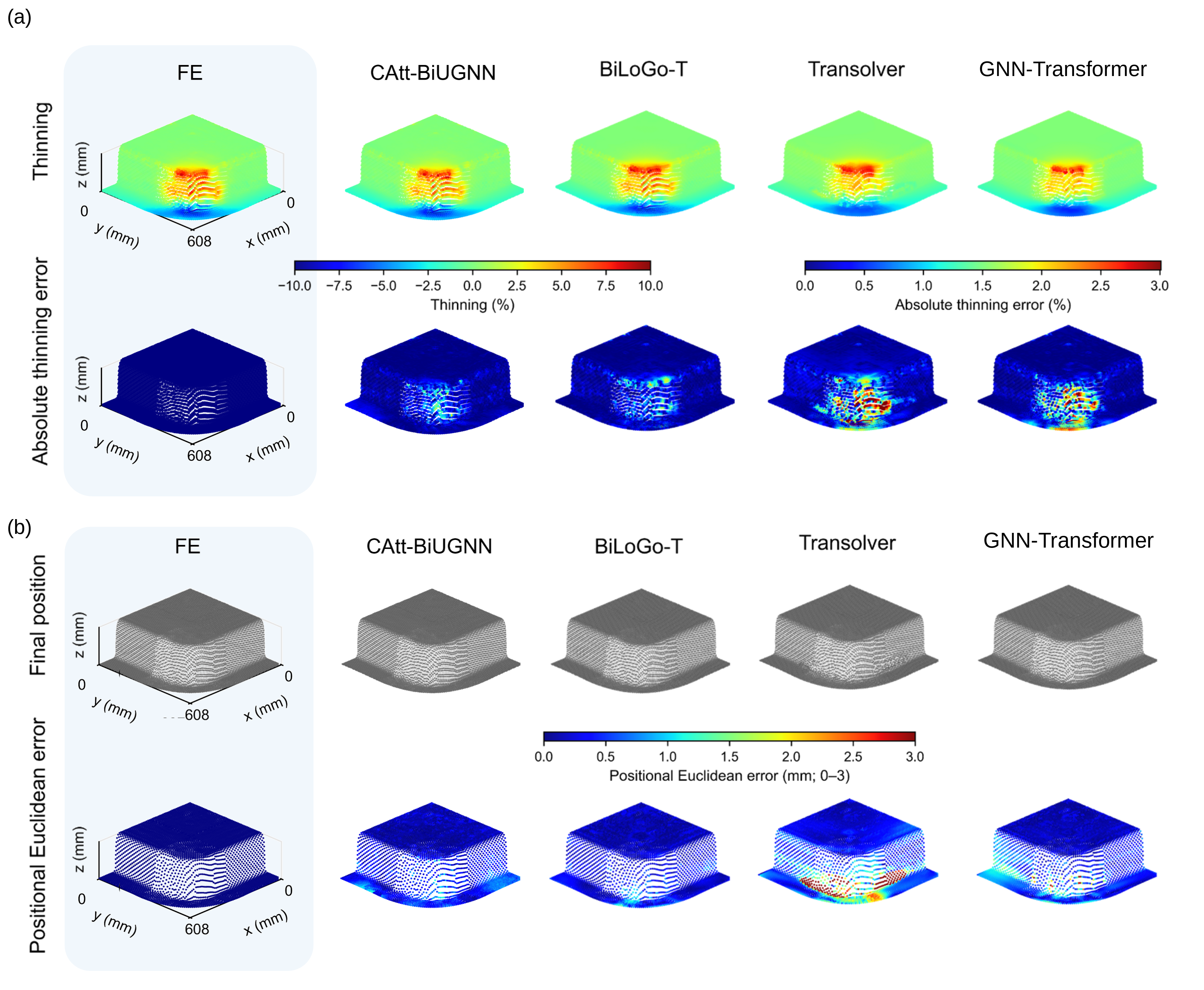}
\caption{Final-step comparison for an example simulation from the Corner test set. (a) Thinning (top) and absolute thinning error (bottom). (b) Final positions (top) and positional Euclidean errors (bottom).}
\label{fig:corner_fields}
\end{figure}

The thinning fields of an example simulation from the test set are shown in Fig.~\ref{fig:corner_fields}(a).
In the FE result, the largest thinning occurs on the side wall at the corner, and the thinning varies strongly over short distances within this region.
CAtt-BiUGNN and BiLoGo-T reproduce both the extent of this region and the variation within it.
For Transolver, the high-thinning region is larger and more uniform than in the FE result, and for GNN-Transformer, part of the variation within this region is not captured.
These differences may indicate a smoothing of the localised thinning.
As a result, the absolute thinning errors of Transolver and GNN-Transformer are large over the corner wall, whereas those of CAtt-BiUGNN and BiLoGo-T are smaller and limited to a few elements.
The final positions and the positional errors of the same simulation are shown in Fig.~\ref{fig:corner_fields}(b).
For Transolver, the largest positional errors occur in the lower part of the corner wall and in the flange, where the material is drawn from the flange into the side walls.
For GNN-Transformer, the errors in this region are smaller, but remain larger than those of CAtt-BiUGNN and BiLoGo-T.
The local accuracy near the corner is compared over all test designs in Section~\ref{sec:results_smoothing}.

\FloatBarrier
\subsubsection{Corner+Rib dataset}
\label{sec:results_rib}
The results on the Corner+Rib dataset are listed in Table~\ref{tab:results_rib}.
BiLoGo-U gives the lowest displacement errors and the lowest threshold-based thinning error, and CAtt-BiUGNN gives the second-lowest values.
Compared with CAtt-BiUGNN, BiLoGo-U reduces the mean positional Euclidean error by 8.3\%, the position MAE by 7.2\% and the threshold-based thinning error by 10.7\%, and reduces the top-0.25\% critical relative thinning error from 5.11\% to 3.86\%.
BiLoGo with the GINO latent grid gives the lowest top-0.25\% critical relative thinning error of 2.25\%, but its other errors are higher than those of BiLoGo-U and CAtt-BiUGNN.
In addition, BiLoGo-U requires considerably less GPU memory than CAtt-BiUGNN, as shown in Section~\ref{sec:results_cost}.
Transolver, GNN-Transformer and the original UPT give the largest errors on this dataset.
The differences between the global modules are examined in Section~\ref{sec:results_global}.

\begin{table}[!htbp]
\centering
\caption{Prediction accuracy on the Corner+Rib dataset at the final learned step of the 10-step rollout. The values are averaged over the test simulations. The best value in each column is shown in bold, and the second-best value is underlined.}
\label{tab:results_rib}
\resizebox{\textwidth}{!}{%
\begin{tabular}{lcccc}
\hline
Model & $e_{\mathrm{pos}}$ (mm) & Position MAE (mm) & Threshold-based thinning error (\%) & Top-0.25\% critical thinning error \\
\hline
CAtt-BiUGNN \cite{Zhao2026CAttBiGNN} & \underline{0.2637} & \underline{0.1248} & \underline{0.206} & 5.11\% \\
Transolver \cite{Wu2024Transolver} & 1.3091 & 0.6243 & 0.989 & 16.75\% \\
GNN-Transformer & 0.7410 & 0.3575 & 0.726 & 10.96\% \\
UPT \cite{Alkin2024UPT} & 1.3466 & 0.6366 & 0.486 & 9.40\% \\
\hline
BiLoGo-T & 0.4493 & 0.2146 & 0.391 & \underline{3.37\%} \\
BiLoGo + Transolver++ & 0.6526 & 0.3170 & 0.507 & 6.77\% \\
BiLoGo + GINO & 0.2698 & 0.1282 & 0.211 & \textbf{2.25\%} \\
BiLoGo-U & \textbf{0.2419} & \textbf{0.1158} & \textbf{0.184} & 3.86\% \\
\hline
\end{tabular}}
\end{table}

\begin{figure}[!htbp]
\centering
\includegraphics[width=\textwidth]{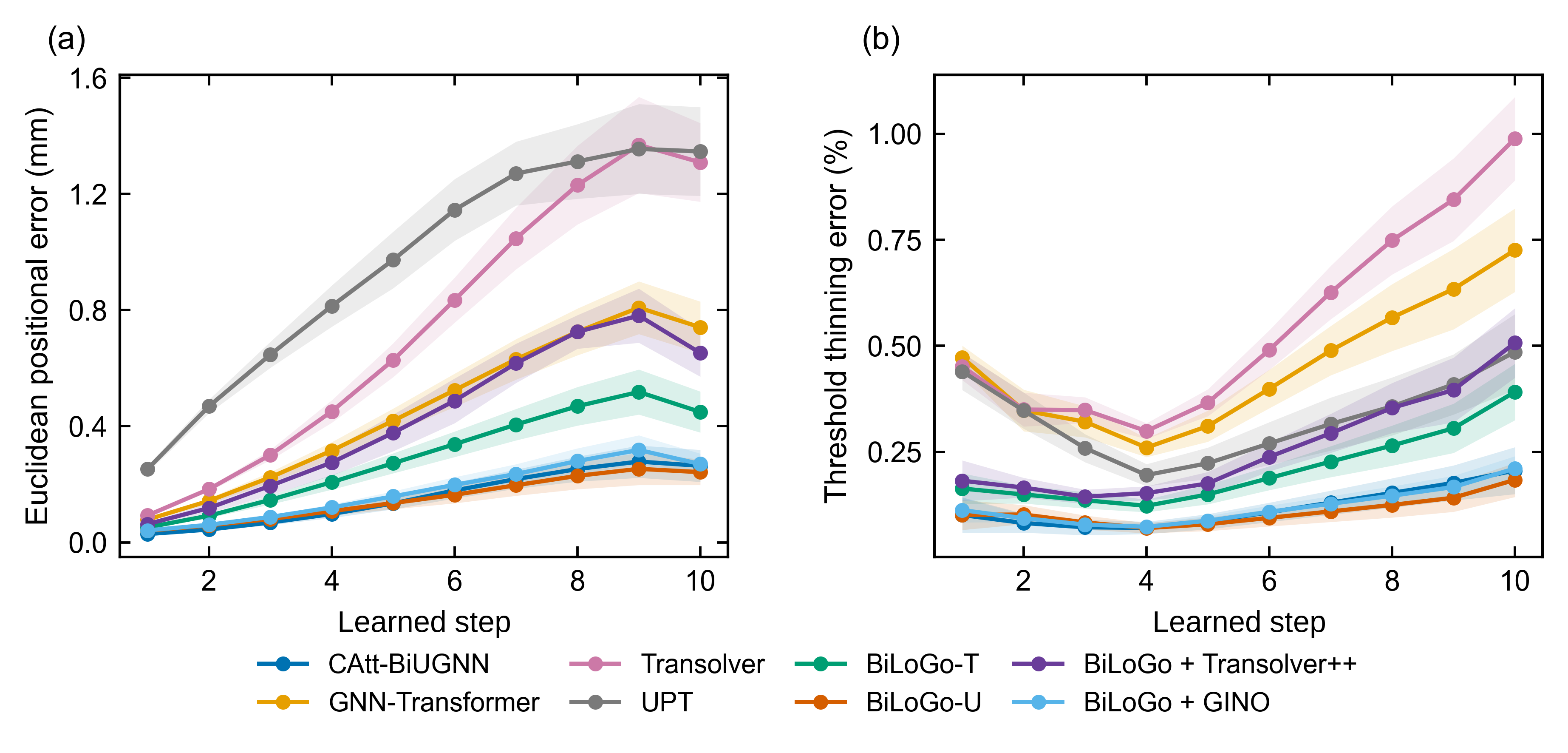}
\caption{Errors over the 10 learned steps of the rollout on the Corner+Rib test set. (a) Mean positional Euclidean error. (b) Threshold-based thinning error. The lines show the mean over the test simulations, and the shaded regions show the standard deviation.}
\label{fig:rib_rollout}
\end{figure}

The evolution of the errors over the rollout is shown in Fig.~\ref{fig:rib_rollout}.
BiLoGo-U, BiLoGo with the GINO latent grid and CAtt-BiUGNN give the lowest positional errors at all learned steps, and the differences between the models increase as the rollout proceeds.
The positional errors of Transolver, GNN-Transformer and the original UPT increase faster than those of the other models, which indicates a stronger accumulation of errors during the rollout.
The threshold-based thinning error of most models decreases over the first four steps and then increases.
This may be attributed to the small number of elements whose thinning exceeds the threshold in the early steps, so that the error is computed over a small set of elements.

\begin{figure}[!htbp]
\centering
\includegraphics[width=\textwidth,trim=150 8 155 12,clip]{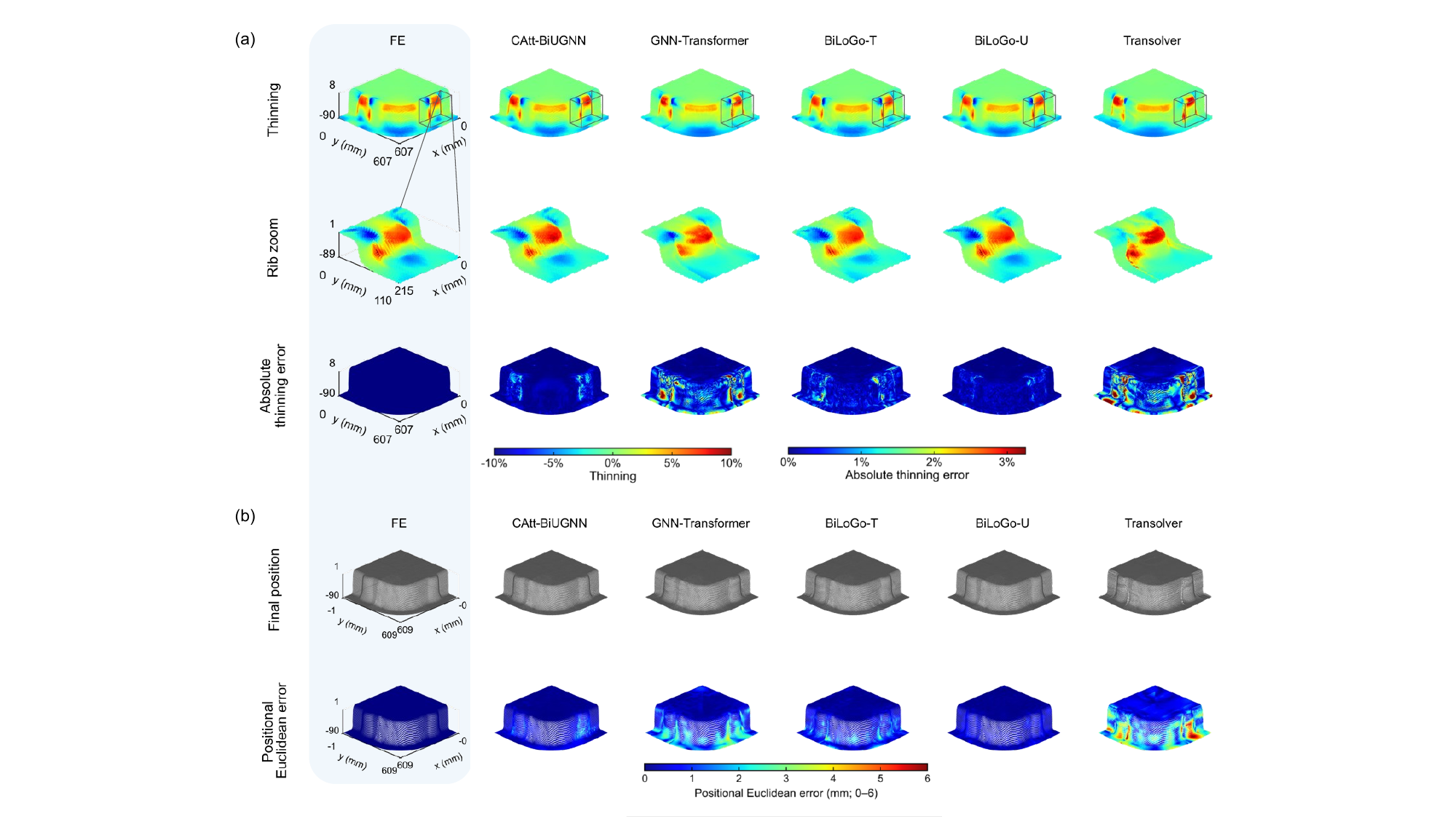}
\caption{Final-step comparison for an example simulation from the Corner+Rib test set. (a) Thinning (top), thinning near one rib (middle) and absolute thinning error (bottom). (b) Final positions (top) and positional Euclidean errors (bottom).}
\label{fig:rib_fields}
\end{figure}

The thinning fields of an example simulation from the test set are shown in Fig.~\ref{fig:rib_fields}(a).
In the FE result, the largest thinning near the rib is concentrated in a small region with large changes in thinning at its boundary.
CAtt-BiUGNN, BiLoGo-T and BiLoGo-U reproduce the shape and the peak value of this region.
For GNN-Transformer, the high-thinning region is more spread out and its peak value is lower than in the FE result, which is consistent with its low smoothing index reported in Section~\ref{sec:results_smoothing}.
For Transolver, the shape and the position of this region also differ from the FE result.
As a result, the absolute thinning errors of GNN-Transformer and Transolver are large near the ribs and the corner, whereas those of CAtt-BiUGNN and BiLoGo-U remain small over most of the component.
The final positions and the positional errors of the same simulation are shown in Fig.~\ref{fig:rib_fields}(b).
For GNN-Transformer, BiLoGo-T and Transolver, the largest positional errors occur on the side walls near the ribs and the corner.
For BiLoGo-U and CAtt-BiUGNN, the positional errors remain small over the whole component.
The local accuracy near the ribs is compared in Section~\ref{sec:results_smoothing}.

\FloatBarrier
\subsection{Smoothing of localised responses}
\label{sec:results_smoothing}
\begin{figure}[H]
\centering
\includegraphics[width=0.85\textwidth]{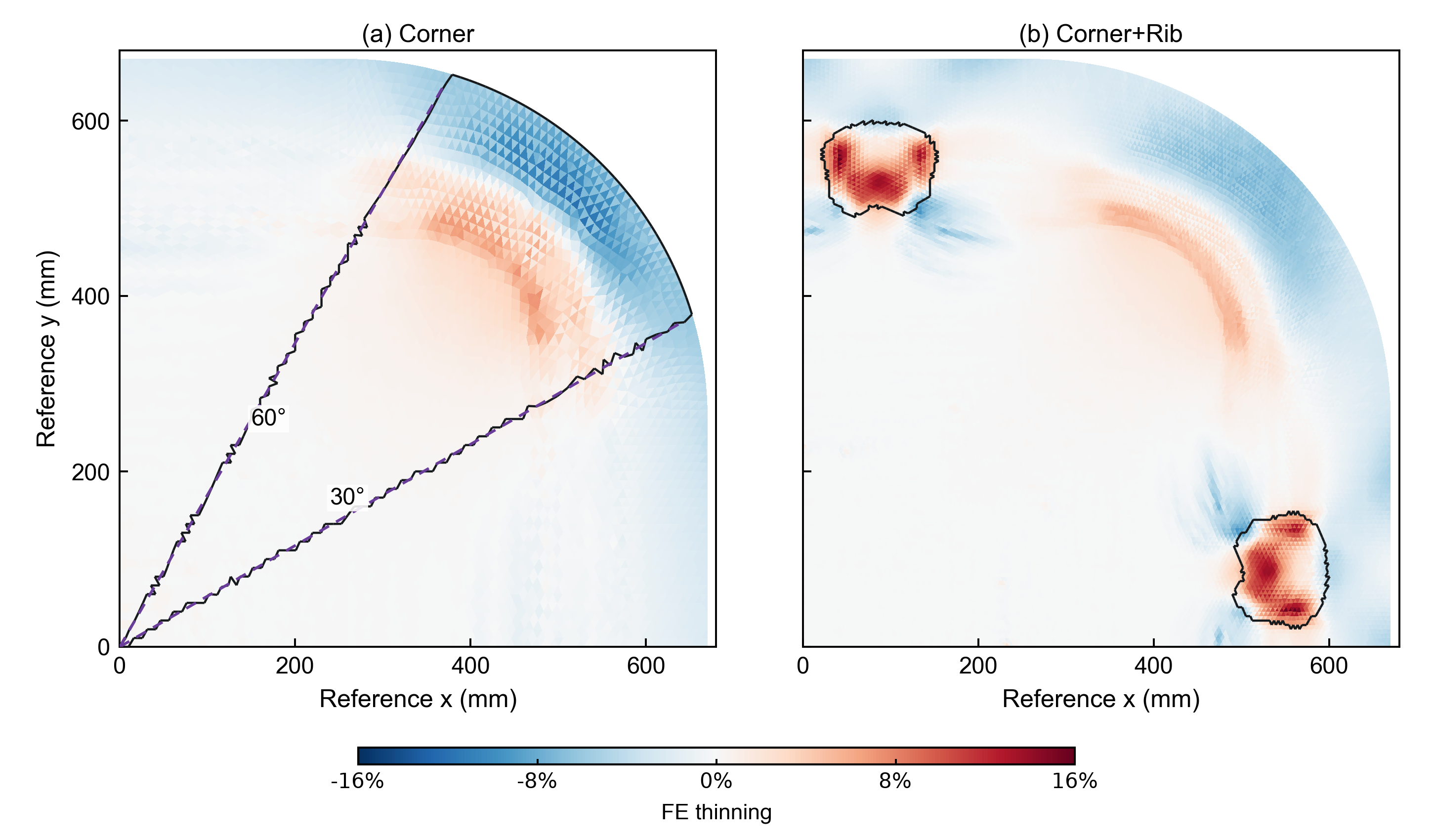}
\caption{Regions used for the local metrics, shown on the undeformed configuration with the FE thinning of one test design at the final learned step. (a) Corner region of the Corner dataset. (b) Rib region of the Corner+Rib dataset. The black lines show the region boundaries.}
\label{fig:local_regions}
\end{figure}

This section compares the local accuracy of the predicted thinning in the corner region of the Corner dataset and in the rib region of the Corner+Rib dataset.
Local metrics are evaluated at the final learned step in the two regions shown in Fig.~\ref{fig:local_regions}.
On the Corner dataset, the corner region contains the elements whose centres in the undeformed configuration lie between 30$^{\circ}$ and 60$^{\circ}$ from the symmetry planes, measured about the intersection of the two symmetry planes.
On the Corner+Rib dataset, the rib region contains the elements whose centres in the undeformed configuration lie within 20~mm of the ribs.
The region thinning error is the mean absolute thinning error over the region.
The smoothing index $S$ compares the local variation of the predicted thinning with that of the FE thinning:
\begin{equation}
S = \left( \frac{\sum_{(i,j) \in \mathcal{P}} w_{ij} \left( \hat{\theta}_i - \hat{\theta}_j \right)^2}{\sum_{(i,j) \in \mathcal{P}} w_{ij} \left( \theta_i - \theta_j \right)^2} \right)^{1/2},
\qquad
w_{ij} = \frac{1}{\lVert \mathbf{X}_i - \mathbf{X}_j \rVert^2}.
\label{eq:smoothing_index}
\end{equation}
Here, $\mathcal{P}$ is the set of pairs of elements in the region that share an edge, $\theta_i$ and $\hat{\theta}_i$ are the FE and predicted thinning of element $i$, and $\mathbf{X}_i$ is the centre of element $i$ in the undeformed configuration.
A value of $S$ smaller than 1 indicates that the predicted thinning field is smoother than the FE field.
Since $S$ only compares the magnitude of the local variation, the thinning-gradient error $G$ is also evaluated:
\begin{equation}
G = \left( \frac{1}{|\mathcal{P}|} \sum_{(i,j) \in \mathcal{P}} w_{ij} \left[ \left( \hat{\theta}_i - \hat{\theta}_j \right) - \left( \theta_i - \theta_j \right) \right]^2 \right)^{1/2}.
\label{eq:gradient_error}
\end{equation}
In addition, the mean positional Euclidean error averaged over all 10 learned steps, $\bar{e}_{\mathrm{pos}}$, is reported.

The local metrics in the corner region of the Corner dataset are listed in Table~\ref{tab:local_corner}.
Transolver and GNN-Transformer give smoothing indices below 0.5, and their region thinning errors and thinning-gradient errors are two to four times those of BiLoGo-T.
BiLoGo-T gives a smoothing index close to 1, and its local errors are comparable to those of CAtt-BiUGNN.

\begin{table}[H]
\centering
\caption{Local accuracy near the corner on the Corner test set. The region thinning error, $S$ and $G$ are evaluated at the final learned step, and $\bar{e}_{\mathrm{pos}}$ is averaged over all 10 learned steps. The values are averaged over the 25 test designs. The best value in each column is shown in bold, and the second-best value is underlined. For $S$, the value closest to 1 is regarded as the best.}
\label{tab:local_corner}
\resizebox{\textwidth}{!}{%
\begin{tabular}{lcccc}
\hline
Model & Region thinning error (\%) & $S$ & $G$ (\%/mm) & $\bar{e}_{\mathrm{pos}}$ (mm) \\
\hline
UMGN \cite{Zhao2026RUGNN} & 0.305 & 0.295 & 0.141 & 0.332 \\
RUGNN \cite{Zhao2026RUGNN} & 0.302 & 0.301 & 0.140 & 0.298 \\
CAtt-BiUGNN \cite{Zhao2026CAttBiGNN} & \textbf{0.127} & \underline{1.009} & \textbf{0.030} & \underline{0.251} \\
Transolver \cite{Wu2024Transolver} & 0.443 & 0.453 & 0.147 & 0.517 \\
GNN-Transformer & 0.322 & 0.469 & 0.137 & 0.270 \\
\hline
BiLoGo-T & \underline{0.157} & \textbf{1.001} & \underline{0.037} & \textbf{0.189} \\
\hline
\end{tabular}}
\end{table}

The local metrics in the rib region of the Corner+Rib dataset are listed in Table~\ref{tab:local_rib}.
The three Transformer-based neural operators give smoothing indices between 0.41 and 0.56, so the local variation of their predicted thinning is approximately half of that of the FE result.
With the same types of tokeniser, BiLoGo increases the smoothing index from 0.56 to 0.87 for the physical slices and from 0.41 to 0.94 for the supernodes, and reduces the region thinning error and the thinning-gradient error by 60--73\%.
The smoothing indices of BiLoGo-U and BiLoGo with the GINO latent grid are close to that of CAtt-BiUGNN, and their region thinning errors are lower.
However, BiLoGo with the GINO latent grid requires approximately three times the training GPU memory of BiLoGo-U, which is similar to that of CAtt-BiUGNN, and more than ten times its inference GPU memory, as shown in Section~\ref{sec:results_cost}.
Therefore, BiLoGo-U is considered the most suitable variant for the Corner+Rib dataset.

\begin{table}[H]
\centering
\caption{Local accuracy near the ribs on the Corner+Rib test set. The region thinning error, $S$ and $G$ are evaluated at the final learned step, and $\bar{e}_{\mathrm{pos}}$ is averaged over all 10 learned steps. The values are averaged over the 50 test designs. The best value in each column is shown in bold, and the second-best value is underlined. For $S$, the value closest to 1 is regarded as the best.}
\label{tab:local_rib}
\resizebox{\textwidth}{!}{%
\begin{tabular}{lcccc}
\hline
Model & Region thinning error (\%) & $S$ & $G$ (\%/mm) & $\bar{e}_{\mathrm{pos}}$ (mm) \\
\hline
CAtt-BiUGNN \cite{Zhao2026CAttBiGNN} & 0.608 & \textbf{0.989} & 0.167 & \underline{0.156} \\
Transolver \cite{Wu2024Transolver} & 1.838 & 0.562 & 0.480 & 0.745 \\
GNN-Transformer & 1.487 & 0.489 & 0.454 & 0.461 \\
UPT \cite{Alkin2024UPT} & 1.202 & 0.415 & 0.442 & 0.959 \\
\hline
BiLoGo-T & 0.585 & 0.874 & 0.193 & 0.295 \\
BiLoGo + GINO & \textbf{0.346} & \underline{0.957} & \underline{0.140} & 0.177 \\
BiLoGo-U & \underline{0.375} & 0.943 & \textbf{0.121} & \textbf{0.151} \\
\hline
\end{tabular}}
\end{table}

\begin{figure}[H]
\centering
\includegraphics[width=0.75\textwidth]{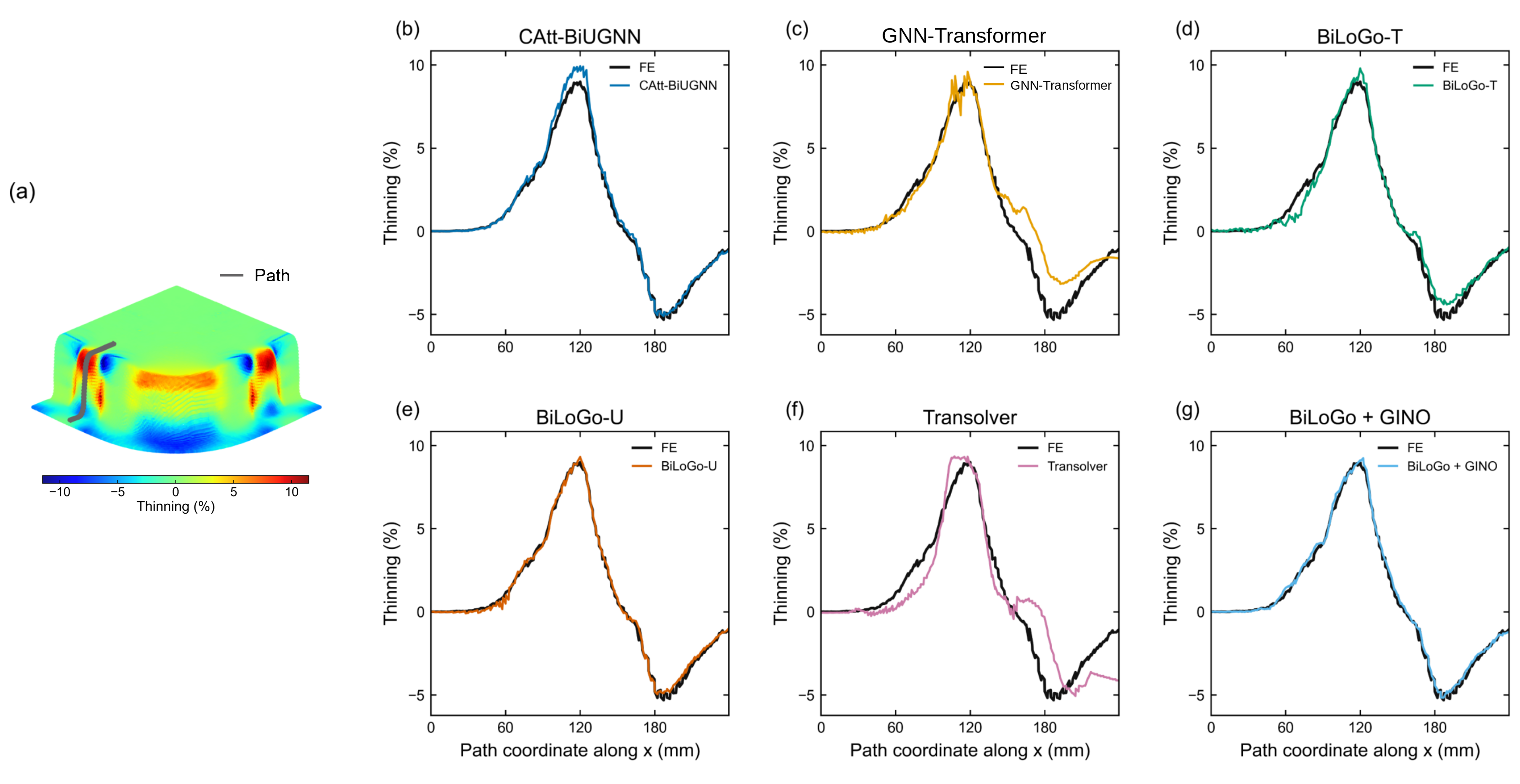}
\caption{Thinning along a path that crosses one rib, for one test simulation of the Corner+Rib dataset at the final learned step. (a) Position of the path on the FE thinning field. (b)--(g) FE and predicted thinning along the path.}
\label{fig:rib_path}
\end{figure}

To illustrate these differences, the thinning along a path that crosses one rib in one test design is shown in Fig.~\ref{fig:rib_path}.
Along this path, the FE thinning increases to a peak of approximately 9\% on the rib and decreases to approximately $-5$\% next to it.
BiLoGo-U and BiLoGo with the GINO latent grid follow both the peak and the minimum closely.
BiLoGo-T reproduces the position of the peak, but deviates from the FE result near the minimum.
CAtt-BiUGNN slightly overestimates the peak.
GNN-Transformer shows oscillations near the peak and deviates from the FE result beyond the peak, and Transolver shifts the position of the peak and does not reproduce the minimum.

On both datasets, BiLoGo gives smoothing indices between 0.87 and 1.00, whereas the Transformer-based neural operators with the same types of tokeniser give values between 0.41 and 0.56.

\FloatBarrier
\subsection{Ablation study of the BiLoGo components}
\label{sec:results_components}

The contribution of each component of BiLoGo is examined by an ablation study on the Corner dataset, in which one component is removed at a time.
The results are listed in Table~\ref{tab:ablation}.
Removing the kinematics-conditioned element update increases the threshold-based thinning error by 58.2\% and the top-1\% critical relative thinning error from 3.57\% to 7.53\%, while the mean positional Euclidean error increases by 7.4\%.
This agrees with the purpose of this component, which is to condition the element update on the kinematics of the element nodes.
Removing the gated fusion increases the positional error by 27.8\% and the thinning error by 27.1\%, which shows that the local state features are needed in addition to the recovered global features.
Removing the dual stream, so that the state and contact inputs are compressed into a single token stream, increases the positional error by 49.4\% and the thinning error by 30.6\%.
Removing the cross-attention, so that the contact tokens do not query the state tokens, gives slightly lower displacement errors, with a mean positional Euclidean error of 0.269~mm compared with 0.271~mm.
However, both thinning errors increase: the threshold-based thinning error increases from 0.170\% to 0.173\%, and the top-1\% critical relative thinning error from 3.57\% to 3.93\%.
On the Corner dataset, removing the dual stream gives a much larger increase in error than removing the cross-attention, so the benefit of the two streams may come mainly from compressing the state and contact inputs separately.
Removing the bipartite GraphSAGE layers, both before and after the global module, gives the largest increase in the displacement error, with a positional error of 0.717~mm, which shows that the local node--element coupling is necessary before and after the global module.

\begin{table}[!htbp]
\centering
\caption{Ablation study of the BiLoGo components on the Corner dataset at the final learned step of the 10-step rollout. The values are averaged over the test simulations. The best value in each column is shown in bold, and the second-best value is underlined.}
\label{tab:ablation}
\resizebox{\textwidth}{!}{%
\begin{tabular}{lcccc}
\hline
Model & $e_{\mathrm{pos}}$ (mm) & Position MAE (mm) & Threshold-based thinning error (\%) & Top-1\% critical thinning error \\
\hline
BiLoGo-T & \underline{0.2713} & \underline{0.1266} & \textbf{0.170} & \textbf{3.57\%} \\
w/o kinematics-conditioned element update & 0.2915 & 0.1355 & 0.269 & 7.53\% \\
w/o gated fusion & 0.3467 & 0.1614 & 0.216 & 4.47\% \\
w/o cross-attention & \textbf{0.2692} & \textbf{0.1251} & \underline{0.173} & 3.93\% \\
w/o dual stream & 0.4053 & 0.1918 & 0.222 & \underline{3.82\%} \\
w/o bipartite GraphSAGE & 0.7167 & 0.3352 & 0.275 & 5.78\% \\
\hline
\end{tabular}}
\end{table}

The same five components are also removed from BiLoGo-U on the Corner+Rib dataset, and the results are listed in Table~\ref{tab:ablation_rib}.
In the variant without the dual stream, the state and contact inputs are merged and compressed into 4096 supernode tokens, which equals the total number of tokens of the two streams in BiLoGo-U.
On this dataset, removing any of the five components increases all four errors.
Since all variants are also trained with the autoregressive stage described in Section~\ref{sec:setup}, this shows that the components remain effective when the model is trained on its own predicted rollouts.
Removing the kinematics-conditioned element update increases the threshold-based thinning error by 27.7\%, while the mean positional Euclidean error remains almost unchanged, which agrees with the result on the Corner dataset.
Removing the gated fusion increases the positional error by 15.0\% and the thinning error by 13.0\%.
Removing the bipartite GraphSAGE layers increases the positional error by 12.9\% and the thinning error by 13.6\%.
This increase is much smaller than on the Corner dataset, which may be because the supernodes already keep the spatial locality of the compression.
In contrast to the Corner dataset, removing the cross-attention gives the largest increase in the positional error, 24.7\%, while the thinning error increases by 7.6\%.
Removing the dual stream increases the positional error by 7.9\% and the thinning error by 20.7\%, and gives the largest top-0.25\% critical relative thinning error of 4.51\%.
On this dataset, the cross-attention therefore mainly affects the displacement, while the separate compression of the state and contact inputs mainly affects the thinning.
In addition, without any one of the five components, BiLoGo-U no longer gives lower displacement and threshold-based thinning errors than CAtt-BiUGNN (Table~\ref{tab:results_rib}).

\begin{table}[!htbp]
\centering
\caption{Ablation study of the BiLoGo-U components on the Corner+Rib dataset at the final learned step of the 10-step rollout. The values are averaged over the test simulations. The best value in each column is shown in bold, and the second-best value is underlined.}
\label{tab:ablation_rib}
\resizebox{\textwidth}{!}{%
\begin{tabular}{lcccc}
\hline
Model & $e_{\mathrm{pos}}$ (mm) & Position MAE (mm) & Threshold-based thinning error (\%) & Top-0.25\% critical thinning error \\
\hline
BiLoGo-U & \textbf{0.2419} & \textbf{0.1158} & \textbf{0.184} & \textbf{3.86\%} \\
w/o kinematics-conditioned element update & \underline{0.2429} & \underline{0.1169} & 0.235 & 4.34\% \\
w/o gated fusion & 0.2783 & 0.1328 & 0.208 & 3.92\% \\
w/o cross-attention & 0.3017 & 0.1440 & \underline{0.198} & \underline{3.88\%} \\
w/o dual stream & 0.2611 & 0.1254 & 0.222 & 4.51\% \\
w/o bipartite GraphSAGE & 0.2732 & 0.1318 & 0.209 & 4.10\% \\
\hline
\end{tabular}}
\end{table}

\FloatBarrier
\subsection{Effect of the global module on the Corner+Rib dataset}
\label{sec:results_global}

Section~\ref{sec:results_components} examined how the local message passing and the fusion and element update module affect the accuracy.
This subsection focuses on the global module.
All other components of BiLoGo are kept, and only the tokeniser is replaced, so that the effect of the tokeniser on the accuracy and the computational cost can be examined.
The comparison is made on the Corner+Rib dataset, in which the changes in the response are concentrated near the ribs.
Four tokenisers are compared: the physical slices of BiLoGo-T and the locally adaptive slicing of Transolver++, which group the entities by their features, and the latent grid of GINO and the supernodes of BiLoGo-U, which keep the spatial locality of the compression.
The accuracy of each variant is compared with that of the hierarchical graph model CAtt-BiUGNN, and the computational cost is compared in Section~\ref{sec:results_cost}.

The results of the four variants are listed in Table~\ref{tab:results_rib}.
The two feature-based tokenisers give the largest errors among these variants.
The mean positional Euclidean error is 0.449~mm with the physical slices of BiLoGo-T and 0.653~mm with the locally adaptive slices of Transolver++.
The two spatially local representations give considerably lower errors: 0.270~mm with the latent grid of GINO and 0.242~mm with the supernodes of BiLoGo-U.
The same order is obtained for the threshold-based thinning error.
Compared with CAtt-BiUGNN, which gives a mean positional Euclidean error of 0.264~mm, only BiLoGo-U gives lower displacement and threshold-based thinning errors, and BiLoGo with the GINO latent grid gives similar errors.
Since all other components are the same in the four variants, these differences may be attributed mainly to the tokeniser.
The number of tokens also differs between the variants.
However, BiLoGo-T was also trained with 128 slices per stream, and it gave larger errors than with 16 slices.

These results are consistent with the analysis in Section~\ref{sec:method_compression}.
In slice-based compression, entities are assigned to tokens according to their features, and the recovered global features of all entities are combinations of the same few tokens.
The large changes in thinning over a short distance near the ribs may therefore be smoothed in the recovered global features.
The locally adaptive slicing of Transolver++ changes how the slices are formed, but the assignment is still based on the features, and it does not improve the accuracy in this case.
In contrast, each supernode and each cell of the latent grid collects information from a local region of the blank only, so that the recovered global update can vary over short distances.
Between the two spatially local representations, BiLoGo-U gives lower errors and requires considerably less GPU memory, as shown in Section~\ref{sec:results_cost}.

The slice assignment of BiLoGo-T on the Dome and Corner datasets is shown in \ref{app:slices}.
On both datasets, the weights of each slice form continuous regions on the blank, such as the flange, the side wall and the top face, and the size of these regions is similar to that of the variation in the response.
This may explain why the physical slices give accurate predictions on these two datasets.

\begin{figure}[!htbp]
\centering
\includegraphics[width=\textwidth]{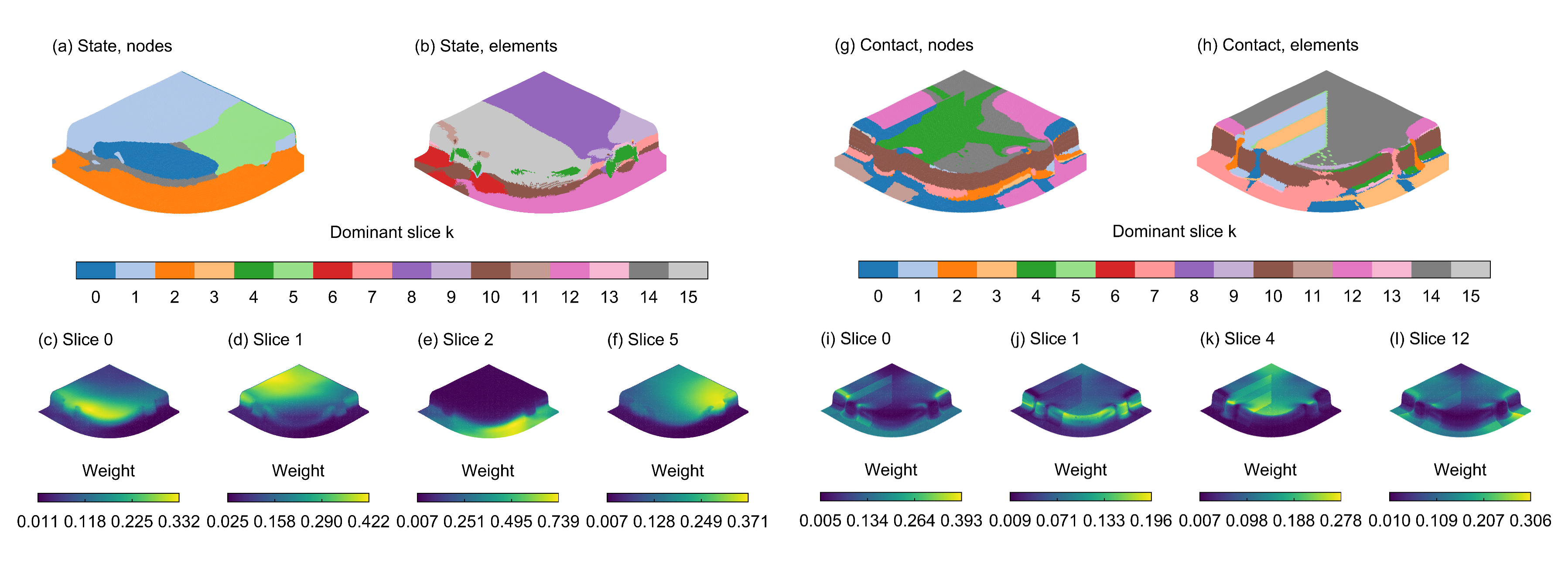}
\caption{Slice assignment of BiLoGo-T for one test simulation of the Corner+Rib dataset at learned step 9, with the FE state as input. (a)--(f) State stream. (g)--(l) Contact stream. (a), (g) Dominant slice of each node. (b), (h) Dominant slice of each element. (c)--(f) Assignment weights of state slices 0, 1, 2 and 5. (i)--(l) Assignment weights of contact slices 0, 1, 4 and 12. Each weight panel uses its own colour scale.}
\label{fig:token_slices}
\end{figure}

In contrast, the slice assignment of BiLoGo-T on the Corner+Rib dataset is shown in Fig.~\ref{fig:token_slices}.
In the state stream, the dominant slices form a few large continuous regions, such as the top face, the side walls and the flange (Fig.~\ref{fig:token_slices}(a) and (b)).
For example, state slice 1 has large weights on the top face, and state slice 2 in the flange (Fig.~\ref{fig:token_slices}(d) and (e)).
In the contact stream, the side walls, including both ribs and the corner, are mainly assigned to the same slice, and regions that are far apart, such as the two sides of the top face, are also assigned to the same slice (Fig.~\ref{fig:token_slices}(g) and (h)).
In both streams, the nodes and elements in the same region are often assigned to different slices, which shows that the tokeniser distinguishes the two entity types, as intended by the entity-type embedding described in Section~\ref{sec:method_inputs}.
The assignment weights of four contact slices are shown in Fig.~\ref{fig:token_slices}(i)--(l).
Contact slice 0 has large weights at the ribs on both side walls, and contact slice 1 has large weights along the lower part of the side walls, including the ribs and the corner.
The information near a rib is therefore combined with the information from the other rib and from the rest of the side walls in the same token.
Contact slice 4 has large weights in the region in contact with the punch, and contact slice 12 has large weights at the two outer ends of the flange, which are far apart.
These slices group the entities according to their contact condition rather than their position.
In addition, the largest assignment weight of each contact slice is below 0.4, so the information of each entity is distributed over several slices.
The two streams therefore partition the blank in different ways, which illustrates why compressing the state and contact inputs separately reduces the errors in the ablation study in Section~\ref{sec:results_components}.

\begin{figure}[!htbp]
\centering
\includegraphics[width=\textwidth]{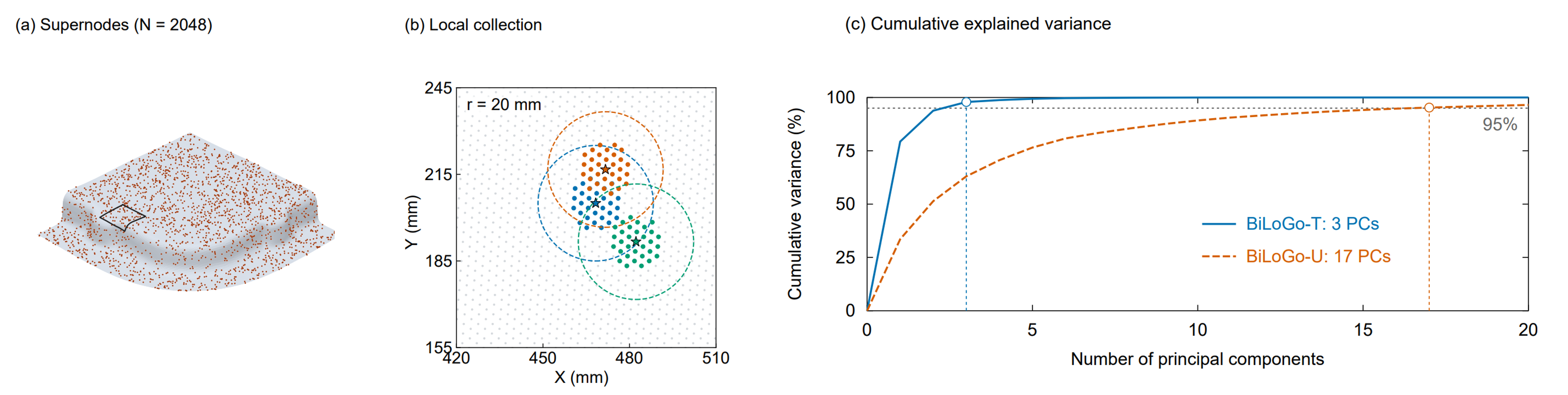}
\caption{Supernodes of BiLoGo-U and comparison of the recovered global features, for the simulation in Fig.~\ref{fig:token_slices}. (a) Positions of the 2048 supernodes. (b) Entities collected by three supernodes in the region marked in (a), with the collection radius $r = 20$~mm, in the reference coordinates. (c) Cumulative explained variance of the principal components of the recovered nodal global features $\mathbf{g}_e$ of BiLoGo-T and BiLoGo-U.}
\label{fig:token_supernodes}
\end{figure}

The supernodes of BiLoGo-U are shown in Fig.~\ref{fig:token_supernodes}(a) and (b).
In contrast to the slices, each supernode collects only the entities within a radius of 20~mm in the reference configuration.
The effect of the compression on the recovered global features is quantified in Fig.~\ref{fig:token_supernodes}(c).
A principal component analysis was applied to the recovered global features $\mathbf{g}_e$ of the 16,049 nodes, which are the outputs of the global module before the gated fusion.
The features were taken from one test simulation at learned step 9, with the FE state as input, and each feature channel was centred but not scaled.
For BiLoGo-T, three principal components explain 95\% of the variance, although the latent dimension is 128 and the number of tokens is 16.
For BiLoGo-U, 17 principal components are required.
This is consistent with the analysis in Section~\ref{sec:method_compression}: the recovered global features of BiLoGo-T lie in a lower-dimensional space than those of BiLoGo-U.
The effect on the predicted thinning is measured directly by the smoothing index in Table~\ref{tab:local_rib}, which is 0.87 for BiLoGo-T and 0.94 for BiLoGo-U near the ribs.

\FloatBarrier
\subsection{Computational cost}
\label{sec:results_cost}

The computational cost is compared on the Corner and Corner+Rib datasets, which have the largest meshes among the three datasets, with 8,008 and 31,595 elements, respectively.
The cost on the Corner dataset is listed in Table~\ref{tab:cost_corner}.
BiLoGo-T reduces the peak training GPU memory by 69.1\%, the training time per epoch by 55.1\% and the inference time per simulation by 57.0\% compared with CAtt-BiUGNN, while giving lower displacement and threshold-based thinning errors, as shown in Section~\ref{sec:results_corner}.
The difference in memory comes mainly from the number of parameters.
The inter-level mappings of CAtt-BiUGNN use a separate weight on each inter-level edge, and these weights are not shared between edges.
Its number of parameters therefore grows with the number of inter-level edges and with the mesh, from 239.5M on the Corner mesh to 678.6M on the Corner+Rib mesh (Table~\ref{tab:cost_rib}).
These parameters, together with their gradients and optimiser states, take most of its training memory.
In BiLoGo, the weights of the message-passing layers and the global module are shared across all entities, so its number of parameters does not depend on the mesh, and its training memory is mainly used by the activations.
Transolver and GNN-Transformer require less memory than CAtt-BiUGNN, but give higher errors.

\begin{table}[!htbp]
\centering
\caption{Computational cost on the Corner dataset, measured on an NVIDIA L4 GPU.}
\label{tab:cost_corner}
\resizebox{\textwidth}{!}{%
\begin{tabular}{lcccc}
\hline
Model & Trainable & Training time & Training GPU & Inference time \\
 & parameters & (s/epoch) & memory (GiB) & (s/simulation) \\
\hline
UMGN \cite{Zhao2026RUGNN} & 82.9M & 49.25 & 1.590 & 1.025 \\
RUGNN \cite{Zhao2026RUGNN} & 85.3M & 65.99 & 1.753 & 0.986 \\
CAtt-BiUGNN \cite{Zhao2026CAttBiGNN} & 239.5M & 129.11 & 4.497 & 1.285 \\
Transolver \cite{Wu2024Transolver} & 0.31M & 24.89 & 0.447 & 0.427 \\
GNN-Transformer & 0.97M & 37.62 & 0.815 & 0.482 \\
BiLoGo-T & 2.13M & 57.97 & 1.389 & 0.553 \\
\hline
\end{tabular}}
\end{table}

\begin{figure}[!htbp]
\centering
\includegraphics[width=\textwidth]{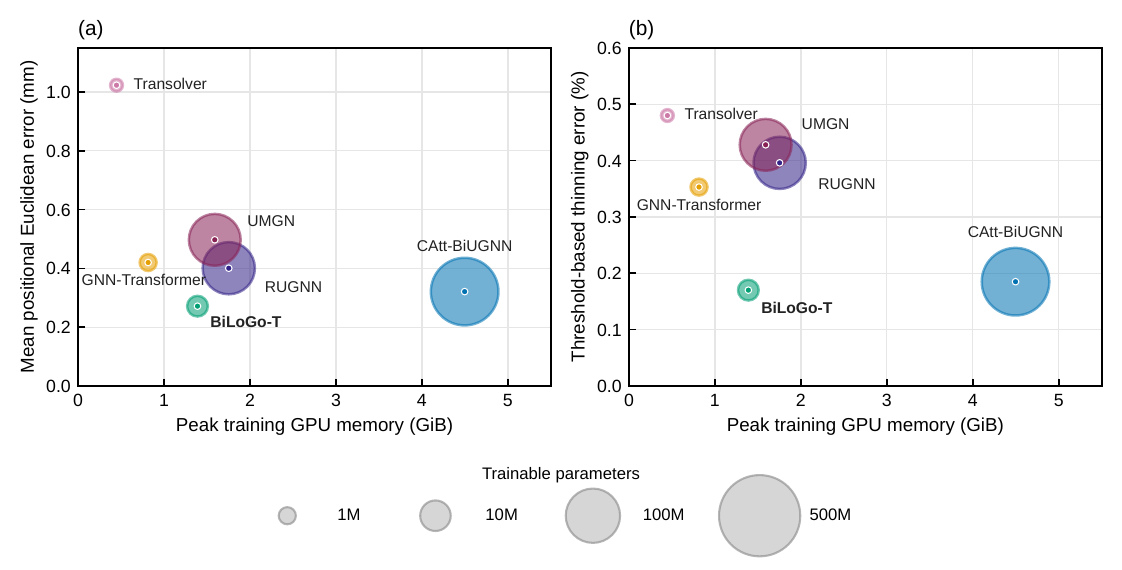}
\caption{Prediction accuracy, peak training GPU memory and number of trainable parameters of the models in Table~\ref{tab:cost_corner}. (a) Mean positional Euclidean error. (b) Threshold-based thinning error. The bubble area increases with the square root of the number of trainable parameters, as indicated by the reference bubbles.}
\label{fig:accuracy_memory_corner}
\end{figure}

The relation between the accuracy, the training memory and the number of parameters on the Corner dataset is shown in Fig.~\ref{fig:accuracy_memory_corner}.
BiLoGo-T gives the lowest positional and threshold-based thinning errors, and requires less training memory than all hierarchical graph models.
Its training memory is approximately one third of that of CAtt-BiUGNN, and the three hierarchical graph models have approximately 40 to 110 times as many parameters as BiLoGo-T.
The models with lower memory than BiLoGo-T, namely Transolver and GNN-Transformer, give larger errors.

The cost on the Corner+Rib dataset is listed in Table~\ref{tab:cost_rib}.
Compared with CAtt-BiUGNN, BiLoGo-U reduces the peak GPU memory during training by 66.1\%, from 12.86~GiB to 4.36~GiB, with a similar training time per epoch of 132.7~s and 134.6~s, respectively.
As the mesh grows from the Corner to the Corner+Rib dataset, the number of parameters of CAtt-BiUGNN increases from 239.5M to 678.6M, while the number of parameters of BiLoGo remains unchanged.
The reduction in the peak training memory remains about two thirds on both datasets.
The peak GPU memory during inference is also reduced from 3.58~GiB to 0.62~GiB, since the parameters of BiLoGo take a negligible amount of memory.
The inference time per simulation is similar for all models, between 6.7 and 7.6~s, since it is mainly spent on the recomputation of the contact inputs at each step, which is the same for all models.
This time is longer than on the Corner dataset, since the Corner+Rib mesh has approximately four times as many nodes for which the contact inputs are recomputed.
Reducing the time of the contact recomputation, for example by performing it on the GPU and by processing several simulations in a batch, is an important direction for future optimisation.
Among the BiLoGo variants, BiLoGo-U reduces the peak training memory by 64.6\% and the training time per epoch by 33.9\% compared with BiLoGo with the GINO latent grid, which gives the second-lowest errors among these variants.
BiLoGo-U requires 66.6\% more training time and 20.3\% more memory than BiLoGo-T, which may be attributed to the construction of the mesh--supernode mapping and the larger number of tokens in the global interaction.

\begin{table}[!htbp]
\centering
\caption{Computational cost on the Corner+Rib dataset, measured on an NVIDIA A40 GPU.}
\label{tab:cost_rib}
\resizebox{\textwidth}{!}{%
\begin{tabular}{lccccc}
\hline
Model & Trainable & Training time & Training GPU & Inference time & Inference GPU \\
 & parameters & (s/epoch) & memory (GiB) & (s/simulation) & memory (GiB) \\
\hline
CAtt-BiUGNN \cite{Zhao2026CAttBiGNN} & 678.6M & 134.64 & 12.860 & 7.037 & 3.576 \\
Transolver \cite{Wu2024Transolver} & 0.31M & 25.33 & 0.996 & 6.721 & 0.606 \\
GNN-Transformer & 0.97M & 43.10 & 1.608 & 6.840 & 0.612 \\
UPT \cite{Alkin2024UPT} & 1.70M & 88.58 & 0.754 & 6.917 & 0.611 \\
\hline
BiLoGo-T & 2.13M & 79.68 & 3.626 & 6.876 & 0.621 \\
BiLoGo + GINO & 2.87M & 200.87 & 12.331 & 7.620 & 8.869 \\
BiLoGo-U & 2.51M & 132.73 & 4.363 & 7.182 & 0.619 \\
\hline
\end{tabular}}
\end{table}

The coarse graphs of CAtt-BiUGNN are constructed in advance during data preparation, and their construction is not included in its training and inference times.
In contrast, the supernode mapping of BiLoGo-U is constructed within the model during training and inference, and is included in these times.
Even so, the training and inference times of BiLoGo-U are similar to those of CAtt-BiUGNN.
However, the graph hierarchy and the inter-level edge weights of CAtt-BiUGNN are fixed for a given mesh.
As a result, meshes with different topologies cannot be used together to train one CAtt-BiUGNN model.
In contrast, BiLoGo does not require a predefined graph hierarchy, and its number of parameters does not depend on the mesh size, so its training is not restricted to a single mesh topology.

\begin{figure}[!htbp]
\centering
\includegraphics[width=\textwidth]{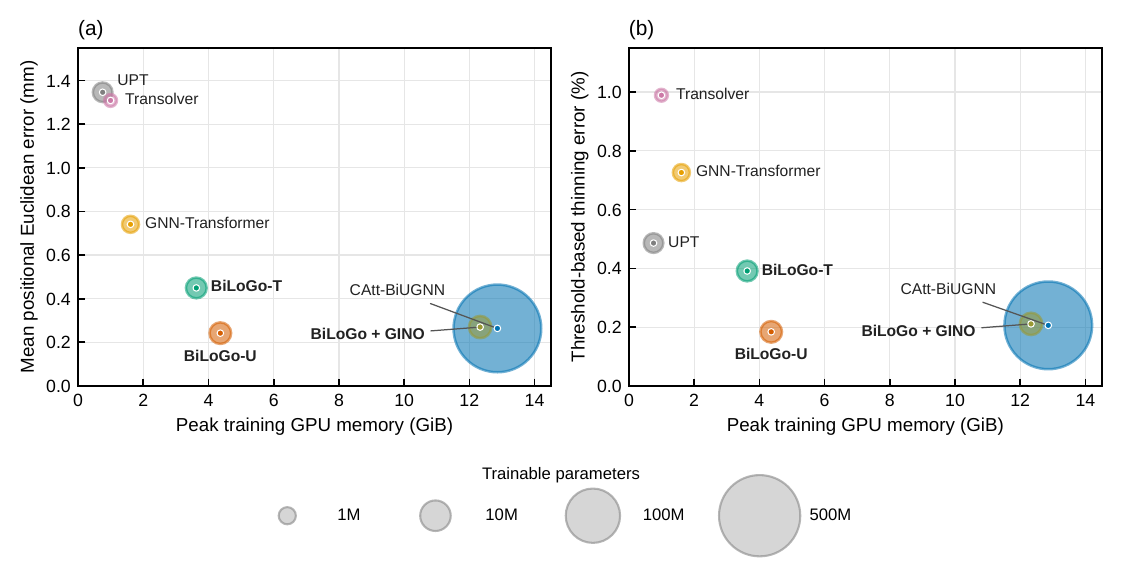}
\caption{Prediction accuracy, peak training GPU memory and number of trainable parameters of the models in Table~\ref{tab:cost_rib}. (a) Mean positional Euclidean error. (b) Threshold-based thinning error. The bubble area increases with the square root of the number of trainable parameters, as indicated by the reference bubbles.}
\label{fig:accuracy_memory}
\end{figure}

The relation between the accuracy, the training memory and the number of parameters on the Corner+Rib dataset is shown in Fig.~\ref{fig:accuracy_memory}.
BiLoGo-U gives the lowest positional and threshold-based thinning errors, and its training memory is approximately one third of that of CAtt-BiUGNN and BiLoGo with the GINO latent grid.
The models with lower memory than BiLoGo-U give larger errors.

For reference, the FE simulation of one Corner+Rib design takes approximately 1600~s on an Intel Core i7-12700 CPU with 12 cores, while the rollout of BiLoGo-U takes 7.2~s on one NVIDIA A40 GPU.

\FloatBarrier
\subsection{Selection of the global module and limitations}
\label{sec:results_limitations}
The results on the three datasets, in which the response becomes increasingly localised, show how the choice of the global module affects the accuracy.
On the Dome dataset, in which the response varies smoothly over the blank, BiLoGo-T with the physical slices gives the lowest displacement errors and the lowest threshold-based thinning error.
On the Corner dataset, BiLoGo-T also gives the lowest displacement errors and requires 69.1\% less training memory than CAtt-BiUGNN.
On the Corner+Rib dataset, in which the response changes over short distances near the ribs, the physical slices give considerably larger errors than the spatially local representations.
Therefore, when the response is concentrated near local geometric features, a spatially local tokeniser may be preferred.
Among the spatially local representations tested here, the supernode tokeniser of BiLoGo-U gives the lowest displacement and threshold-based thinning errors with considerably less GPU memory than the latent grid of GINO.

This work also has some limitations.
First, the mesh topology is the same for all simulations within each dataset.
Applying BiLoGo to meshes that change between simulations and to larger components is a natural extension of this work.
Second, the kinematics-conditioned element update is based on the membrane deformation of the elements, which dominates the thinning in the present simulations, as discussed in Section~\ref{sec:problem_local}.
For processes in which bending has a larger effect on the thinning, the descriptor may be extended with information on the change in curvature.
Third, the process conditions, including the material, the friction coefficient, the blank-holder force and the stamping speed, are fixed within each dataset, and the temperature and the stress state of the blank are not used as inputs.
The effect of these quantities on the material flow, which can be significant in hot stamping, is therefore represented only implicitly through the training data.
Fourth, the thinning is used here as the indicator of formability.
In practice, splitting also depends on the strain path and the forming limit of the material, which are not predicted by the model.

\section{Conclusion}
\label{sec:conclusion}

In sheet forming, localised responses, such as the thinning near sharp radii, corners and reinforcing ribs, often determine whether a component can be formed.
Transformer-based neural operators propagate information efficiently across the mesh, but tend to smooth such localised responses.
This study developed BiLoGo to reduce this smoothing while keeping the low memory cost of a Transformer-based neural operator.
Guided by the FE update, BiLoGo connects local node--element message passing with a Transformer-based neural operator through three modules.
The bidirectional node--element GraphSAGE model predicts each quantity on its own FE entity.
The dual-stream Transformer-based neural operator compresses the blank state and the tool contact into separate tokens, so that the contact information is not averaged with the state features.
The fusion and element update module keeps the local features in the gated fusion and conditions the element update on the kinematics of the element nodes.
BiLoGo was evaluated on three forming datasets with increasingly localised responses, namely the Dome, Corner and Corner+Rib datasets.
The main conclusions are as follows.

First, BiLoGo reduces the smoothing of localised responses.
On all three datasets, BiLoGo gave the lowest positional and threshold-based thinning errors among the tested models, including on the Corner+Rib dataset, in which the response is concentrated near the ribs.
Near the ribs, the smoothing index of the predicted thinning increased from 0.41--0.56 for the Transformer-based neural operators to 0.87 for BiLoGo-T and 0.94 for BiLoGo-U, which use the same types of tokeniser.
Near the corner on the Corner dataset, it increased from 0.45 for Transolver to 1.00 for BiLoGo-T.
The ablation studies showed that removing any of the five components increased all four errors on the Corner+Rib dataset, and the kinematics-conditioned element update mainly reduced the thinning error.

Second, this improvement is achieved with a low memory cost.
Compared with the hierarchical GNN, which gives the lowest errors among the existing models on the Corner and Corner+Rib datasets, BiLoGo-T reduced the mean positional error by 15.5\% on the Corner dataset and BiLoGo-U by 8.3\% on the Corner+Rib dataset.
The peak GPU memory during training was reduced by 69.1\% and 66.1\%, respectively, and the number of parameters of BiLoGo does not change with the mesh.

Third, the global module of BiLoGo can be replaced, and the tokeniser affects how well localised responses are predicted.
On the Dome and Corner datasets, BiLoGo-T with physical slices gave the lowest mean errors among the tested models.
On the Corner+Rib dataset, in which the response is concentrated near the ribs, BiLoGo-U with spatial supernodes reduced the mean positional error by 46.2\% compared with BiLoGo-T with feature-based physical slices.

These results indicate that the proposed BiLoGo can reduce the smoothing of localised responses while keeping the low memory cost of Transformer-based neural operators.
This may make learned simulators more suitable for formability assessment in design studies that involve local geometric features such as ribs.
Future work will apply transfer learning to adapt the trained model to new components and forming processes.
Datasets in which the global geometry and the local geometric features vary at the same time will also be studied.
In addition, since the number of parameters of BiLoGo does not depend on the mesh, the framework will be extended to meshes that change between simulations, so that it can be applied to more diverse stamped components.

\FloatBarrier

\section*{Acknowledgement}
The authors acknowledge funding support from UKRI (UKRI221: AI-Driven Design for Forming High-Performance Vehicle Parts), as well as PhD scholarships from Imperial College London. The authors would like to thank Keysight for their technical support with the PAM-STAMP software. For the purpose of open access, the authors have applied a Creative Commons Attribution (CC BY) license to any Author Accepted Manuscript version arising.

\section*{CRediT authorship contribution statement}

\textbf{Yingxue Zhao}: Conceptualization, Methodology, Software, Investigation, Data curation, Formal analysis, Validation, Visualization, Writing – original draft, Writing – review \& editing.
\textbf{Haoran Li}: Methodology, Formal analysis, Validation, Writing – review \& editing.
\textbf{Haosu Zhou}: Methodology, Formal analysis, Writing – review \& editing.
\textbf{Tobias Pfaff}: Methodology, Validation, Supervision, Writing – review \& editing.
\textbf{Nan Li}: Conceptualization, Methodology, Formal analysis, Validation, Supervision, Project administration, Funding acquisition, Writing – review \& editing.

\section*{Declaration of competing interest}

The authors declare that they have no known competing financial interests
or personal relationships that could have appeared to influence the work
reported in this paper.

\section*{Data availability}
The data supporting the findings of this study are available upon reasonable request.

\appendix
\setcounter{figure}{0}
\setcounter{table}{0}

\section{Slice assignment of BiLoGo-T}
\label{app:slices}

\begin{figure}[p]
\centering
\includegraphics[width=\textwidth]{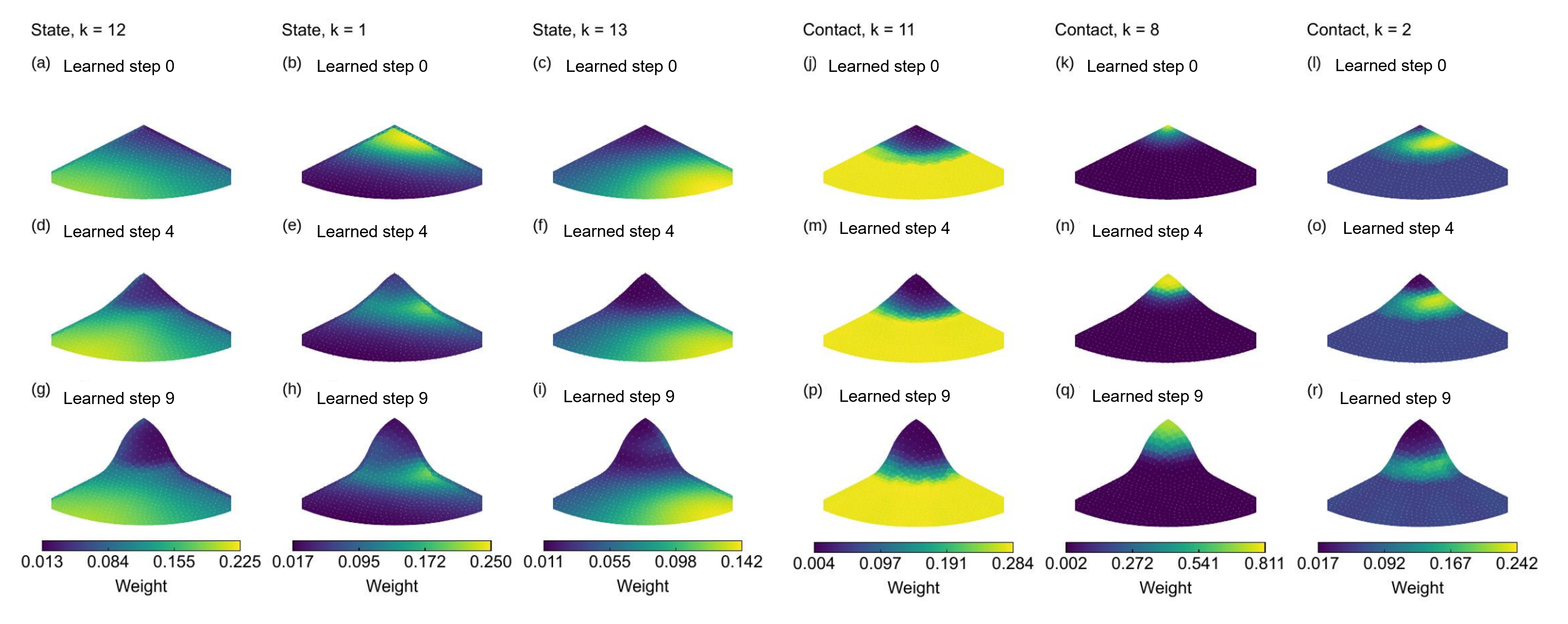}
\caption{Assignment weights of three slices of BiLoGo-T for one test simulation of the Dome dataset at learned steps 0, 4 and 9, with the FE state as input. (a)--(i) State stream. (j)--(r) Contact stream. Here, $k$ denotes the slice index. The three panels of each slice share one colour scale, and different slices use different colour scales.}
\label{fig:token_dome}
\end{figure}

\begin{figure}[p]
\centering
\includegraphics[width=\textwidth]{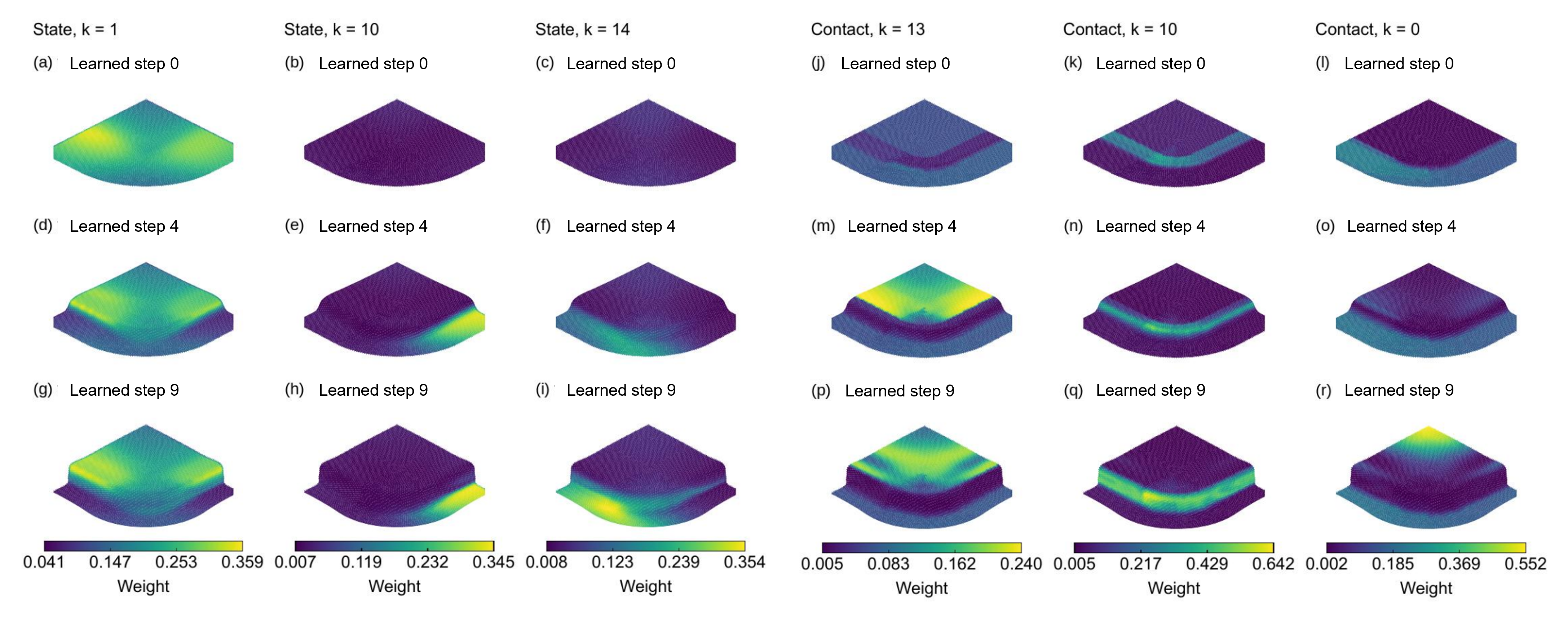}
\caption{Assignment weights of three slices of BiLoGo-T for one test simulation of the Corner dataset at learned steps 0, 4 and 9, with the FE state as input. (a)--(i) State stream. (j)--(r) Contact stream. Here, $k$ denotes the slice index. The three panels of each slice share one colour scale, and different slices use different colour scales.}
\label{fig:token_corner}
\end{figure}

Figs.~\ref{fig:token_dome} and \ref{fig:token_corner} show the assignment weights of three slices in the state and contact streams of BiLoGo-T on the Dome and Corner datasets, for one test simulation of each dataset at learned steps 0, 4 and 9 with the FE state as input.
In both streams, the weights of each slice form continuous regions on the blank, and these regions change as the forming proceeds.
On the Dome dataset, the contact slices form bands around the centre of the dome: slice 11 has large weights in the flange, slice 8 at the top of the dome, and slice 2 in a ring on the dome wall (Fig.~\ref{fig:token_dome}(j)--(r)).
On the Corner dataset, contact slice 13 has large weights on the top face, and contact slice 10 has large weights in a curved band, which corresponds to the side wall as the forming proceeds (Fig.~\ref{fig:token_corner}(j)--(r)).
The state slices show a similar type of assignment.
For example, state slice 10 on the Corner dataset has small weights at step 0, and large weights develop in one part of the flange at step 4 and remain at step 9 (Fig.~\ref{fig:token_corner}(b), (e) and (h)).

\FloatBarrier 


 \biboptions{numbers,sort&compress}
 \bibliographystyle{elsarticle-num} 
 \bibliography{ref}

@article{Attar2021Feasibility,
  author        = {Attar, Hamid Reza and Zhou, Haosu and Foster, Alistair and Li, Nan},
  title         = {{Rapid feasibility assessment of components to be formed through hot stamping: A deep learning approach}},
  journal       = {Journal of Manufacturing Processes},
  year          = {2021},
  volume        = {68},
  pages         = {1650--1671},
  doi           = {10.1016/j.jmapro.2021.06.011},
  url           = {https://doi.org/10.1016/j.jmapro.2021.06.011},
}

@inproceedings{Attar2021HFQ,
  author        = {Attar, Hamid Reza and Zhou, Haosu and Li, Nan},
  title         = {{Deformation and thinning field prediction for HFQ\textregistered{} formed panel components using convolutional neural networks}},
  booktitle     = {IOP Conference Series: Materials Science and Engineering},
  year          = {2021},
  volume        = {1157},
  number        = {1},
  pages         = {012079},
  doi           = {10.1088/1757-899X/1157/1/012079},
  url           = {https://doi.org/10.1088/1757-899X/1157/1/012079},
}

@article{Attar2023Optimisation,
  author        = {Attar, Hamid Reza and Foster, Alistair and Li, Nan},
  title         = {{Development of a deep learning platform for sheet stamping geometry optimisation under manufacturing constraints}},
  journal       = {Engineering Applications of Artificial Intelligence},
  year          = {2023},
  volume        = {123},
  pages         = {106295},
  doi           = {10.1016/j.engappai.2023.106295},
  url           = {https://doi.org/10.1016/j.engappai.2023.106295},
}

@article{Nguyen2025FormingLimits,
  author        = {Nguyen, Nhat-Tam and Tran, Minh Tien and Nguyen, Xuan Minh and Lee, Ho Won and Kang, Seong-Hoon and Oh, Young-Seok and Kim, Hyunki and Kim, Dong-Kyu},
  title         = {{Physics-guided machine learning for forming-limit assessments of advanced high-strength steels}},
  journal       = {International Journal of Mechanical Sciences},
  year          = {2025},
  volume        = {287},
  pages         = {109959},
  doi           = {10.1016/j.ijmecsci.2025.109959},
  url           = {https://doi.org/10.1016/j.ijmecsci.2025.109959},
}

@inproceedings{Zhao2025BatteryBox,
  author        = {Zhao, Yingxue and Li, Haoran and Zhou, Haosu and Attar, Hamid Reza and Pfaff, Tobias and Li, Nan},
  title         = {{Rapid prediction of material deformation in hot stamping of battery box geometries using graph neural network}},
  booktitle     = {Journal of Physics: Conference Series},
  year          = {2025},
  volume        = {3104},
  number        = {1},
  pages         = {012057},
  doi           = {10.1088/1742-6596/3104/1/012057},
  url           = {https://doi.org/10.1088/1742-6596/3104/1/012057},
}

@article{Zhou2025BlankShape,
  author        = {Zhou, Haosu and Li, Haoran and Zhao, Yingxue and Childs, Peter R. N. and Li, Nan},
  title         = {{Image-based Artificial Intelligence-driven modelling for blank shape optimisation in sheet metal forming}},
  journal       = {Materials \& Design},
  year          = {2025},
  volume        = {256},
  pages         = {114302},
  doi           = {10.1016/j.matdes.2025.114302},
  url           = {https://doi.org/10.1016/j.matdes.2025.114302},
}

@article{Zhou2025Multilevel,
  author        = {Zhou, Haosu and Zhao, Yingxue and Li, Haoran and Pfaff, Tobias and Li, Nan},
  title         = {{A multi-level graph-based surrogate model for real-time high-fidelity sheet forming simulations}},
  journal       = {Advanced Engineering Informatics},
  year          = {2025},
  volume        = {66},
  pages         = {103458},
  doi           = {10.1016/j.aei.2025.103458},
  url           = {https://doi.org/10.1016/j.aei.2025.103458},
}

@article{Li2026Crashworthiness,
  author        = {Li, Haoran and Zhao, Yingxue and Zhou, Haosu and Pfaff, Tobias and Li, Nan},
  title         = {{A graph neural network surrogate model for mesh-based crashworthiness prediction of vehicle panel components}},
  journal       = {Results in Engineering},
  year          = {2026},
  volume        = {30},
  pages         = {110925},
  doi           = {10.1016/j.rineng.2026.110925},
  url           = {https://doi.org/10.1016/j.rineng.2026.110925},
}

@misc{Li2026MaskMorph,
  author        = {Li, Haoran and Lehrer, Tobias and Zhao, Yingxue and Zhou, Haosu and Stocker, Philipp and Pfaff, Tobias and Wagner, Marcus and Li, Nan},
  title         = {{Mask-Morph Graph U-Net: A Generalisable Mesh-Based Surrogate for Crashworthiness Field Prediction under Large Geometric Variation}},
  howpublished  = {arXiv preprint},
  year          = {2026},
  doi           = {10.48550/arXiv.2605.15231},
  eprint        = {2605.15231},
  archivePrefix = {arXiv},
  primaryClass  = {cs.LG},
  url           = {https://arxiv.org/abs/2605.15231},
}

@misc{Luo2026StampFormer,
  author        = {Luo, Jiajie and Mohamed, Mohamed and Hassan, Osama and Zhou, Haosu and Zhao, Yingxue and Li, Haoran and Li, Xinrun and Shao, Zhutao and Long, Yang and Li, Nan and Li, Jichun},
  title         = {{StampFormer: A Physics-Guided Material-Geometry-Coupled Multimodal Model for Rapid Prediction of Physical Fields in Sheet Metal Stamping}},
  howpublished  = {arXiv preprint},
  year          = {2026},
  doi           = {10.48550/arXiv.2605.18835},
  eprint        = {2605.18835},
  archivePrefix = {arXiv},
  primaryClass  = {cs.LG},
  url           = {https://arxiv.org/abs/2605.18835},
}

@article{Nguyen2026FormabilityDeepONet,
  author        = {Nguyen, Nhat-Tam and Nguyen, Xuan Minh and Nguyen, Thanh Hai and Lee, Ho Won and Kang, Seong-Hoon and Oh, Young-Seok and Kim, Dong-Kyu},
  title         = {{Physics-informed deep operator network for formability evaluation of dual-phase steels}},
  journal       = {International Journal of Mechanical Sciences},
  year          = {2026},
  volume        = {317},
  pages         = {111504},
  doi           = {10.1016/j.ijmecsci.2026.111504},
  url           = {https://doi.org/10.1016/j.ijmecsci.2026.111504},
}

@article{Wang2026Springback,
  author        = {Wang, Hui and Lim, Heonyong and Yoon, Jeong Whan},
  title         = {{Stress-aware graph neural network for springback prediction considering material variations}},
  journal       = {Journal of Materials Processing Technology},
  year          = {2026},
  volume        = {348},
  pages         = {119189},
  doi           = {10.1016/j.jmatprotec.2025.119189},
  url           = {https://doi.org/10.1016/j.jmatprotec.2025.119189},
}

@misc{Zhao2026CAttBiGNN,
  author        = {Zhao, Yingxue and Li, Haoran and Zhou, Haosu and Pfaff, Tobias and Li, Nan},
  title         = {{A finite-element-inspired bipartite graph learned simulator for manufacturability assessment in large-deformation sheet forming}},
  howpublished  = {arXiv preprint},
  year          = {2026},
  doi           = {10.48550/arXiv.2605.22845},
  eprint        = {2605.22845},
  archivePrefix = {arXiv},
  primaryClass  = {cs.CE},
  url           = {https://arxiv.org/abs/2605.22845},
}

@article{Zhao2026RUGNN,
  author        = {Zhao, Yingxue and Chen, Qianyi and Li, Haoran and Zhou, Haosu and Attar, Hamid Reza and Pfaff, Tobias and Wu, Tailin and Li, Nan},
  title         = {{Recurrent U-Net-based Graph Neural Network (RUGNN) for accurate deformation predictions in sheet material forming}},
  journal       = {Advanced Engineering Informatics},
  year          = {2026},
  volume        = {69},
  pages         = {104021},
  doi           = {10.1016/j.aei.2025.104021},
  url           = {https://doi.org/10.1016/j.aei.2025.104021},
}

@inproceedings{Pfaff2021MeshGraphNet,
  author        = {Pfaff, Tobias and Fortunato, Meire and Sanchez-Gonzalez, Alvaro and Battaglia, Peter W.},
  title         = {{Learning Mesh-Based Simulation with Graph Networks}},
  booktitle     = {International Conference on Learning Representations},
  year          = {2021},
  url           = {https://openreview.net/forum?id=roNqYL0_XP},
}

@article{Maurizi2022MechanicalFields,
  author        = {Maurizi, Marco and Gao, Chao and Berto, Filippo},
  title         = {{Predicting stress, strain and deformation fields in materials and structures with graph neural networks}},
  journal       = {Scientific Reports},
  year          = {2022},
  volume        = {12},
  pages         = {21834},
  doi           = {10.1038/s41598-022-26424-3},
  url           = {https://doi.org/10.1038/s41598-022-26424-3},
}

@article{Gladstone2024SteadyGNN,
  author        = {Gladstone, Rini Jasmine and Rahmani, Helia and Suryakumar, Vishvas and Meidani, Hadi and D'Elia, Marta and Zareei, Ahmad},
  title         = {{Mesh-based GNN surrogates for time-independent PDEs}},
  journal       = {Scientific Reports},
  year          = {2024},
  volume        = {14},
  pages         = {3394},
  doi           = {10.1038/s41598-024-53185-y},
  url           = {https://doi.org/10.1038/s41598-024-53185-y},
}

@article{Lino2022RotationMultiscale,
  author        = {Lino, Mario and Fotiadis, Stathi and Bharath, Anil A. and Cantwell, Chris D.},
  title         = {{Multi-scale rotation-equivariant graph neural networks for unsteady Eulerian fluid dynamics}},
  journal       = {Physics of Fluids},
  year          = {2022},
  volume        = {34},
  number        = {8},
  pages         = {087110},
  doi           = {10.1063/5.0097679},
  url           = {https://doi.org/10.1063/5.0097679},
}

@article{Yang2022AMGNET,
  author        = {Yang, Zhishuang and Dong, Yidao and Deng, Xiaogang and Zhang, Laiping},
  title         = {{AMGNET: multi-scale graph neural networks for flow field prediction}},
  journal       = {Connection Science},
  year          = {2022},
  volume        = {34},
  number        = {1},
  pages         = {2500--2519},
  doi           = {10.1080/09540091.2022.2131737},
  url           = {https://doi.org/10.1080/09540091.2022.2131737},
}

@article{Barwey2023GraphAutoencoder,
  author        = {Barwey, Shivam and Shankar, Varun and Viswanathan, Venkatasubramanian and Maulik, Romit},
  title         = {{Multiscale graph neural network autoencoders for interpretable scientific machine learning}},
  journal       = {Journal of Computational Physics},
  year          = {2023},
  volume        = {495},
  pages         = {112537},
  doi           = {10.1016/j.jcp.2023.112537},
  url           = {https://doi.org/10.1016/j.jcp.2023.112537},
}

@inproceedings{Cao2023BSMS,
  author        = {Cao, Yadi and Chai, Menglei and Li, Minchen and Jiang, Chenfanfu},
  title         = {{Efficient Learning of Mesh-Based Physical Simulation with Bi-Stride Multi-Scale Graph Neural Network}},
  booktitle     = {Proceedings of the 40th International Conference on Machine Learning},
  year          = {2023},
  volume        = {202},
  pages         = {3541--3558},
  series        = {Proceedings of Machine Learning Research},
  url           = {https://proceedings.mlr.press/v202/cao23a.html},
}

@article{Deshpande2024MAgNET,
  author        = {Deshpande, Saurabh and Bordas, St{\'{e}}phane P. A. and Lengiewicz, Jakub},
  title         = {{MAgNET: A graph U-Net architecture for mesh-based simulations}},
  journal       = {Engineering Applications of Artificial Intelligence},
  year          = {2024},
  volume        = {133},
  pages         = {108055},
  doi           = {10.1016/j.engappai.2024.108055},
  url           = {https://doi.org/10.1016/j.engappai.2024.108055},
}

@article{Perera2024AdaptiveMesh,
  author        = {Perera, Roberto and Agrawal, Vinamra},
  title         = {{Multiscale graph neural networks with adaptive mesh refinement for accelerating mesh-based simulations}},
  journal       = {Computer Methods in Applied Mechanics and Engineering},
  year          = {2024},
  volume        = {429},
  pages         = {117152},
  doi           = {10.1016/j.cma.2024.117152},
  url           = {https://doi.org/10.1016/j.cma.2024.117152},
}

@inproceedings{Deng2025EvoMesh,
  author        = {Deng, Huayu and Zhu, Xiangming and Wang, Yunbo and Yang, Xiaokang},
  title         = {{EvoMesh: Adaptive Physical Simulation with Hierarchical Graph Evolutions}},
  booktitle     = {Proceedings of the 42nd International Conference on Machine Learning},
  year          = {2025},
  volume        = {267},
  pages         = {13327--13346},
  series        = {Proceedings of Machine Learning Research},
  url           = {https://proceedings.mlr.press/v267/deng25j.html},
}

@inproceedings{Mousavi2025RIGNO,
  author        = {Mousavi, Sepehr and Wen, Shizheng and Lingsch, Levi E. and Herde, Maximilian and Raoni{\'{c}}, Bogdan and Mishra, Siddhartha},
  title         = {{RIGNO: A Graph-based Framework For Robust And Accurate Operator Learning For PDEs On Arbitrary Domains}},
  booktitle     = {Advances in Neural Information Processing Systems},
  year          = {2025},
  volume        = {38},
  pages         = {166615--166677},
  doi           = {10.52202/085713-5018},
  url           = {https://proceedings.nips.cc/paper_files/paper/2025/hash/dcb91f43033bb1d367d1848806dee98d-Abstract-Conference.html},
}

@inproceedings{Wurth2025ROBIN,
  author        = {W{\"{u}}rth, Tobias and Freymuth, Niklas and Neumann, Gerhard and K{\"{a}}rger, Luise},
  title         = {{Diffusion-Based Hierarchical Graph Neural Networks for Simulating Nonlinear Solid Mechanics}},
  booktitle     = {Advances in Neural Information Processing Systems},
  year          = {2025},
  volume        = {38},
  pages         = {144338--144367},
  doi           = {10.52202/085713-4336},
  url           = {https://proceedings.nips.cc/paper_files/paper/2025/hash/bcd3c132bb19544815c5e660a2b044bd-Abstract-Conference.html},
}

@article{Gao2024FEIH,
  author        = {Gao, Rui and Deo, Indu Kant and Jaiman, Rajeev K.},
  title         = {{A finite element-inspired hypergraph neural network: Application to fluid dynamics simulations}},
  journal       = {Journal of Computational Physics},
  year          = {2024},
  volume        = {504},
  pages         = {112866},
  doi           = {10.1016/j.jcp.2024.112866},
  url           = {https://doi.org/10.1016/j.jcp.2024.112866},
}

@article{Gao2024FSI,
  author        = {Gao, Rui and Jaiman, Rajeev K.},
  title         = {{Predicting fluid-structure interaction with graph neural networks}},
  journal       = {Physics of Fluids},
  year          = {2024},
  volume        = {36},
  number        = {1},
  pages         = {013622},
  doi           = {10.1063/5.0182801},
  url           = {https://doi.org/10.1063/5.0182801},
}

@article{Li2025HeterogeneousNonlinear,
  author        = {Li, Yongcheng and Wang, Changsheng and Hou, Wenbin},
  title         = {{A universal surrogate modeling method based on heterogeneous graph neural network for nonlinear analysis}},
  journal       = {Computer Methods in Applied Mechanics and Engineering},
  year          = {2025},
  volume        = {437},
  pages         = {117793},
  doi           = {10.1016/j.cma.2025.117793},
  url           = {https://doi.org/10.1016/j.cma.2025.117793},
}

@article{Rezaei2025FiniteOperator,
  author        = {Rezaei, Shahed and Najian Asl, Reza and Faroughi, Shirko and Asgharzadeh, Mahdi and Harandi, Ali and Najafi Koopas, Rasoul and Laschet, Gottfried and Reese, Stefanie and Apel, Markus},
  title         = {{A finite operator learning technique for mapping the elastic properties of microstructures to their mechanical deformations}},
  journal       = {International Journal for Numerical Methods in Engineering},
  year          = {2025},
  volume        = {126},
  number        = {1},
  pages         = {e7637},
  doi           = {10.1002/nme.7637},
  url           = {https://doi.org/10.1002/nme.7637},
}

@inproceedings{Alkin2024UPT,
  title = {Universal Physics Transformers: A Framework for Efficiently Scaling Neural Operators},
  author = {Alkin, Benedikt and F{\"u}rst, Andreas and Schmid, Simon
            and Gruber, Lukas and Holzleitner, Markus
            and Brandstetter, Johannes},
  booktitle = {Advances in Neural Information Processing Systems},
  volume = {37},
  year = {2024},
  doi = {10.52202/079017-0793}
}

@article{Wang2026NEHGNN,
  author        = {Wang, Hui and Wu, Jianjun and Liu, Long and Liu, Yvdong and Yang, Zekun},
  title         = {{Probability wrinkling analysis of spatial tubes forming considering mixed uncertainties}},
  journal       = {Thin-Walled Structures},
  year          = {2026},
  volume        = {229},
  pages         = {115152},
  doi           = {10.1016/j.tws.2026.115152},
  url           = {https://doi.org/10.1016/j.tws.2026.115152},
}

@inproceedings{Li2021FNO,
  author        = {Li, Zongyi and Kovachki, Nikola Borislavov and Azizzadenesheli, Kamyar and Liu, Burigede and Bhattacharya, Kaushik and Stuart, Andrew M. and Anandkumar, Anima},
  title         = {{Fourier Neural Operator for Parametric Partial Differential Equations}},
  booktitle     = {International Conference on Learning Representations},
  year          = {2021},
  url           = {https://openreview.net/forum?id=c8P9NQVtmnO},
}

@inproceedings{Hao2023GNOT,
  author        = {Hao, Zhongkai and Wang, Zhengyi and Su, Hang and Ying, Chengyang and Dong, Yinpeng and Liu, Songming and Cheng, Ze and Song, Jian and Zhu, Jun},
  title         = {{GNOT: A General Neural Operator Transformer for Operator Learning}},
  booktitle     = {Proceedings of the 40th International Conference on Machine Learning},
  year          = {2023},
  volume        = {202},
  pages         = {12556--12569},
  series        = {Proceedings of Machine Learning Research},
  url           = {https://proceedings.mlr.press/v202/hao23c.html},
}

@misc{Fortunato2022MSMGN,
  title = {MultiScale MeshGraphNets},
  author = {Fortunato, Meire and Pfaff, Tobias
            and Wirnsberger, Peter and Pritzel, Alexander
            and Battaglia, Peter},
  year = {2022},
  eprint = {2210.00612},
  archivePrefix = {arXiv},
  url = {https://arxiv.org/abs/2210.00612}
}

@inproceedings{Li2023GINO,
  author        = {Li, Zongyi and Kovachki, Nikola Borislavov and Choy, Chris and Li, Boyi and Kossaifi, Jean and Otta, Shourya Prakash and Nabian, Mohammad Amin and Stadler, Maximilian and Hundt, Christian and Azizzadenesheli, Kamyar and Anandkumar, Anima},
  title         = {{Geometry-Informed Neural Operator for Large-Scale 3D PDEs}},
  booktitle     = {Advances in Neural Information Processing Systems},
  year          = {2023},
  volume        = {36},
  doi           = {10.52202/075280-1556},
  url           = {https://proceedings.neurips.cc/paper_files/paper/2023/hash/70518ea42831f02afc3a2828993935ad-Abstract-Conference.html},
}

@inproceedings{Wu2024Transolver,
  author        = {Wu, Haixu and Luo, Huakun and Wang, Haowen and Wang, Jianmin and Long, Mingsheng},
  title         = {{Transolver: A Fast Transformer Solver for PDEs on General Geometries}},
  booktitle     = {Proceedings of the 41st International Conference on Machine Learning},
  year          = {2024},
  volume        = {235},
  pages         = {53681--53705},
  series        = {Proceedings of Machine Learning Research},
  url           = {https://proceedings.mlr.press/v235/wu24r.html},
}

@article{Chen2025LocalityNO,
  author        = {Chen, Jiangce and Xu, Wenzhuo and Xu, Zeda and Grande Guti{\'{e}}rrez, Noelia and Narra, Sneha Prabha and McComb, Christopher},
  title         = {{Enforcing the principle of locality for physical simulations with neural operators}},
  journal       = {Journal of Computational Physics},
  year          = {2025},
  volume        = {538},
  pages         = {114131},
  doi           = {10.1016/j.jcp.2025.114131},
  url           = {https://doi.org/10.1016/j.jcp.2025.114131},
}

@article{Kaewnuratchadasorn2025GeoPhysicsNO,
  author        = {Kaewnuratchadasorn, Chawit and Wang, Jiaji and Kim, Chul-Woo and Deng, Xiaowei},
  title         = {{Geometry physics neural operator solver for solid mechanics}},
  journal       = {Computer-Aided Civil and Infrastructure Engineering},
  year          = {2025},
  volume        = {40},
  number        = {10},
  pages         = {1388--1404},
  doi           = {10.1111/mice.13405},
  url           = {https://doi.org/10.1111/mice.13405},
}

@article{Lei2025M4GN,
  author        = {Lei, Bo and Castillo, Victor M. and Hu, Yeping},
  title         = {{M4GN: Mesh-based Multi-segment Hierarchical Graph Network for Dynamic Simulations}},
  journal       = {Transactions on Machine Learning Research},
  year          = {2025},
  volume        = {2025},
  url           = {https://openreview.net/forum?id=R3vDbqWa1v},
}

@inproceedings{Luo2025TransolverPP,
  author        = {Luo, Huakun and Wu, Haixu and Zhou, Hang and Xing, Lanxiang and Di, Yichen and Wang, Jianmin and Long, Mingsheng},
  title         = {{Transolver++: An Accurate Neural Solver for PDEs on Million-Scale Geometries}},
  booktitle     = {Proceedings of the 42nd International Conference on Machine Learning},
  year          = {2025},
  volume        = {267},
  pages         = {41432--41449},
  series        = {Proceedings of Machine Learning Research},
  url           = {https://proceedings.mlr.press/v267/luo25o.html},
}

@misc{Iparraguirre2026MGNT,
  author        = {Iparraguirre, Mikel M. and Alfaro, Iciar and Gonzalez, David and Cueto, Elias},
  title         = {{MeshGraphNet-Transformer: Scalable Mesh-based Learned Simulation for Solid Mechanics}},
  howpublished  = {arXiv preprint},
  year          = {2026},
  doi           = {10.48550/arXiv.2601.23177},
  eprint        = {2601.23177},
  archivePrefix = {arXiv},
  primaryClass  = {cs.LG},
  url           = {https://arxiv.org/abs/2601.23177},
}

@inproceedings{Zhou2026Transolver3,
  author        = {Zhou, Hang and Wu, Haixu and Shangguan, Haonan and Ma, Yuezhou and Weng, Huikun and Wang, Jianmin and Long, Mingsheng},
  title         = {{Transolver-3: Scaling Up Transformer Solvers to Industrial-Scale Geometries}},
  booktitle     = {Proceedings of the 43rd International Conference on Machine Learning},
  year          = {2026},
  volume        = {306},
  pages         = {166009--166030},
  series        = {Proceedings of Machine Learning Research},
  url           = {https://proceedings.mlr.press/v306/zhou26ay.html},
}

@article{Perera2022CrackGNN,
  author        = {Perera, Roberto and Guzzetti, Davide and Agrawal, Vinamra},
  title         = {{Graph neural networks for simulating crack coalescence and propagation in brittle materials}},
  journal       = {Computer Methods in Applied Mechanics and Engineering},
  year          = {2022},
  volume        = {395},
  pages         = {115021},
  doi           = {10.1016/j.cma.2022.115021},
  url           = {https://doi.org/10.1016/j.cma.2022.115021},
}

@article{Xu2022SuperMeshing,
  author        = {Xu, Qingfeng and Nie, Zhenguo and Xu, Handing and Zhou, Haosu and Attar, Hamid Reza and Li, Nan and Xie, Fugui and Liu, Xin-Jun},
  title         = {{SuperMeshing: A new deep learning architecture for increasing the mesh density of physical fields in metal forming numerical simulation}},
  journal       = {Journal of Applied Mechanics},
  year          = {2022},
  volume        = {89},
  number        = {1},
  pages         = {011002},
  doi           = {10.1115/1.4052195},
  url           = {https://doi.org/10.1115/1.4052195},
}

@article{Zhou2024CrackMPNN,
  author        = {Zhou, X.-P. and Feng, K.},
  title         = {{MPNN based graph networks as learnable physics engines for deformation and crack propagation in solid mechanics}},
  journal       = {International Journal of Solids and Structures},
  year          = {2024},
  volume        = {291},
  pages         = {112695},
  doi           = {10.1016/j.ijsolstr.2024.112695},
  url           = {https://doi.org/10.1016/j.ijsolstr.2024.112695},
}

@article{Barwey2025MeshSuperResolution,
  author        = {Barwey, Shivam and Pal, Pinaki and Patel, Saumil and Balin, Riccardo and Lusch, Bethany and Vishwanath, Venkatram and Maulik, Romit and Balakrishnan, Ramesh},
  title         = {{Mesh-based super-resolution of fluid flows with multiscale graph neural networks}},
  journal       = {Computer Methods in Applied Mechanics and Engineering},
  year          = {2025},
  volume        = {443},
  pages         = {118072},
  doi           = {10.1016/j.cma.2025.118072},
  url           = {https://doi.org/10.1016/j.cma.2025.118072},
}

@article{GuevaraGarban2025LocalFields,
  author        = {Guevara Garban, Manuel Ricardo and Chemisky, Yves and Cl{\'{e}}ment, Micha{\"{e}}l and Pruli{\`{e}}re, {\'{E}}tienne},
  title         = {{Physics-informed graph neural networks to reconstruct local fields considering finite strain hyperelasticity}},
  journal       = {International Journal for Numerical Methods in Engineering},
  year          = {2025},
  volume        = {126},
  number        = {24},
  pages         = {e70193},
  doi           = {10.1002/nme.70193},
  url           = {https://doi.org/10.1002/nme.70193},
}

@article{Jakobsen2026PlasticLocalization,
  author        = {Jakobsen, Rasmus B. and Kristensen, Tobias S. and Storm, Joep and Rocha, Iuri B. C. M. and Andriollo, Tito},
  title         = {{Predicting plastic strain localization in porous solids using graph neural networks}},
  journal       = {Mechanics of Materials},
  year          = {2026},
  volume        = {219},
  pages         = {105733},
  doi           = {10.1016/j.mechmat.2026.105733},
  url           = {https://doi.org/10.1016/j.mechmat.2026.105733},
}

@article{Lai2026LocallyEnhanced,
  author        = {Lai, Zhihong and Zhao, Luyang and Shao, Qian},
  title         = {{Locally enhanced neural networks for discontinuities in solid mechanics}},
  journal       = {International Journal of Mechanical Sciences},
  year          = {2026},
  volume        = {322},
  pages         = {111660},
  doi           = {10.1016/j.ijmecsci.2026.111660},
  url           = {https://doi.org/10.1016/j.ijmecsci.2026.111660},
}

@article{Chen2024PIERS,
  author        = {Chen, Qianyi and Cao, Jiannong and Lin, Wanyu and Zhu, Songye and Wang, Sumei},
  title         = {{Predicting dynamic responses of continuous deformable bodies: A graph-based learning approach}},
  journal       = {Computer Methods in Applied Mechanics and Engineering},
  year          = {2024},
  volume        = {420},
  pages         = {116669},
  doi           = {10.1016/j.cma.2023.116669},
  url           = {https://doi.org/10.1016/j.cma.2023.116669},
}

@article{Li2025DynamicImpactGNN,
  author        = {Li, Qilin and Huang, Zhijie and Shao, Yanda and Li, Ling and Chen, Wensu and Hao, Hong},
  title         = {{Dynamic graph-based approach for prediction of spatiotemporal response of RC structure to impact loads}},
  journal       = {Computers \& Structures},
  year          = {2025},
  volume        = {316},
  pages         = {107861},
  doi           = {10.1016/j.compstruc.2025.107861},
  url           = {https://doi.org/10.1016/j.compstruc.2025.107861},
}

@article{Tesan2025DigitalHuman,
  author        = {Tes{\'{a}}n, Lucas and Gonz{\'{a}}lez, David and Martins, Pedro and Cueto, El{\'{i}}as},
  title         = {{Thermodynamics-informed graph neural networks for real-time simulation of digital human twins}},
  journal       = {Computational Mechanics},
  year          = {2025},
  volume        = {76},
  pages         = {923--944},
  doi           = {10.1007/s00466-025-02633-1},
  url           = {https://doi.org/10.1007/s00466-025-02633-1},
}

@article{Dubey2026ContactGNN,
  author        = {Dubey, Vijay K. and Haese, Collin E. and G{\"{u}}ltekin, Osman and Dalton, David and Rausch, Manuel K. and Fuhg, Jan},
  title         = {{Graph neural network surrogates for contacting deformable bodies with necessary and sufficient contact detection}},
  journal       = {Computer Methods in Applied Mechanics and Engineering},
  year          = {2026},
  volume        = {448},
  pages         = {118413},
  doi           = {10.1016/j.cma.2025.118413},
  url           = {https://doi.org/10.1016/j.cma.2025.118413},
}

@article{Hernandez2024ThermoGNN,
  author        = {Hern{\'{a}}ndez, Quercus and Bad{\'{i}}as, Alberto and Chinesta, Francisco and Cueto, El{\'{i}}as},
  title         = {{Thermodynamics-informed graph neural networks}},
  journal       = {IEEE Transactions on Artificial Intelligence},
  year          = {2024},
  volume        = {5},
  number        = {3},
  pages         = {967--976},
  doi           = {10.1109/TAI.2022.3179681},
  url           = {https://doi.org/10.1109/TAI.2022.3179681},
}

@article{Wurth2024PIMGN,
  author        = {W{\"{u}}rth, Tobias and Freymuth, Niklas and Zimmerling, Clemens and Neumann, Gerhard and K{\"{a}}rger, Luise},
  title         = {{Physics-informed MeshGraphNets (PI-MGNs): Neural finite element solvers for non-stationary and nonlinear simulations on arbitrary meshes}},
  journal       = {Computer Methods in Applied Mechanics and Engineering},
  year          = {2024},
  volume        = {429},
  pages         = {117102},
  doi           = {10.1016/j.cma.2024.117102},
  url           = {https://doi.org/10.1016/j.cma.2024.117102},
}

@article{Tierz2025LocalThermoGNN,
  author        = {Tierz, Alicia and Alfaro, Ic{\'{i}}ar and Gonz{\'{a}}lez, David and Chinesta, Francisco and Cueto, El{\'{i}}as},
  title         = {{Graph neural networks informed locally by thermodynamics}},
  journal       = {Engineering Applications of Artificial Intelligence},
  year          = {2025},
  volume        = {144},
  pages         = {110108},
  doi           = {10.1016/j.engappai.2025.110108},
  url           = {https://doi.org/10.1016/j.engappai.2025.110108},
}

@article{Lu2021DeepONet,
  author        = {Lu, Lu and Jin, Pengzhan and Pang, Guofei and Zhang, Zhongqiang and Karniadakis, George Em},
  title         = {{Learning nonlinear operators via DeepONet based on the universal approximation theorem of operators}},
  journal       = {Nature Machine Intelligence},
  year          = {2021},
  volume        = {3},
  pages         = {218--229},
  doi           = {10.1038/s42256-021-00302-5},
  url           = {https://doi.org/10.1038/s42256-021-00302-5},
}

@article{Kovachki2023NeuralOperator,
  author        = {Kovachki, Nikola and Li, Zongyi and Liu, Burigede and Azizzadenesheli, Kamyar and Bhattacharya, Kaushik and Stuart, Andrew and Anandkumar, Anima},
  title         = {{Neural Operator: Learning Maps Between Function Spaces With Applications to PDEs}},
  journal       = {Journal of Machine Learning Research},
  year          = {2023},
  volume        = {24},
  number        = {89},
  pages         = {1--97},
  url           = {https://www.jmlr.org/papers/v24/21-1524.html},
}

@article{Herrmann2024DLMechanicsReview,
  author        = {Herrmann, Leon and Kollmannsberger, Stefan},
  title         = {{Deep learning in computational mechanics: a review}},
  journal       = {Computational Mechanics},
  year          = {2024},
  volume        = {74},
  pages         = {281--331},
  doi           = {10.1007/s00466-023-02434-4},
  url           = {https://doi.org/10.1007/s00466-023-02434-4},
}

@article{Zhao2024GNNReview,
  author        = {Zhao, Yingxue and Li, Haoran and Zhou, Haosu and Attar, Hamid Reza and Pfaff, Tobias and Li, Nan},
  title         = {{A review of graph neural network applications in mechanics-related domains}},
  journal       = {Artificial Intelligence Review},
  year          = {2024},
  volume        = {57},
  number        = {11},
  pages         = {315},
  doi           = {10.1007/s10462-024-10931-y},
  url           = {https://doi.org/10.1007/s10462-024-10931-y},
}

@article{Lim2021TFT,
  title   = {Temporal Fusion Transformers for interpretable multi-horizon time series forecasting},
  author  = {Lim, Bryan and Ar{\i}k, Sercan {\"O}. and Loeff, Nicolas and Pfister, Tomas},
  journal = {International Journal of Forecasting},
  volume  = {37},
  number  = {4},
  pages   = {1748--1764},
  year    = {2021}
}

@inproceedings{Bachlechner2021ReZero,
  title     = {{ReZero} is all you need: Fast convergence at large depth},
  author    = {Bachlechner, Thomas and Majumder, Bodhisattwa Prasad and Mao, Henry and Cottrell, Gary and McAuley, Julian},
  booktitle = {Proceedings of the 37th Conference on Uncertainty in Artificial Intelligence},
  series    = {Proceedings of Machine Learning Research},
  volume    = {161},
  pages     = {1352--1361},
  year      = {2021}
}

@inproceedings{Hamilton2017GraphSAGE,
  title     = {Inductive representation learning on large graphs},
  author    = {Hamilton, William L. and Ying, Rex and Leskovec, Jure},
  booktitle = {Advances in Neural Information Processing Systems},
  volume    = {30},
  year      = {2017}
}

@inproceedings{Kingma2015Adam,
  title     = {Adam: A method for stochastic optimization},
  author    = {Kingma, Diederik P. and Ba, Jimmy},
  booktitle = {International Conference on Learning Representations},
  year      = {2015}
}

@article{Azizzadenesheli2024NOReview,
  author        = {Azizzadenesheli, Kamyar and Kovachki, Nikola and Li, Zongyi and Liu-Schiaffini, Miguel and Kossaifi, Jean and Anandkumar, Anima},
  title         = {Neural operators for accelerating scientific simulations and design},
  journal       = {Nature Reviews Physics},
  year          = {2024},
  doi           = {10.1038/s42254-024-00712-5},
  url           = {https://doi.org/10.1038/s42254-024-00712-5},
}

@article{Khodakarami2026SpectralBias,
  author        = {Khodakarami, Siavash and Oommen, Vivek and Bora, Aniruddha and Karniadakis, George Em},
  title         = {Mitigating spectral bias in neural operators via high-frequency scaling for physical systems},
  journal       = {Neural Networks},
  volume        = {193},
  pages         = {108027},
  year          = {2026},
  doi           = {10.1016/j.neunet.2025.108027},
  url           = {https://doi.org/10.1016/j.neunet.2025.108027},
}

@article{Shao2017HFQFormability,
  author        = {Shao, Zhutao and Li, Nan and Lin, Jianguo and Dean, Trevor},
  title         = {{Formability evaluation for sheet metals under hot stamping conditions by a novel biaxial testing system and a new materials model}},
  journal       = {International Journal of Mechanical Sciences},
  year          = {2017},
  volume        = {120},
  pages         = {149--158},
  doi           = {10.1016/j.ijmecsci.2016.11.022}
}

@article{Mohamed2012AA6082,
  author        = {Mohamed, Mohamed S. and Foster, Alistair D. and Lin, Jianguo and Balint, Daniel S. and Dean, Trevor A.},
  title         = {{Investigation of deformation and failure features in hot stamping of AA6082: Experimentation and modelling}},
  journal       = {International Journal of Machine Tools and Manufacture},
  year          = {2012},
  volume        = {53},
  number        = {1},
  pages         = {27--38},
  doi           = {10.1016/j.ijmachtools.2011.07.005}
}

@inproceedings{Rampasek2022GraphGPS,
  author    = {Ramp{\'a}{\v{s}}ek, Ladislav and Galkin, Michael and Dwivedi, Vijay Prakash and Luu, Anh Tuan and Wolf, Guy and Beaini, Dominique},
  title     = {{Recipe for a General, Powerful, Scalable Graph Transformer}},
  booktitle = {Advances in Neural Information Processing Systems},
  volume    = {35},
  year      = {2022}
}

@inproceedings{Janny2023EAGLE,
  author    = {Janny, Steeven and B{\'e}n{\'e}teau, Aur{\'e}lien and Nadri, Madiha and Digne, Julie and Thome, Nicolas and Wolf, Christian},
  title     = {{EAGLE}: Large-Scale Learning of Turbulent Fluid Dynamics with Mesh Transformers},
  booktitle = {International Conference on Learning Representations},
  year      = {2023}
}

@inproceedings{Yu2024HCMT,
  author    = {Yu, Youn-Yeol and Choi, Jeongwhan and Cho, Woojin and Lee, Kookjin and Kim, Nayong and Chang, Kiseok and Woo, ChangSeung and Kim, Ilho and Lee, SeokWoo and Yang, Joon Young and Yoon, Sooyoung and Park, Noseong},
  title     = {{Learning Flexible Body Collision Dynamics with Hierarchical Contact Mesh Transformer}},
  booktitle = {International Conference on Learning Representations},
  year      = {2024}
}

@inproceedings{Wen2025GAOT,
  author    = {Wen, Shizheng and Kumbhat, Arsh and Lingsch, Levi and Mousavi, Sepehr and Zhao, Yizhou and Chandrashekar, Praveen and Mishra, Siddhartha},
  title     = {{Geometry Aware Operator Transformer as an Efficient and Accurate Neural Surrogate for PDEs on Arbitrary Domains}},
  booktitle = {Advances in Neural Information Processing Systems},
  volume    = {38},
  year      = {2025}
}

@inproceedings{Brandstetter2022MPPDE,
  author    = {Brandstetter, Johannes and Worrall, Daniel E. and Welling, Max},
  title     = {{Message Passing Neural PDE Solvers}},
  booktitle = {International Conference on Learning Representations},
  year      = {2022}
}

@inproceedings{Alon2021Bottleneck,
  author    = {Alon, Uri and Yahav, Eran},
  title     = {{On the Bottleneck of Graph Neural Networks and its Practical Implications}},
  booktitle = {International Conference on Learning Representations},
  year      = {2021}
}

@article{Cha2018BeadFormability,
  author  = {Cha, Wan-Gi and M{\"u}ller, Sabrina and Albers, Albert and Volk, Wolfram},
  title   = {Formability consideration during bead optimisation to stiffen deep drawn parts},
  journal = {Production Engineering},
  volume  = {12},
  number  = {6},
  pages   = {691--702},
  year    = {2018},
  doi     = {10.1007/s11740-018-0850-z}
}

@article{Ott2021BeadVariants,
  author  = {Ott, M. and Haberkern, P. and Gruber, M. and Hartmann, C. and Risch, T. and Wunderling, C. and Albers, A. and Volk, W.},
  title   = {Load-specific variant generation of bead cross sections in sheet metal components by unidirectional carbon fibre reinforcement},
  journal = {IOP Conference Series: Materials Science and Engineering},
  volume  = {1157},
  number  = {1},
  pages   = {012093},
  year    = {2021},
  doi     = {10.1088/1757-899X/1157/1/012093}
}

@article{Lehrer2027GeometryAugmented,
  author        = {Lehrer, Tobias and Stocker, Philipp and Wagner, Marcus and Duddeck, Fabian},
  title         = {{Geometry-augmented machine learning for drawability assessment across dissimilar U-shaped geometries under sparse data}},
  journal       = {Advanced Engineering Informatics},
  volume        = {77},
  pages         = {105409},
  year          = {2027},
  doi           = {10.1016/j.aei.2026.105409},
}

@article{McKay1979LHS,
  author        = {McKay, Michael D. and Beckman, Richard J. and Conover, William J.},
  title         = {{A comparison of three methods for selecting values of input variables in the analysis of output from a computer code}},
  journal       = {Technometrics},
  volume        = {21},
  number        = {2},
  pages         = {239--245},
  year          = {1979},
  doi           = {10.1080/00401706.1979.10489755},
}






\end{document}